\documentclass[letterpaper,twocolumn,10pt]{article}
\usepackage{zhanggroup}

\usepackage{tikz}
\usepackage{amsmath}
\usepackage{amssymb}
\usepackage{multirow}
\usepackage{color}
\usepackage{xcolor}
\usepackage{url}
\usepackage{caption}
\usepackage{subcaption}
\usepackage{float}
\usepackage{balance}
\usepackage{graphicx}
\usepackage{textcomp}
\usepackage{booktabs}
\usepackage{colortbl}
\usepackage[ruled,linesnumbered]{algorithm2e}

\DeclareMathOperator*{\argmax}{arg\,max}
\usepackage[absolute]{textpos}
\newcommand{\mypara}[1]{\medskip\noindent{\bf {#1}.}~}

\newcommand{\shadow}{\mathcal{S}}
\newcommand{\model}{\mathcal{M}}
\newcommand{\smodel}{\mathcal{G}}
\newcommand{\knowledge}{\Omega}
\newcommand{\attack}{\mathcal{A}}

\newcommand{\train}{\mathcal{D}_{\it train}}
\newcommand{\test}{\mathcal{D}_{\it test}}
\newcommand{\shadowData}{\mathcal{D}_{\it shadow}}

\newcommand{\targetData}{\mathcal{D}_{\it target}}

\begin{document}
% ======================================================

% ======================================================
\begin{textblock}{15}(1.9,1)
To Appear in 2021 ACM SIGSAC Conference on Computer and Communications Security, November 2021
\end{textblock}
% ======================================================

% ======================================================
\title{\bf Membership Leakage in Label-Only Exposures}
% ======================================================
\date{}

\author{
Zheng Li and Yang Zhang\ \ \
\\
\textit{CISPA Helmholtz Center for Information Security}
}

\maketitle

% ======================================================
\begin{abstract}
Machine learning (ML) has been widely adopted in various privacy-critical applications, e.g., face recognition and medical image analysis. 
However, recent research has shown that ML models are vulnerable to attacks against their training data. 
Membership inference is one major attack in this domain: Given a data sample and model, an adversary aims to determine whether the sample is part of the model's training set. 
Existing membership inference attacks leverage the confidence scores returned by the model as their inputs (score-based attacks). 
However, these attacks can be easily mitigated if the model only exposes the predicted label, i.e., the final model decision.

In this paper, we propose decision-based membership inference attacks and demonstrate that label-only exposures are also vulnerable to membership leakage. 
In particular, we develop two types of decision-based attacks, namely transfer attack and boundary attack.
Empirical evaluation shows that our decision-based attacks can achieve remarkable performance, and even outperform the previous score-based attacks in some cases.
We further present new insights on the success of membership inference based on quantitative and qualitative analysis, i.e., member samples of a model are more distant to the model's decision boundary than non-member samples.
Finally, we evaluate multiple defense mechanisms against our decision-based attacks and show that our two types of attacks can bypass most of these defenses.\footnote{Our code is available at \url{https://github.com/zhenglisec/Decision-based-MIA}.}
\end{abstract}
% ======================================================

% ======================================================
\section{Introduction}
% ======================================================

Machine learning (ML) has witnessed tremendous progress over the past decade and has been applied across a wide range of privacy-critical applications, such as face recognition~\cite{ZDH17,KSMB16} and medical image analysis~\cite{KEEKF15,SWFJH10, BFDB11}.
Such developments rely on not only novel training algorithms and architectures, but also access to sensitive and private data, such as health data. 
Various recent research~\cite{SDSOJ19,SZHBFB19,SSSS17,YGFJ18,TLGYW18,HHB19,LBG17,SSM19,HYYBGC21,LBWBWTGC18,LLR21,ZJPWLS20} has shown that ML models are vulnerable to privacy attacks.
One major attack in this domain is membership inference: An adversary aims to determine whether or not a data sample is used to train a target ML model.

\begin{figure}[!t]
\centering
\includegraphics[width=1\columnwidth]{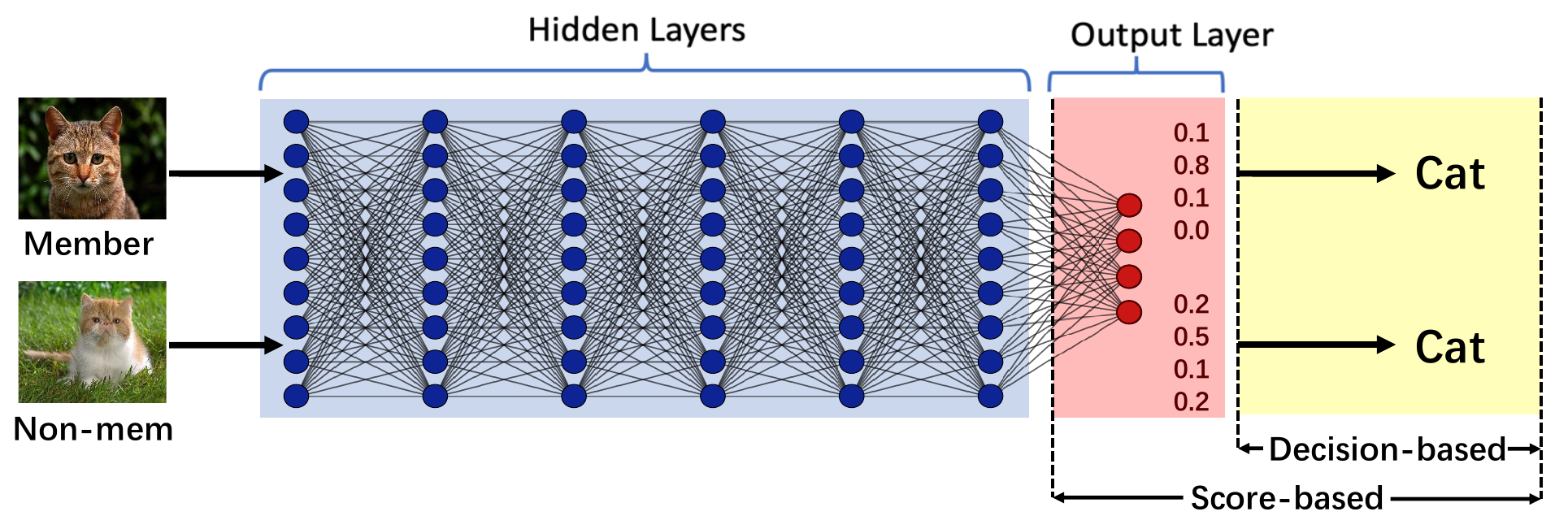}
\caption{An illustration of accessible components of the target model for each of the two threat models. A score-based threat model assumes access to the output layer; a decision-based threat model assumes access to the predicted label alone.}
\label{fig:illustration}
\end{figure}

Existing membership inference attacks~\cite{SSSS17,LBG17,SZHBFB19,YGFJ18,SSM19,HYYBGC21,LLR21} rely on the confidence scores (e.g. class probabilities or logits) returned by a target ML model as their inputs.
The success of membership inference is due to the inherent overfitting property of ML models, i.e., an ML model is more confident facing a data sample it was trained on, and this confidence is reflected in the model's output scores. 
See \autoref{fig:illustration} for an illustration of accessible components of an ML model for such score-based threat model. 
A major drawback for these score-based attacks is that they can be trivially mitigated
if the model only exposes the predicted label, i.e., the final model decision, instead of confidence scores. 
The fact that score-based attacks can be easily averted makes it more difficult to evaluate whether a model is truly vulnerable to membership inference or not, which may lead to premature claims of privacy for ML models.

This motivates us to focus on a new category of membership inference attacks that has so far received fairly little attention, namely \textit{Decision-based attacks}.
Here, the adversary solely relies on the final decision of the target model, i.e., the top-1 predicted label, as their attack model's input.
It is more realistic to evaluate the vulnerability of a machine learning system under the decision-based attacks with sole access to the model's final decision. 
First, compared to score-based attacks, decision-based attacks are much more relevant in real-world applications where confidence scores are rarely accessible. 
Furthermore, decision-based attacks have the potential to be much more robust to the state-of-the-art defenses, such as confidence score perturbation~\cite{JSBZG19,YSXCZ20,NSH18}. 
In label-only exposure, a naive \emph{baseline attack}~\cite{YGFJ18} infers that a candidate sample is a member of a target model if it is predicted correctly by the model. 
However, this baseline attack cannot distinguish between members and non-members that are both correctly classified as shown in \autoref{fig:illustration}.

In this paper, we propose two types of decision-based attacks under different scenarios, namely \emph{transfer attack} and \emph{boundary attack}. 
In the following, we abstractly introduce each of them. 

\mypara{Transfer Attack}
We assume the adversary has an auxiliary dataset (namely shadow dataset) that comes from the same distribution as the target model's training set. 
The assumption also holds for previous score-based attacks~\cite{SSSS17,LBG17,SZHBFB19,SSM19}. 
The adversary first queries the target model in a manner analog to cryptographic oracle, thereby relabeling the shadow dataset by the target model's predicted labels. 
Then, the adversary can use the relabeled shadow dataset to construct a local shadow model to mimic the behavior of the target model. 
In this way, the relabeled shadow dataset contains sufficient information from the target model, and membership information can also be transferred to the shadow model.
Finally, the adversary can leverage the shadow model to launch a score-based membership inference attack locally. 

\mypara{Boundary Attack}
Collecting data, especially sensitive and private data, is a non-trivial task.
Thus, we consider a more difficult and realistic scenario in which there is no shadow dataset and shadow model available. 
To compensate for the lack of information in this scenario, we shift the focus from the target model’s output to the input. 
Here, our key intuition is that it is harder to perturb member data samples to different classes than non-member data samples. 
The adversary queries the target model on candidate data samples, and perturb them to change the model's predicted labels. 
Then the adversary can exploit the magnitude of the perturbation to differentiate member and non-member data samples. 

Extensive experimental evaluation shows that both of our attacks achieve strong performance.
In particular, our boundary attack in some cases even outperforms the previous score-based attacks.
This demonstrates the severe membership risks stemming from ML models. 
In addition, we present a new perspective on the success of current membership inference and show that the distance between a sample and an ML model's decision boundary is strongly correlated with the sample's membership status.

Finally, we evaluate our attacks on multiple defense mechanisms: generalization enhancement\cite{SHKSS14,TLGYW18,SZHBFB19}, privacy enhancement~\cite{ACGMMTZ16} and confidence score perturbation~\cite{NSH18,JSBZG19,YSXCZ20}.
The results show that our attacks can bypass most of the defenses, unless heavy regularization is applied. 
However heavy regularization can lead to a significant degradation of the model accuracy.

In general, our contributions can be summarized as the following:
\begin{itemize}
\item We perform a systematic investigation on membership leakage in label-only exposures of ML models, and introduce decision-based membership inference attacks, which is highly relevant for real-world applications and important to gauge model privacy.
\item We propose two types of decision-based attacks under different scenarios, including transfer attack and boundary attack. Extensive experiments demonstrate that our two types of attacks achieve better performances than the baseline attack, and even outperform the previous score-based attacks in some cases.
\item We propose a new perspective on the reasons for the success of membership inference, and perform a quantitative and qualitative analysis to demonstrate that members of an ML model are more distant from the model's decision boundary than non-members.
\item We evaluate multiple defenses against our decision-based attacks and show that our novel attacks can still achieve reasonable performance unless heavy regularization is applied. 
\end{itemize}

\mypara{Paper Organization}
The rest of this paper is organized as follows. 
\autoref{sec:preli} presents the definitions of membership inference for the ML models, threat models, datasets, and model architectures used in this paper. 
\autoref{sec:adv1} and \autoref{sec:adv2} introduce our two attack methods and evaluation methods. 
In \autoref{sec:analysis}, we provide an in-depth analysis of the success of membership inference. 
\autoref{sec:defenses} provides multiple defenses against decision-based attacks. 
\autoref{sec:related} presents related work, and \autoref{sec:con} concludes the paper.

% ======================================================
\section{Preliminaries}
\label{sec:preli}
% ======================================================

% ======================================================
\subsection{Membership Leakage in Machine Learning Models}
% ======================================================

Membership leakage in ML models emerges when an adversary aims to determine whether a candidate data sample is used to train a certain ML model. 
More formally, given a candidate data sample $x$, a trained ML model $\model$, and external knowledge of an adversary, denoted by $\knowledge$, a membership inference attack $\attack$ can be defined as the following function.
\[
\attack: x, \model, \knowledge \rightarrow \{0, 1\}.
\]
Here, 0 means $x$ is not a member of $\model$'s training set and 1 otherwise. 
The attack model $\attack$ is essentially a binary classifier. 
Depending on the assumptions, it can be constructed in different ways, which will be presented in later sections.

\begin{table*}[!t]
\centering
\caption{
An overview of membership inference threat models.
``\checkmark'' means the adversary needs the knowledge and ``-'' indicates the knowledge is not necessary.
}
\scalebox{0.83}
{
\begin{tabular}{cccccccc} 
\toprule
\multirow{2}{*}{Attack Category} &\multirow{2}{*}{Attacks} & Target Model's & Training Data &Shadow& Detailed Model Prediction & Final Model Prediction\\
& & Structure &  Distribution&Model &(e.g. probabilities or logits) & (e.g. max class label) \\
\midrule
Score-based & \cite{SSSS17,LBG17,SZHBFB19,YGFJ18,SSM19,HYYBGC21,LLR21}
& \checkmark or - & \checkmark or -& \checkmark or - &  \checkmark &  \checkmark \\
\cmidrule(lr){1-7}
\multirow{3}{*}{Decision-based}& Baseline attack~\cite{YGFJ18} & - &  \checkmark & -&- &\checkmark \\
\cmidrule(lr){2-7}
& Transfer attack & - &  \checkmark & \checkmark& - &\checkmark \\
& Boundary attack & - & - & - & - &  \checkmark\\
\bottomrule
\end{tabular}
}
\label{tab:attackoverview}
\end{table*}

% ======================================================
\subsection{Threat Model}
% ======================================================

Here, we outline the threat models considered in this paper. The threat models are summarized in \autoref{tab:attackoverview}. 
There are two existing categories of attacks, i.e., score-based attacks and decision-based attacks. 
The general idea of score-based attacks is to exploit the detailed output (i.e., confidence score) of the target model to launch an attack. 
In decision-based attacks, an adversary cannot access to confidence scores, but relies on the final predictions of the target model launch an attack. 
The baseline attack predicts a data sample as a member of the training set when the model classifies it correctly. 
However, this naive and simple approach does not work at all in the case shown in \autoref{fig:illustration}. 
In the following, we introduce the adversarial knowledge that an adversary has in our decision-based attacks.

\mypara{Adversarial Knowledge} 
For our decision-based attacks, the adversary only has black-box access to the target model, i.e., they are not able to extract a candidate data sample's membership status from the confidence scores. 
Concretely, our threat model comprises the following entities. 
(1) Final decision of the target model $\model$, i.e., predicted label. 
(2) A shadow dataset $\shadowData$ drawn from the same distribution as target model's dataset $\targetData$. 
(3) A local shadow model $\shadow$ trained using the shadow dataset $\shadowData$.
For boundary attack, the adversary only has the knowledge of (1). 

% ======================================================
\subsection{Datasets and Target Model Architecture}
% ======================================================

\mypara{Datasets}
We consider four benchmark datasets of different size and complexity, namely CIFAR-10~\cite{CIFAR}, CIFAR-100~\cite{CIFAR}, GTSRB~\cite{GTSRB}, and Face~\cite{FACE}, to conduct our experiments.
Since the images in GTSRB are of different sizes, we resize them to 64$\times$64 pixels. 
For the Face dataset, we only consider people with more than 40 images, which leaves us with 19 people’s data, i.e., 19 classes. We describe them in detail in Appendix \autoref{appendix:datasets}.

\mypara{Target Model Architecture}
Typically, for image classification tasks, we use neural networks which is adopted across a wide of applications. 
In this work, we build the target model using 4 convolutional layers and 4 pooling layers with 2 hidden layers containing 256 units each at last. 
The target models are trained for 200 training epochs, iteratively using Adam algorithm with a batch-size of 128 and a fixed learning rate of 0.001. 

% ======================================================
\section{Transfer Attack}
\label{sec:adv1}
% ======================================================

\begin{figure*}[!ht]
\centering
\begin{subfigure}{0.5\columnwidth}
\includegraphics[width=\columnwidth]{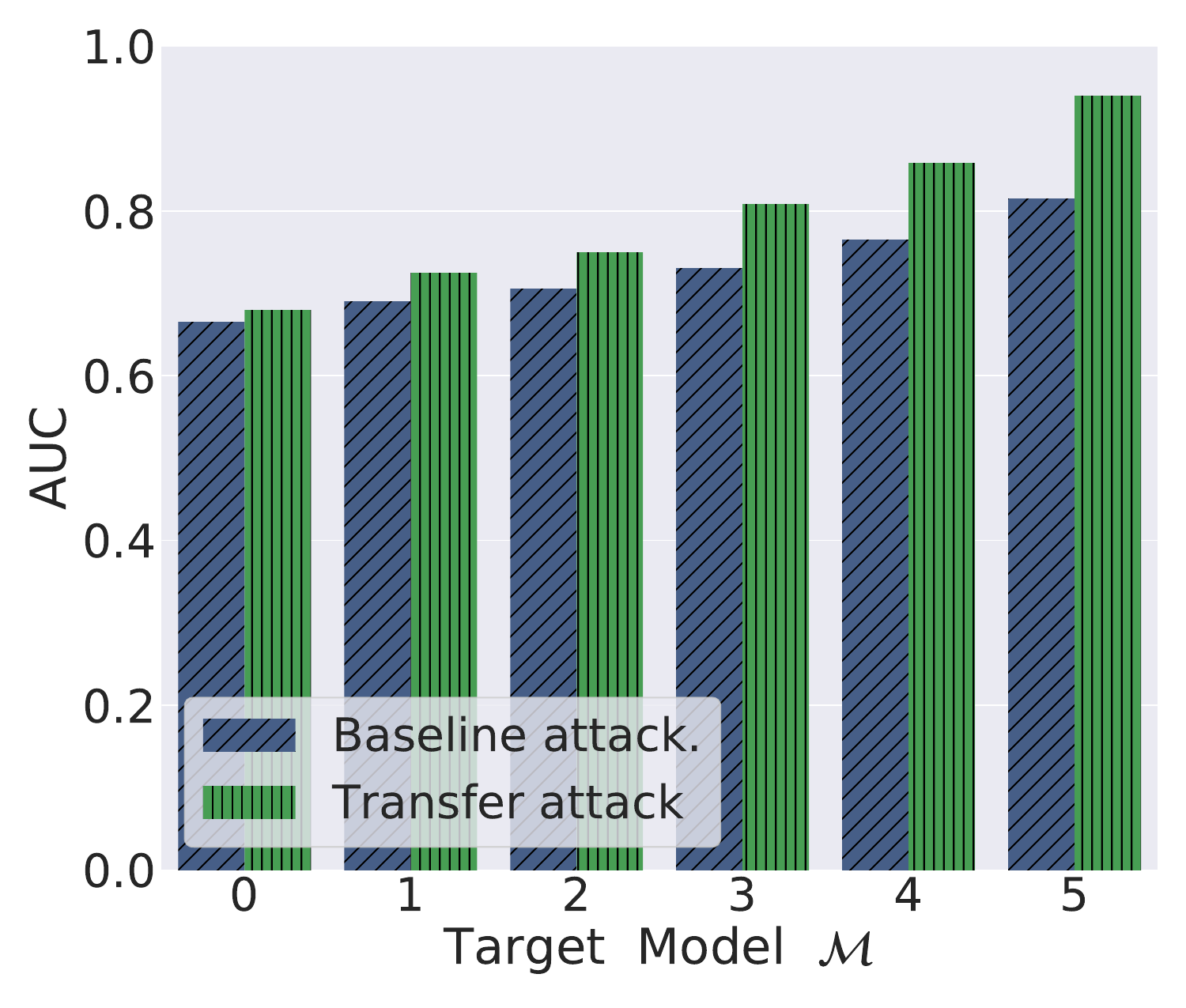}
\caption{CIFAR-10}
\label{fig:compareCIFAR10}
\end{subfigure}
\begin{subfigure}{0.5\columnwidth}
\includegraphics[width=\columnwidth]{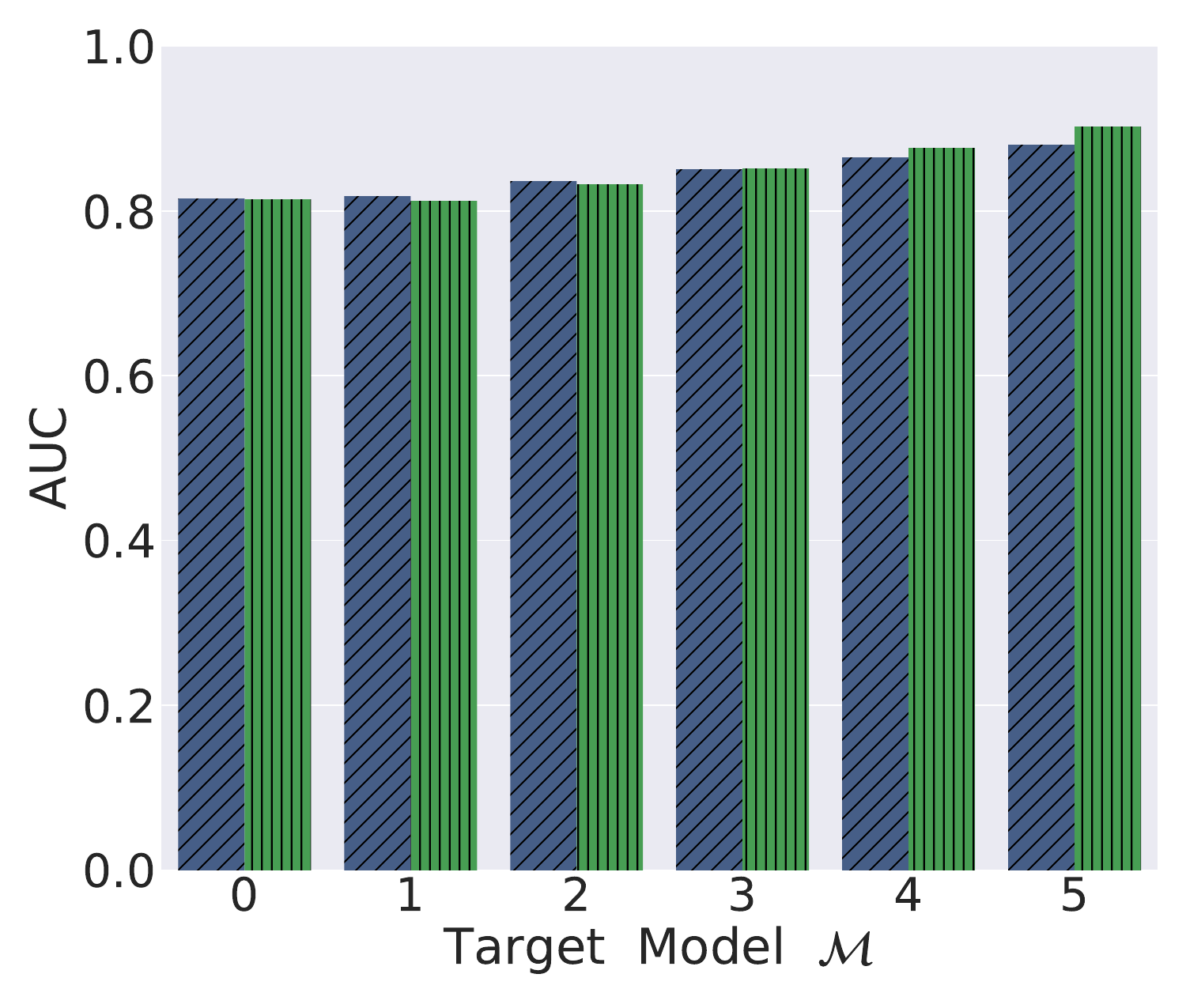}
\caption{CIFAR-100}
\label{fig:compareCIFAR100}
\end{subfigure}
\begin{subfigure}{0.5\columnwidth}
\includegraphics[width=\columnwidth]{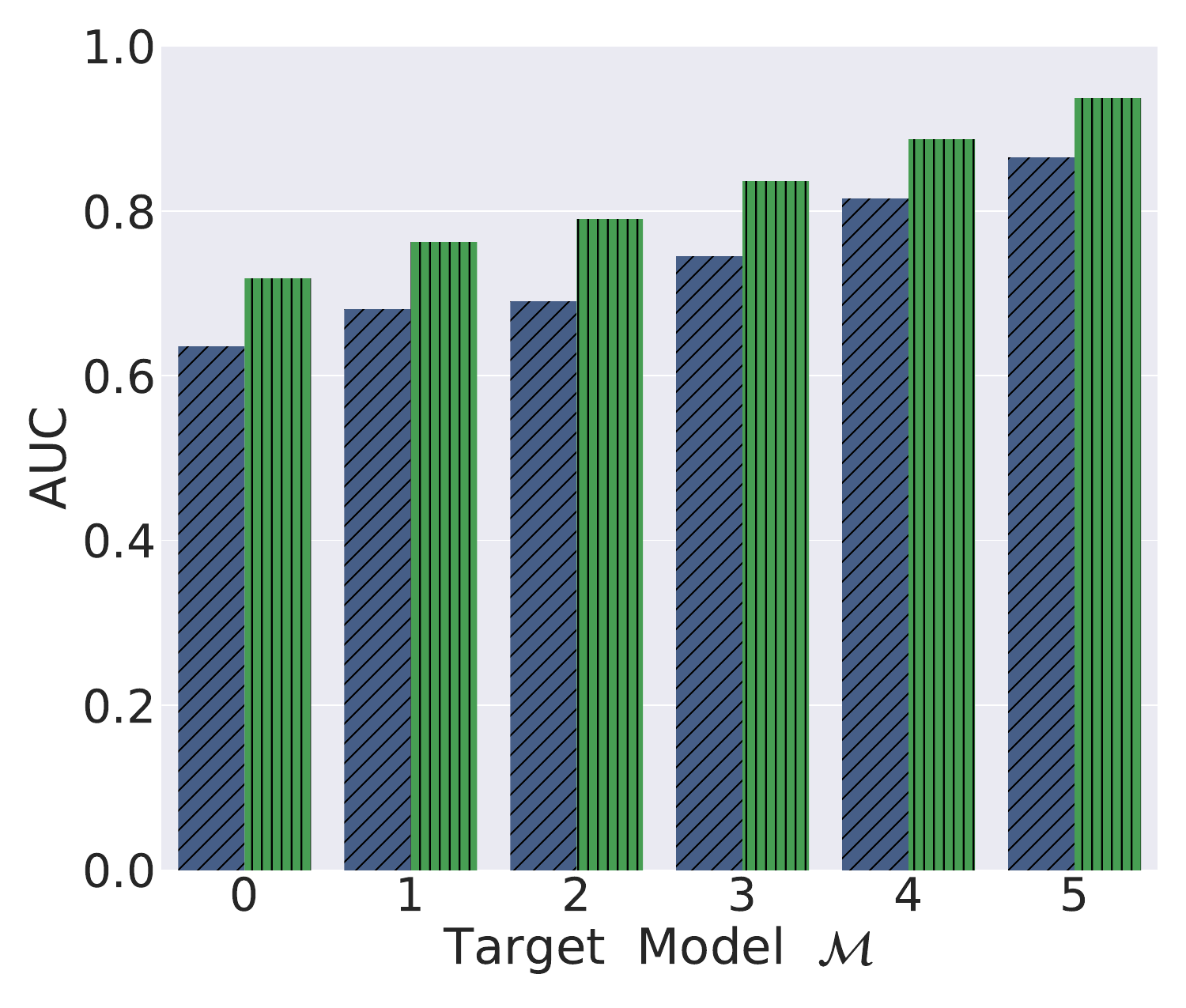}
\caption{GTSRB}
\label{fig:compareGTSRB}
\end{subfigure}
\begin{subfigure}{0.5\columnwidth}
\includegraphics[width=\columnwidth]{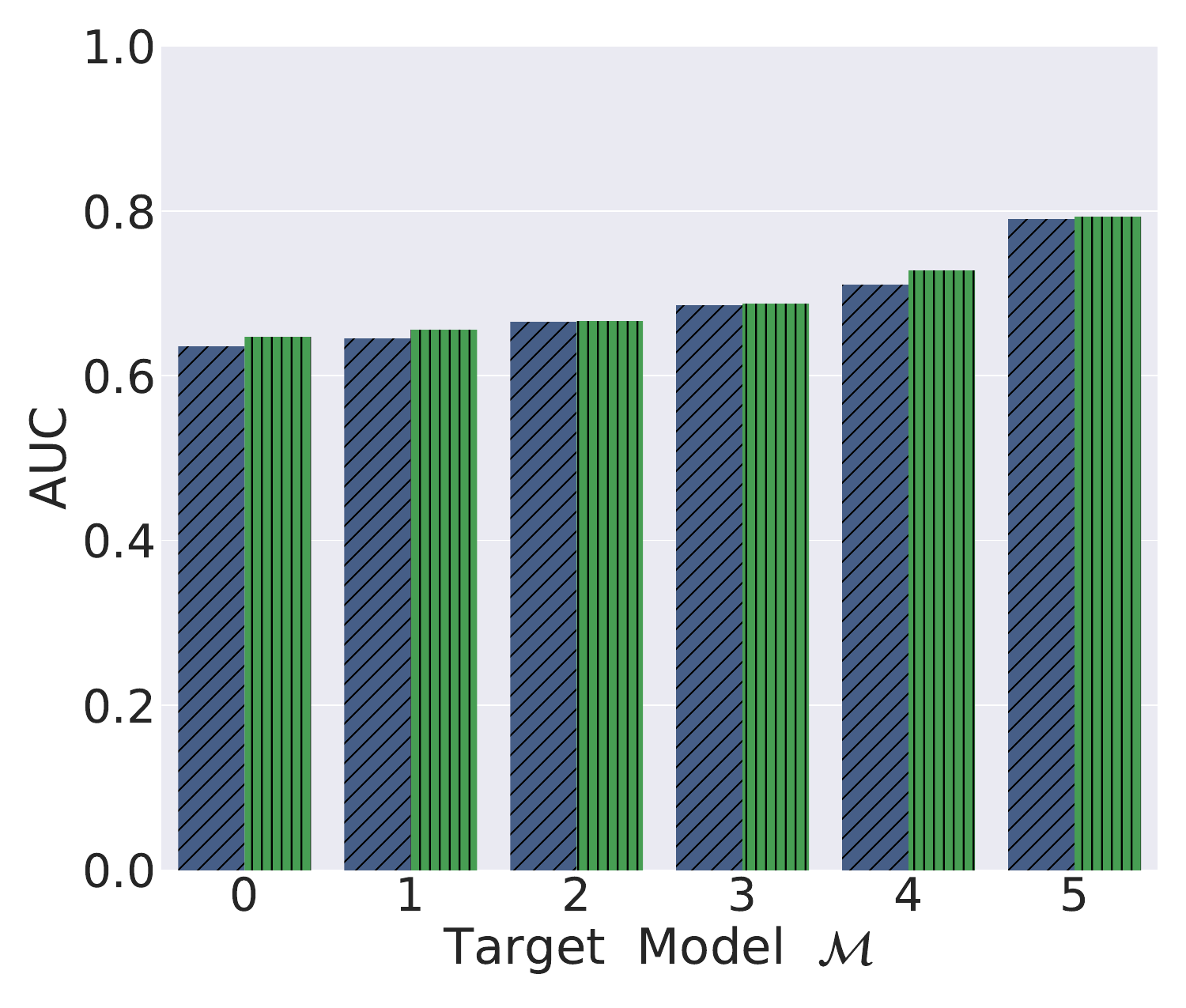}
\caption{Face}
\label{fig:compareFace}
\end{subfigure}
\caption{Comparison of our transfer attack performance with the baseline attack by Yeom et al.~\cite{YGFJ18}. 
The x-axis represents the target model being attacked and the y-axis represents the AUC score.}
\label{fig:adv1_auc}
\end{figure*} 

In this section, we present the first type of decision-based attacks, i.e., transfer attack.
We start by introducing our key intuition. 
Then, we describe the attack methodology. 
Finally, we present the evaluation results.

% ======================================================
\subsection{Key Intuition}
% ======================================================

The intuition of this attack is that the transferability property holds between shadow model $\shadow$ and target model $\model$. 
Almost all related works~\cite{PMG16,DMPJBONR19,LCLS16,NKKKP19} focus on the transferability of adversarial examples, i.e., adversarial examples can transfer between models trained for the same task. 
Unlike these works, we focus on the transferability of membership information for benign data samples, i.e., the member and non-member data samples behaving differently in $\model$ will also behave differently in $\shadow$. 
Then we can leverage the shadow model to launch a score-based membership inference attack.

% ======================================================
\subsection{Methodology}
% ======================================================

The transfer attack’s methodology can be divided into four stages, namely shadow dataset relabeling, shadow model architecture selection, shadow model training, and membership inference. 
The algorithm can be found in Appendix \autoref{alg:transfer}.

\mypara{Shadow Dataset Relabeling}
As aforementioned, the adversary has a shadow dataset $\shadowData$ drawn from the same distribution as the target model $\model$'s dataset $\targetData$. 
To train a shadow model, the first step is to relabel these data samples using the target model $\model$ as an oracle.
In this way, the adversary can establish a connection between the shadow dataset and the target model, which facilitates the shadow model to be more similar to the target model in the next step.

\mypara{Shadow Model Architecture Selection}
As the adversary knows the main task of the target model, it can build the shadow model using high-level knowledge of the classification task (e.g., convolutional networks are appropriate for vision). 
As in prior score-based attacks, we also use the same architecture of target models to build our shadow models. 
Note that we emphasize that the adversary does not have the knowledge of the concrete architecture of the target model, and in \autoref{adv1_results}, we also show that a wide range of architecture choices yield similar attack performance.

\mypara{Shadow Model Training}
The adversary trains the shadow model $\shadow$ with the relabeled shadow dataset $\shadowData$ in conjunction with classical training techniques.

\mypara{Membership Inference}
Finally, the adversary feeds a candidate data sample into the shadow model $\shadow$ to calculate its cross-entropy loss with the ground truth label.
\begin{align}
\mathrm{CELoss}=-\sum_{i=0}^{K} \mathbf{1}_{y} \log (p_{i}),
\label{equ:CELoss}
\end{align}
where $\mathbf{1}_{y}$ is the one-hot encoding of the ground truth label $y$, $p_{i}$ is the probability that the candidate sample belongs to class $i$, and $K$ is the number of classes. 
If the loss value is smaller than a threshold, the adversary then determines the sample being a member and vice versa. 
The adversary can pick a suitable threshold depending on their requirements, as in many machine learning applications.~\cite{BHPZ17,PTC18,HZHBTWB19,FLJLPR14,SZHBFB19,JSBZG19}. 
In our evaluation, we mainly use area under the ROC curve (AUC) which is threshold independent as our evaluation metric.

% ======================================================
\subsection{Experimental Setup}
\label{adv1_setup}
% ======================================================

Following the attack strategy, we split each dataset into $\targetData$ and $\shadowData$: One is used to train and test the target model, and the other is used to train the shadow model $\shadow$ after relabeled by the target model. 
For evaluation, $\targetData$ is also split into two: One is used to train the target model $\model$, i.e., $\train$, and serves as the member samples of the target model, while the other $\test$ serves as the non-member samples. 

It is well known that the inherent overfitting drives ML models to be vulnerable to membership leakage~\cite{SSSS17,SZHBFB19}.
To show the variation of the attack performance on each dataset, we train 6 target models $\model$-0, $\model$-1, ..., $\model$-5 using different size of the training set $\train$, exactly as performed in the prior work by Shokri et al.~\cite{SSSS17} and many subsequent works~\cite{TLGYW18,SZHBFB19,SSM19,LBG17}. 
The sizes of $\train$, $\test$, and $\shadowData$ are summarized in Appendix \autoref{table:datasetsplity}. 

We execute the evaluation on randomly reshuffled data samples from $\targetData$, and select sets of the same size (i.e, equal number of members and non-members) to maximize the uncertainty of inference, thus the baseline performance is equivalent to random guessing. 
We adopt AUC as our evaluation metric which is threshold independent. 
In addition, we further discuss methods to pick threshold for our attack later in this section.

\begin{figure}[!t]
\centering
\includegraphics[width=0.75\columnwidth]{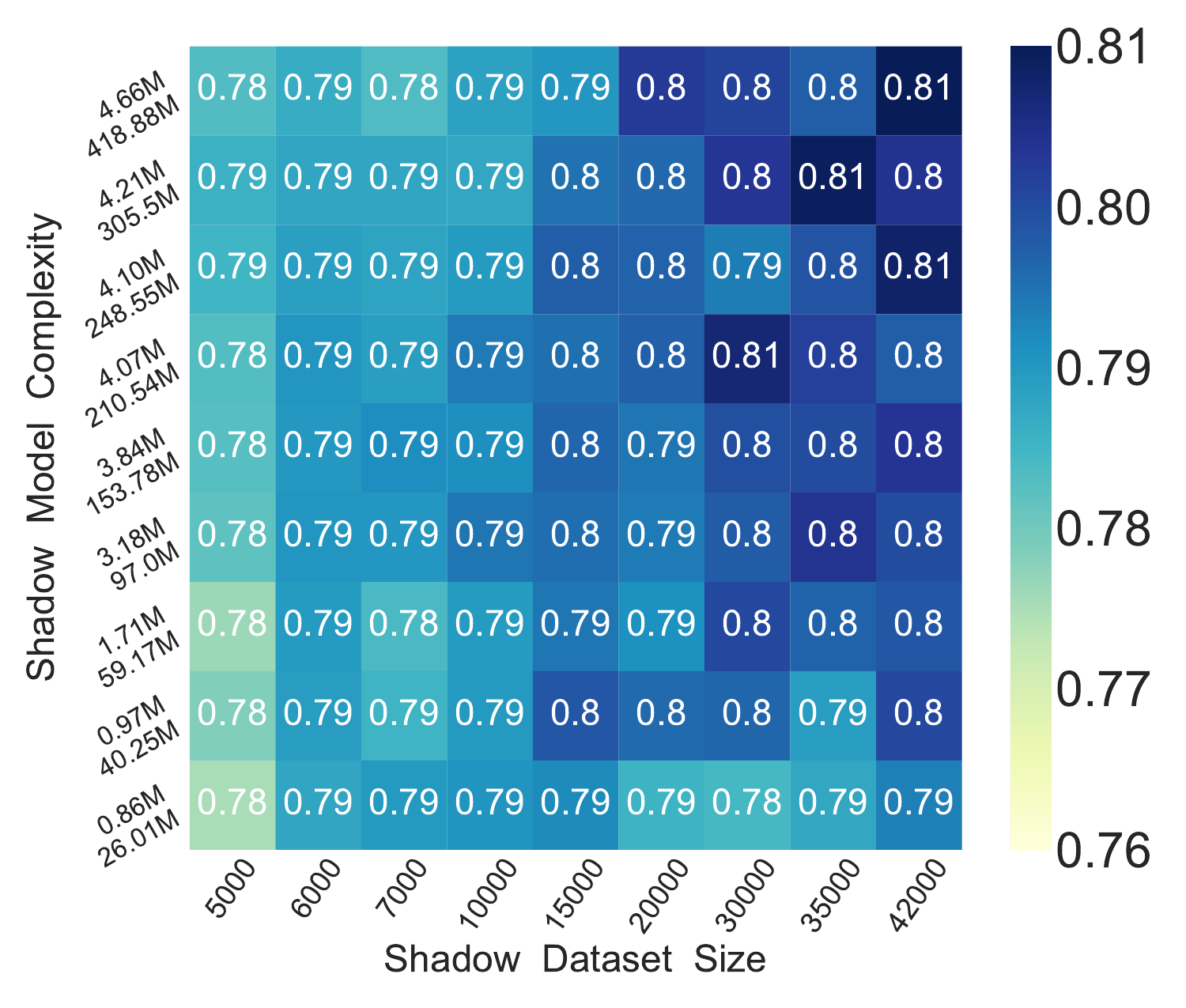}
\caption{Attack AUC under the effect of changing the dataset size  and shadow model complexity (upper is the number of parameters, lower is the computational complexity FLOPs). 
The target model ($\model$-0, CIFAR-100)'s training set size is 7,000, and complexity is 3.84M parameters and 153.78M FLOPs.}
\label{fig:heatmap}
\end{figure}

\begin{figure*}[!t]
\centering
\begin{subfigure}{0.49\columnwidth}
\includegraphics[width=\linewidth]{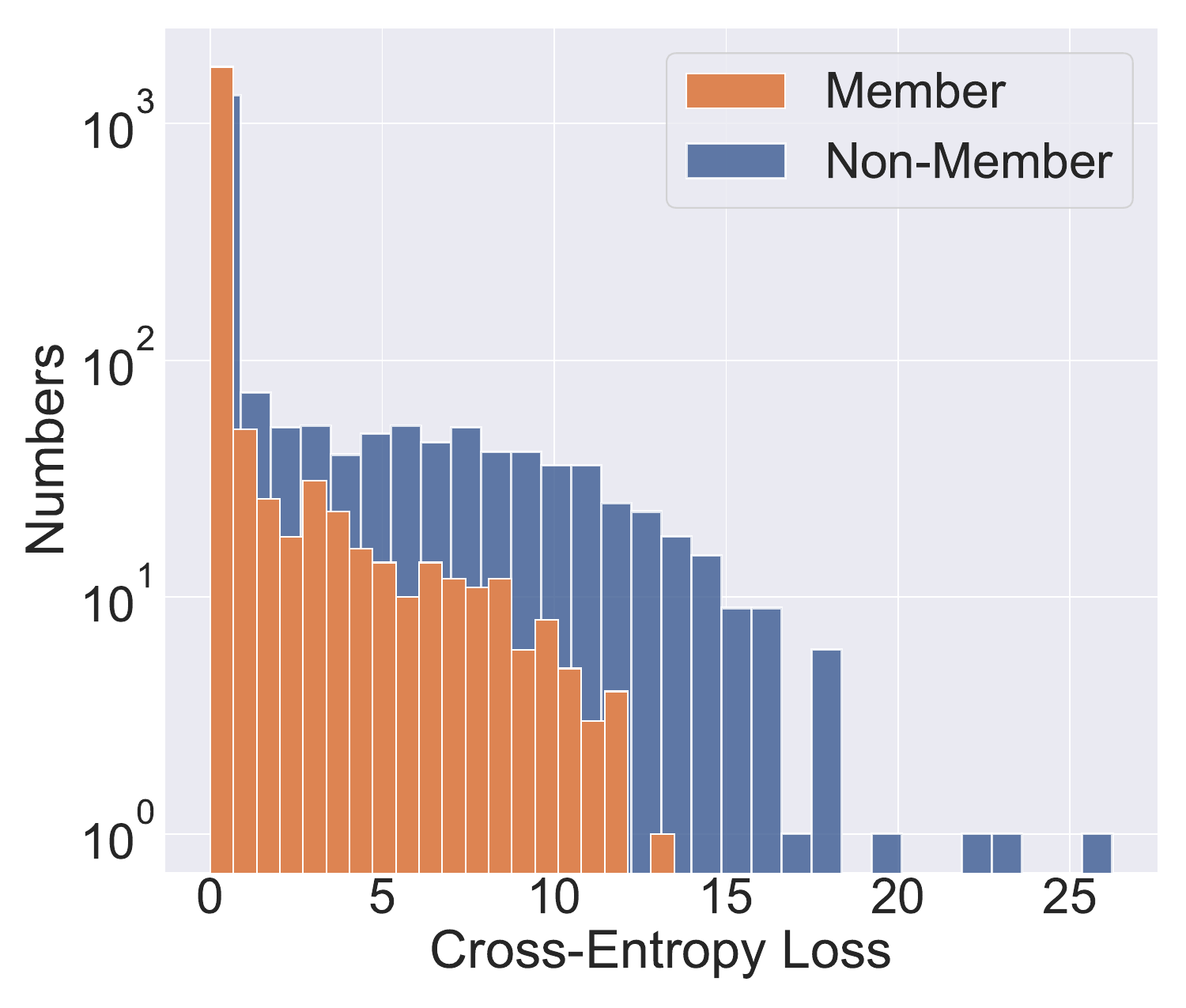}
\caption{CIFAR-10, $\model$-0}
\label{fig:loss-cifar10-0} 
\end{subfigure}
\begin{subfigure}{0.49\columnwidth}
\includegraphics[width=\linewidth]{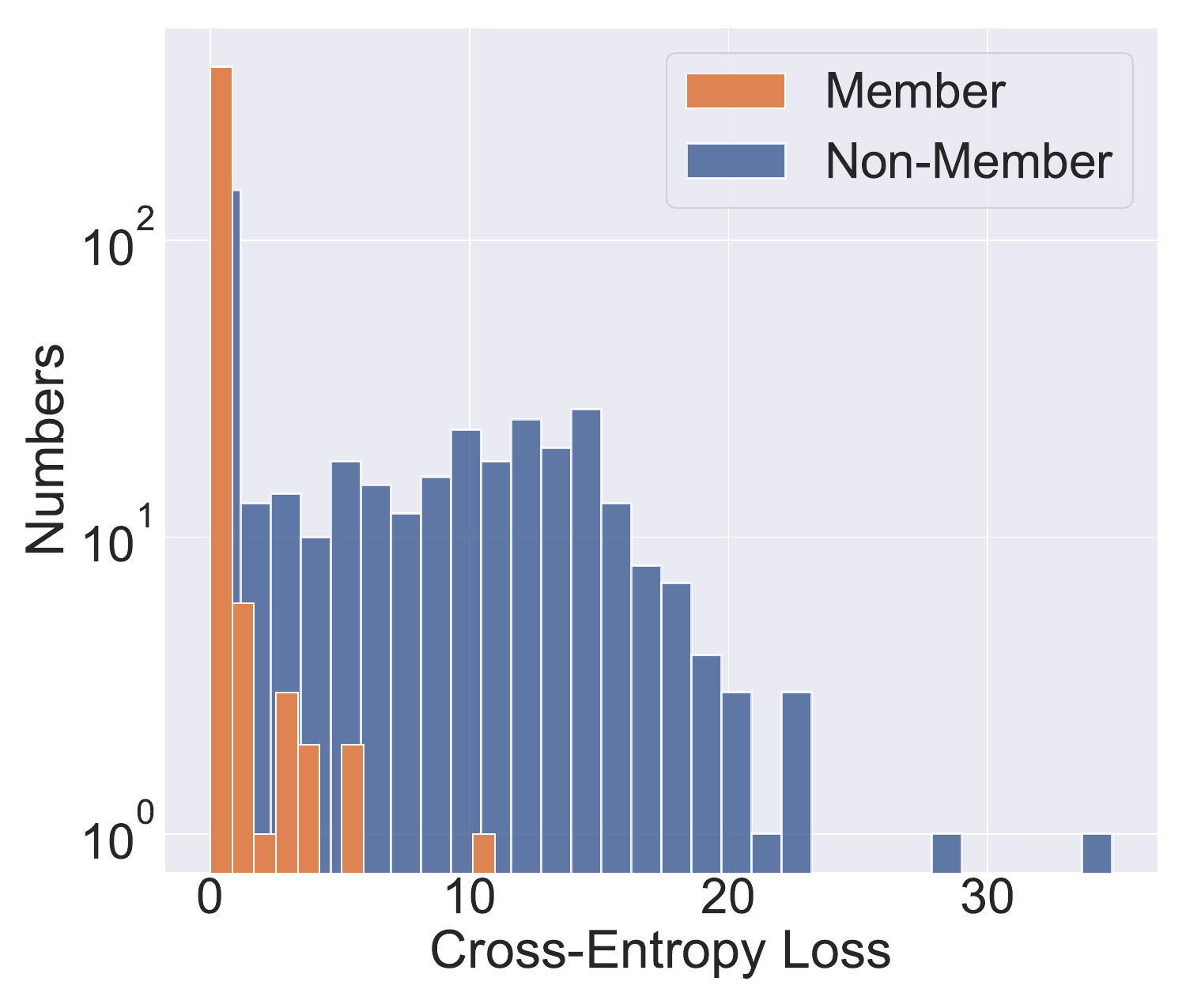}
\caption{CIFAR-10, $\model$-5}
\label{fig:loss-cifar10-5}
\end{subfigure}
\begin{subfigure}{0.49\columnwidth}
\includegraphics[width=\linewidth]{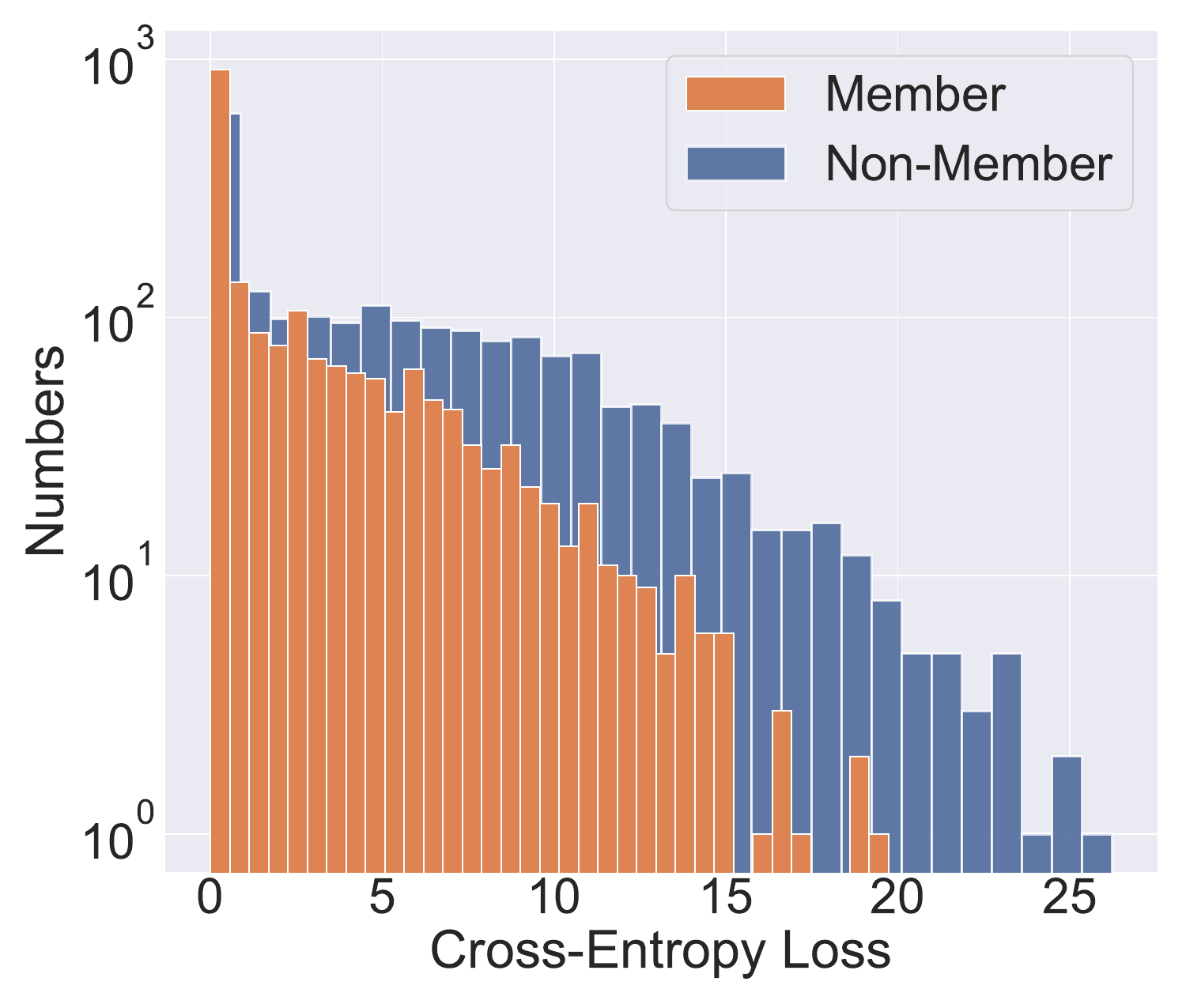}
\caption{CIFAR-100, $\model$-0}
\label{fig:loss-cifar100-0} 
\end{subfigure}
\begin{subfigure}{0.49\columnwidth}
\includegraphics[width=\linewidth]{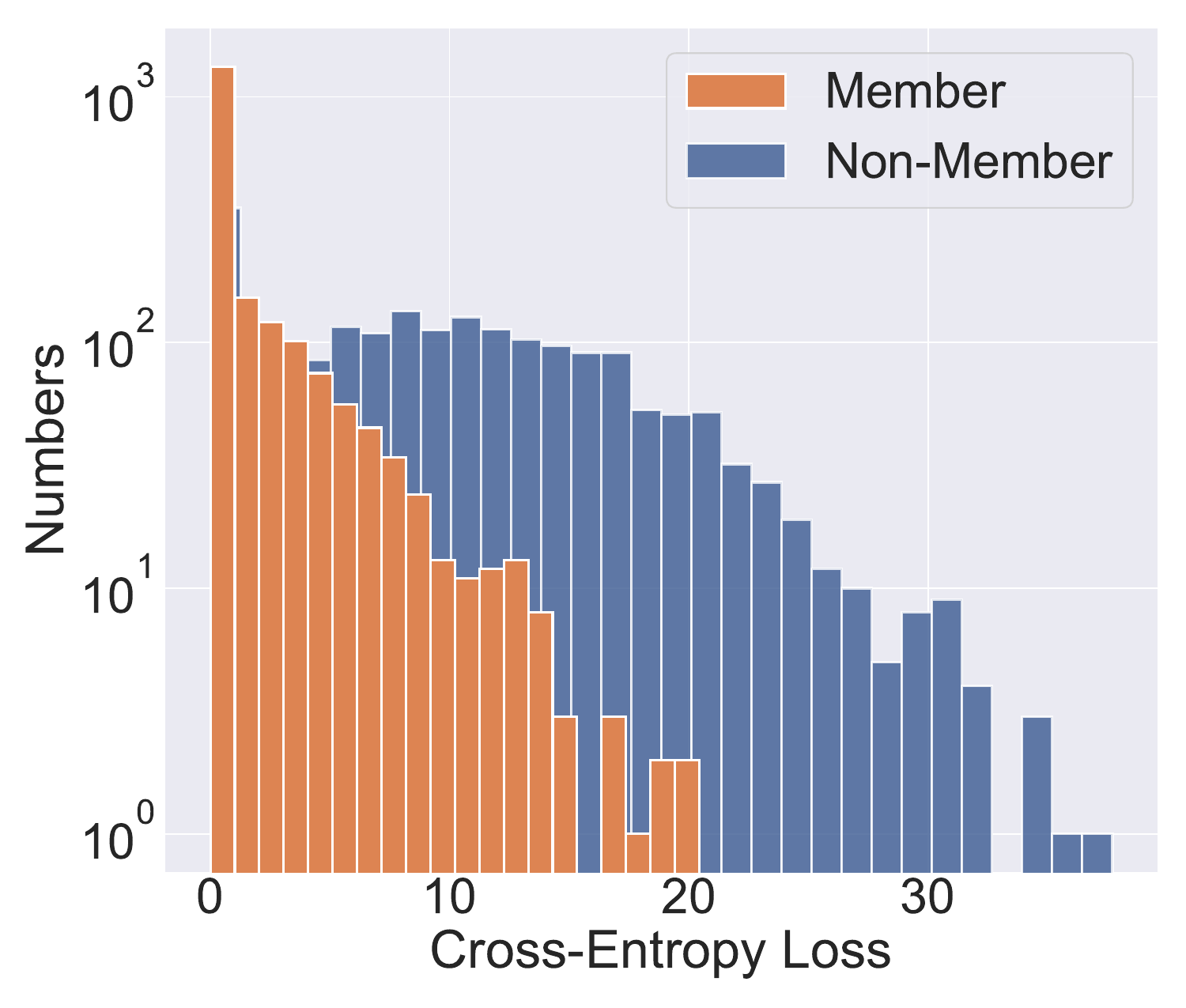}
\caption{CIFAR-100, $\model$-5}
\label{fig:loss-cifar100-5}
\end{subfigure}
\caption{The cross entropy loss distribution obtained from the shadow model. 
The x-axis represents the loss value and the y-axis represents the number of the loss.}
\label{fig:loss_distribution}
\end{figure*}

\begin{figure}[!t]
\centering
\begin{subfigure}{0.49\columnwidth}
\includegraphics[width=\linewidth]{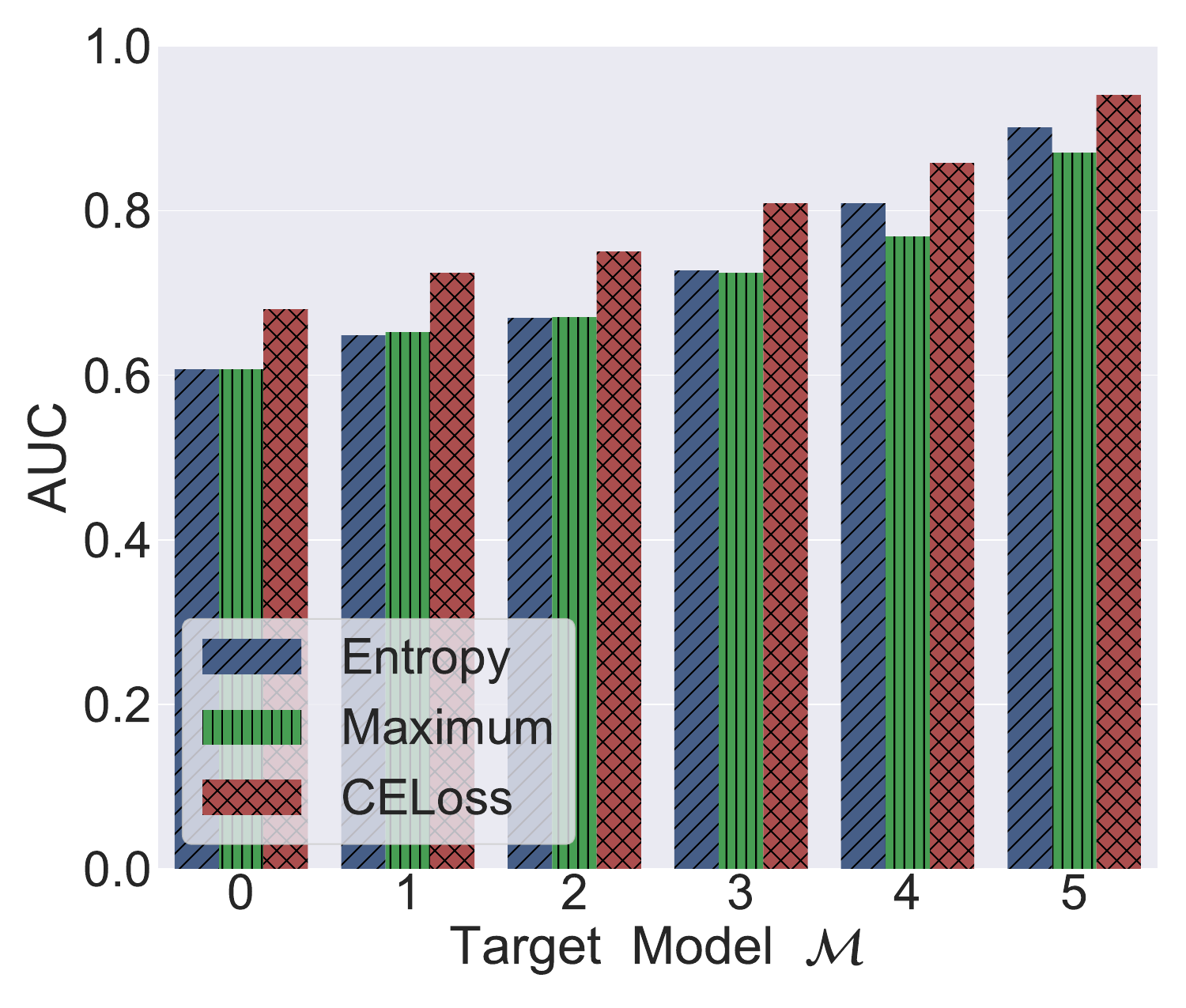}
\caption{CIFAR-10}
\end{subfigure}
\begin{subfigure}{0.49\columnwidth}
\includegraphics[width=\linewidth]{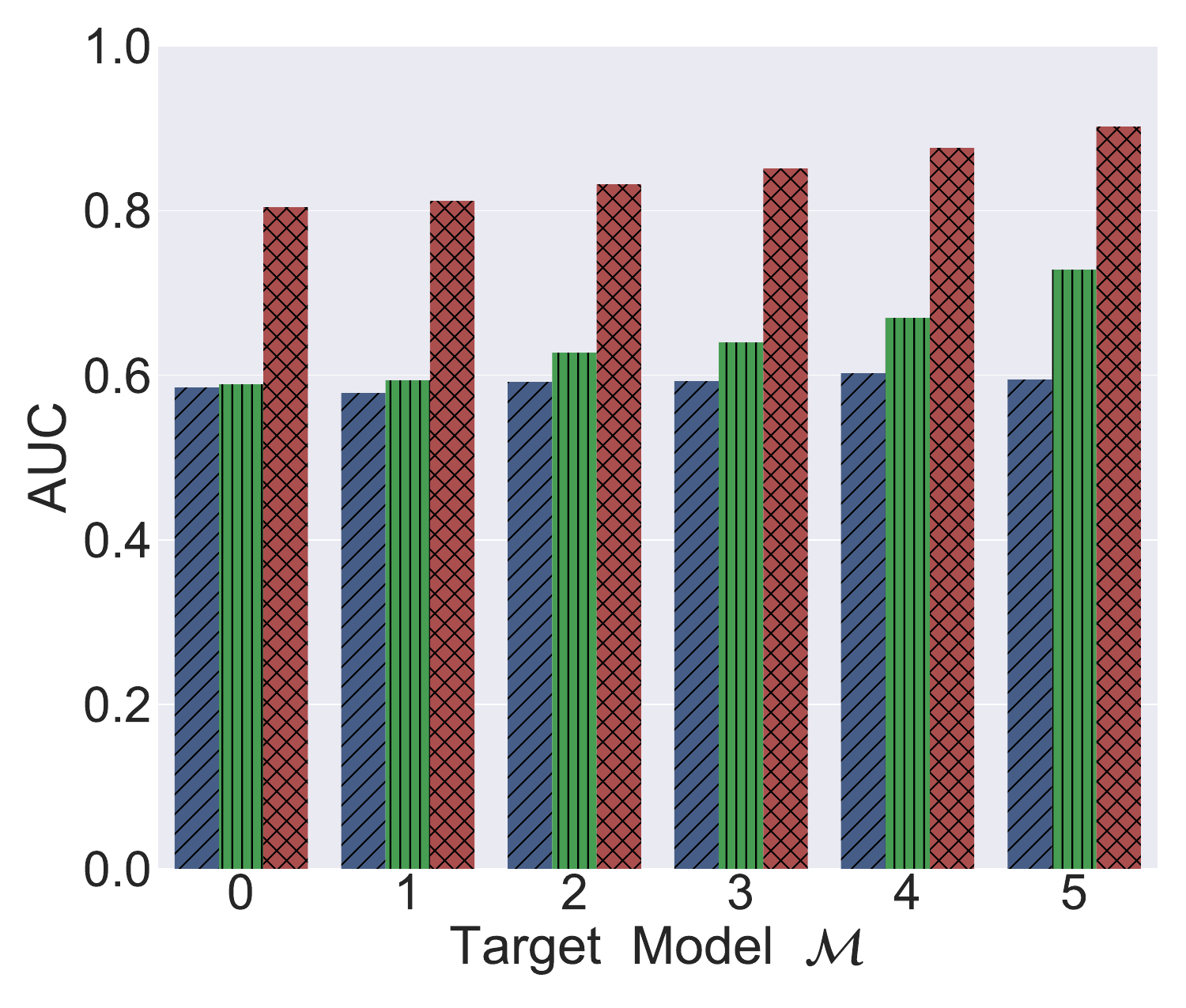}
\caption{CIFAR-100}
\end{subfigure}
\caption{Attack AUC for three different statistical measures. 
The x-axis represents the target model being attacked and the y-axis represents the AUC score.}
\label{fig:max_entropy}
\end{figure}

% ======================================================
\subsection{Results}
\label{adv1_results}
% ======================================================

\mypara{Attack AUC Performance}
\autoref{fig:adv1_auc} depicts the performance of our transfer attack and baseline attack. 
First, we can observe that our transfer attack performs at least on-par with the baseline attack. 
More encouragingly, on the CIFAR-10 and GTSRB datasets, our transfer attack achieves better performance than the baseline attack.
For example, in \autoref{fig:adv1_auc} ($\model$-5, CIFAR-10), the AUC score of the transfer attack is 0.94, while that of the baseline attack is 0.815. 
The reason why our transfer attack outperforms the baseline attack on CIFAR-10 and GTSRB rather than on CIFAR-100 and Face, is that the size of the shadow dataset for the first two datasets is relatively larger than that of the latter two, compared to the size of each dataset (see Appendix \autoref{table:datasetsplity}). 
In the next experiments, we make the same observation that a larger shadow dataset implies better attack performance. 

\mypara{Effects of the Shadow Dataset and Model}
We further investigate the effects of shadow dataset size and shadow model complexity (structure and hyper-parameter) on the attack performance.
More concretely, for the target model ($\model$-0, CIFAR-100), we vary the size of the shadow dataset $\shadowData$ from 5,000 to 42,000, where the target training set $\train$ is 7,000. 
We also vary the complexity of the shadow model from 0.86M (number of parameters) and 26.01M (FLOPs,\footnote{FLOPs represent the theoretical amount of floating-point arithmetic needed when feeding a sample into the model.} computational complexity) to 4.86M and 418.88M, where the complexity of the target model is 3.84M and 153.78M, respectively. 
We conduct extensive experiments to simultaneously tune these two hyper-parameters and report the results in~\autoref{fig:heatmap}. 
Through investigation, we make the following observations.
\begin{itemize}
\item  Larger shadow dataset implies more queries to the target model which leads to better attack performance.
\item Even simpler shadow models and fewer shadow datasets (bottom left part) can achieve strong attack performance.
\item In general, the transfer attack is robust even if the shadow model is much different from the target model.
\end{itemize}

\mypara{Effects of Statistical Metrics} 
As prior works~\cite{SSSS17,SZHBFB19} also use other statistical metrics, i.e., maximum confidence scores $Max(p_{i})$ and normalized entropy $\frac{-1}{\log (K)} \sum_{i} p_{i} \log \left(p_{i}\right)$. 
Here, we also conduct experiments with these statistical metrics. 
\autoref{fig:max_entropy} reports the AUC on the CIFAR-10 and CIFAR-100 datasets. 
We can observe that the loss metric achieves the highest performance with respect to the different target models. 
Meanwhile, the AUC score is very close between maximum confidence score and entropy. 
This indicates that the loss metric contains the strongest signal on differentiating member and non-member samples.
We will give an in-depth discussion on this in \autoref{subsec:QualitativeAnalysis}.

\mypara{Loss Distribution of Membership}
To explain why our transfer attack works, \autoref{fig:loss_distribution} further shows the loss distribution of member and non-member samples from the target model calculated on the shadow model ($\model$-0 and $\model$-5 on CIFAR-10 and CIFAR-100).
Though both member and non-member samples are never used to train the shadow model, we still observe a clear difference between their loss distribution.
This verifies our key intuition aforementioned: The transferability of membership information holds between shadow model $\shadow$ and target model $\model$, i.e., the member and non-member samples behaving differently in $\model$ will also behave differently in $\shadow$.

\mypara{Threshold Choosing}
As mentioned before, in the membership inference stage, the adversary needs to make a  manual decision on which threshold to use.
For the transfer attack, since we assume that the adversary has a dataset that comes from the same distribution as the target  model’s dataset, it can rely on the shadow dataset to estimate a threshold by sampling certain part of that dataset as non-member samples.

% ======================================================
\section{Boundary-Attack}
\label{sec:adv2}
% ======================================================

\begin{table*}[!t]
\definecolor{mygray}{gray}{0.9}
\centering
\caption{
The cross entropy between the confidence scores and other labels except for the predicted label. 
ACE represent the Average Cross Entropy.}
\label{table:adv2intuition}
\scalebox{0.83}
{
\begin{tabular}{l|c|c|ccccccccccc }
\toprule
&Truth& Predicted& \multicolumn{10}{c}{Cross Entropy} \\
Status  & Label& Label& 0 & 1 & 2 & 3 & 4 & 5 & 6 & 7 & 8 & 9 & ACE\\
\midrule
\rowcolor{mygray}
(a) Member    & 6 & 6  & 7.8156& 8.3803& 4.1979& 1.0942& 4.1367& 4.3492& -& 7.6328& 1.5522& 1.2923 & \textbf{4.4946}\\
(b) Non-member & 8 & 8 & 2.3274& 0.8761& 0.8239& 2.0793& 1.2275& 0.9791& 1.2373& 1.1152& -& 5.0451 & 1.2218\\
\cmidrule(lr){1-14}
\rowcolor{mygray}
(c) Member     &3 &3 & 1.2995& 5.2842& 5.4212& -& 1.5130& 4.8059& 4.5897& 7.1547& 3.2411& 4.7910& \textbf{4.2334}\\
(d) Non-member &7 &9 & 2.8686& 1.8325& 3.6480& 0.5352& 1.8722& 1.1689& 4.0124& 0.6866& 3.1071& -& 2.1766\\
\bottomrule
\end{tabular}
}
\end{table*}

After demonstrating our transfer attack, we now present our second attack, i.e., boundary attack. 
Since curating auxiliary data requires significant time and monetary investment. Thus, we relax this assumption in this attack. 
The adversary does not have a shadow dataset to train a shadow model. 
All they could rely on is the predicted label from the target model.
To the best of our knowledge, this is by far the most strict setting for membership inference against ML models. 
In the following section, we start with the key intuition description. 
Then, we introduce the attack methodology. 
In the end, we present the evaluation results.

% ======================================================
\subsection{Key Intuition}
\label{subsec:keyintuition}
% ======================================================

Our intuition behind this attack follows a general observation of the overfitting nature of ML models.
Concretely, an ML model is more confident in predicting data samples that it is trained on. 
In contrast to the prior score-based attacks\cite{SSSS17,LBG17,SZHBFB19,YGFJ18,SSM19,HYYBGC21,LLR21} that directly exploit confidence scores as analysis objects, we place our focus on the antithesis of this observation, i.e., since the ML model is more confident on member data samples, it should be much harder to change its mind.

\begin{figure}[!t]
\centering
\begin{subfigure}{0.49\columnwidth}
\includegraphics[width=\linewidth]{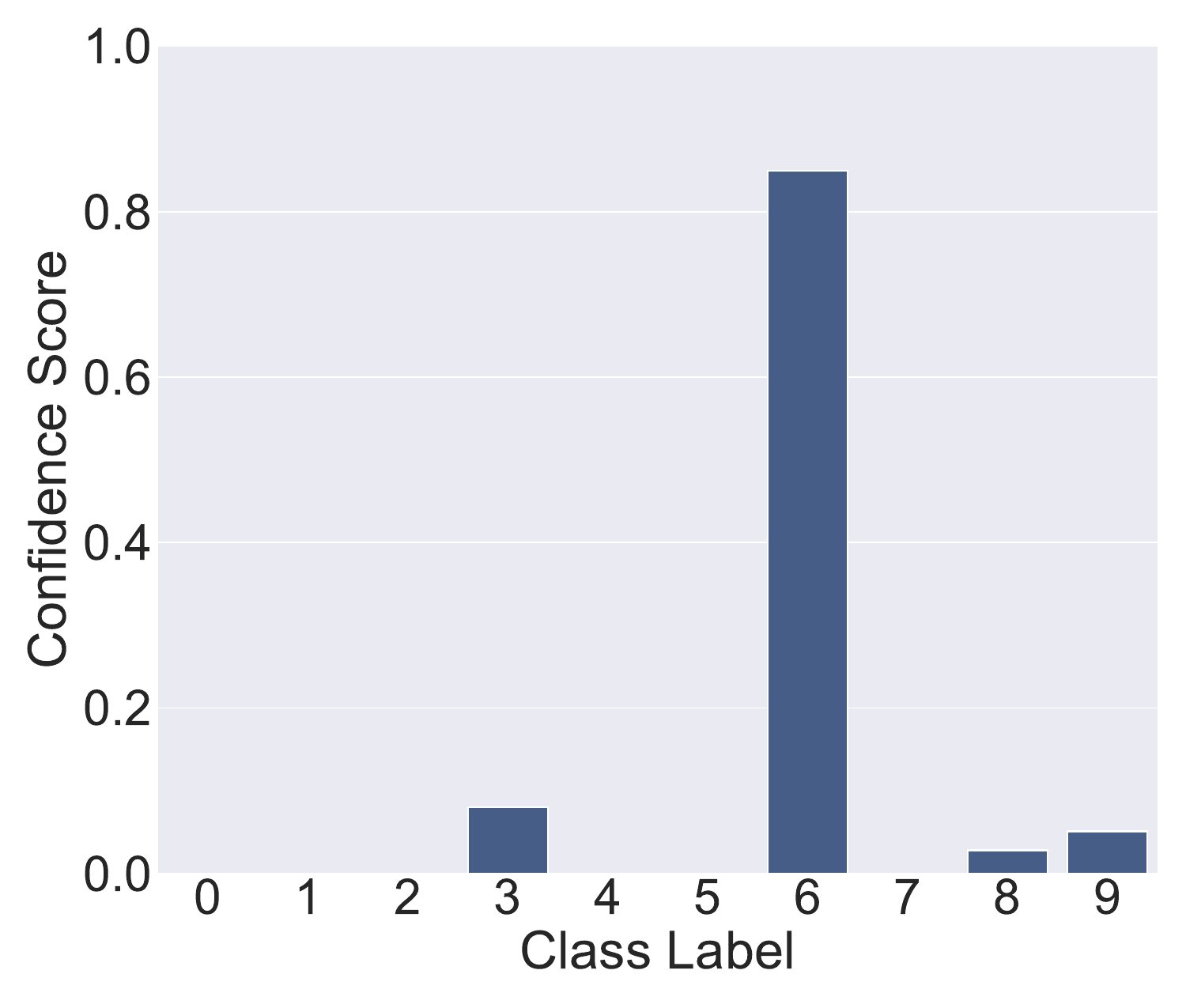}
\caption{Member data sample}
\label{fig:post-a} 
\end{subfigure}
\begin{subfigure}{0.49\columnwidth}
\includegraphics[width=\linewidth]{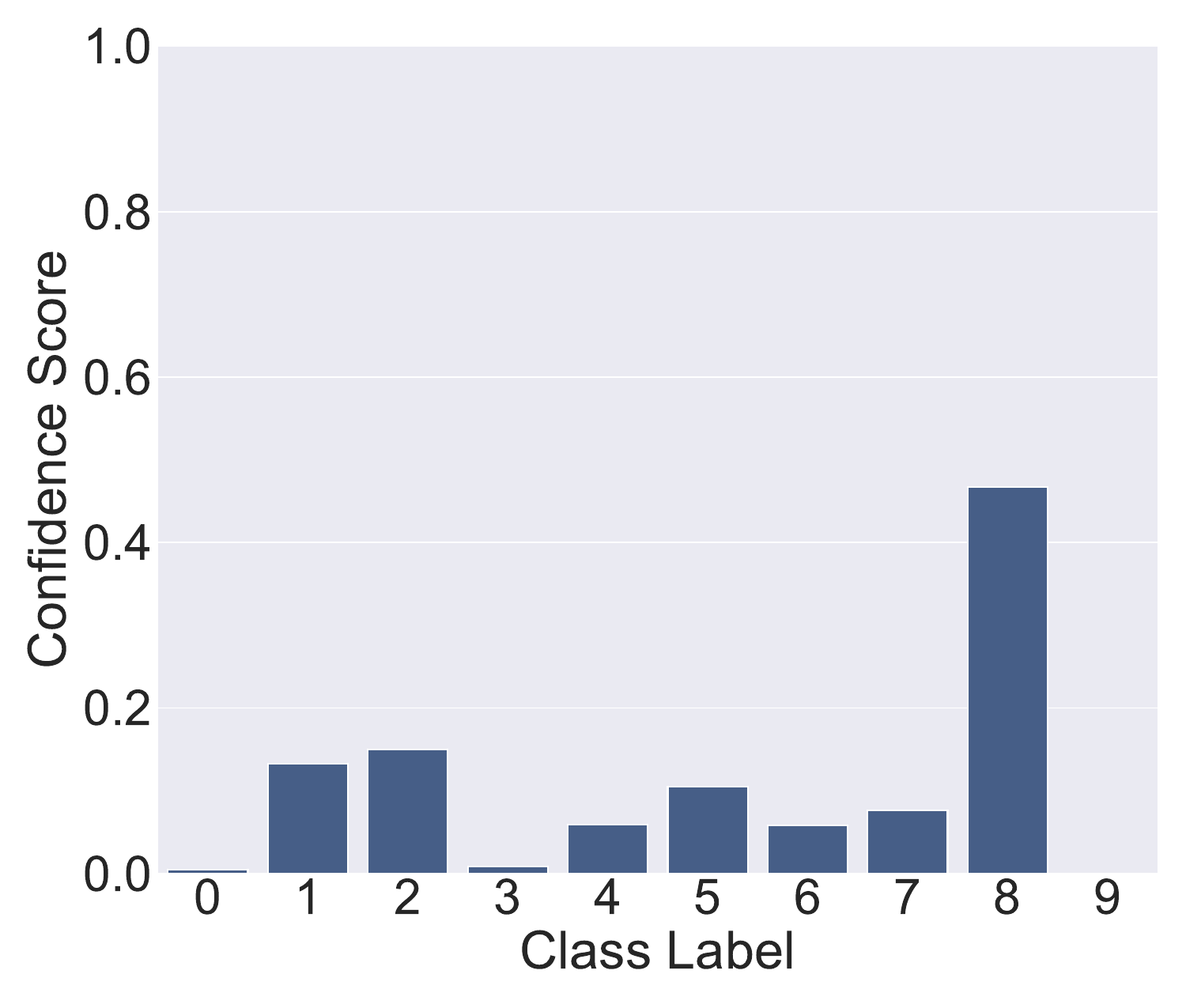}
\caption{Non-member data sample}
\label{fig:post-b}
\end{subfigure}
\begin{subfigure}{0.49\columnwidth}
\includegraphics[width=\linewidth]{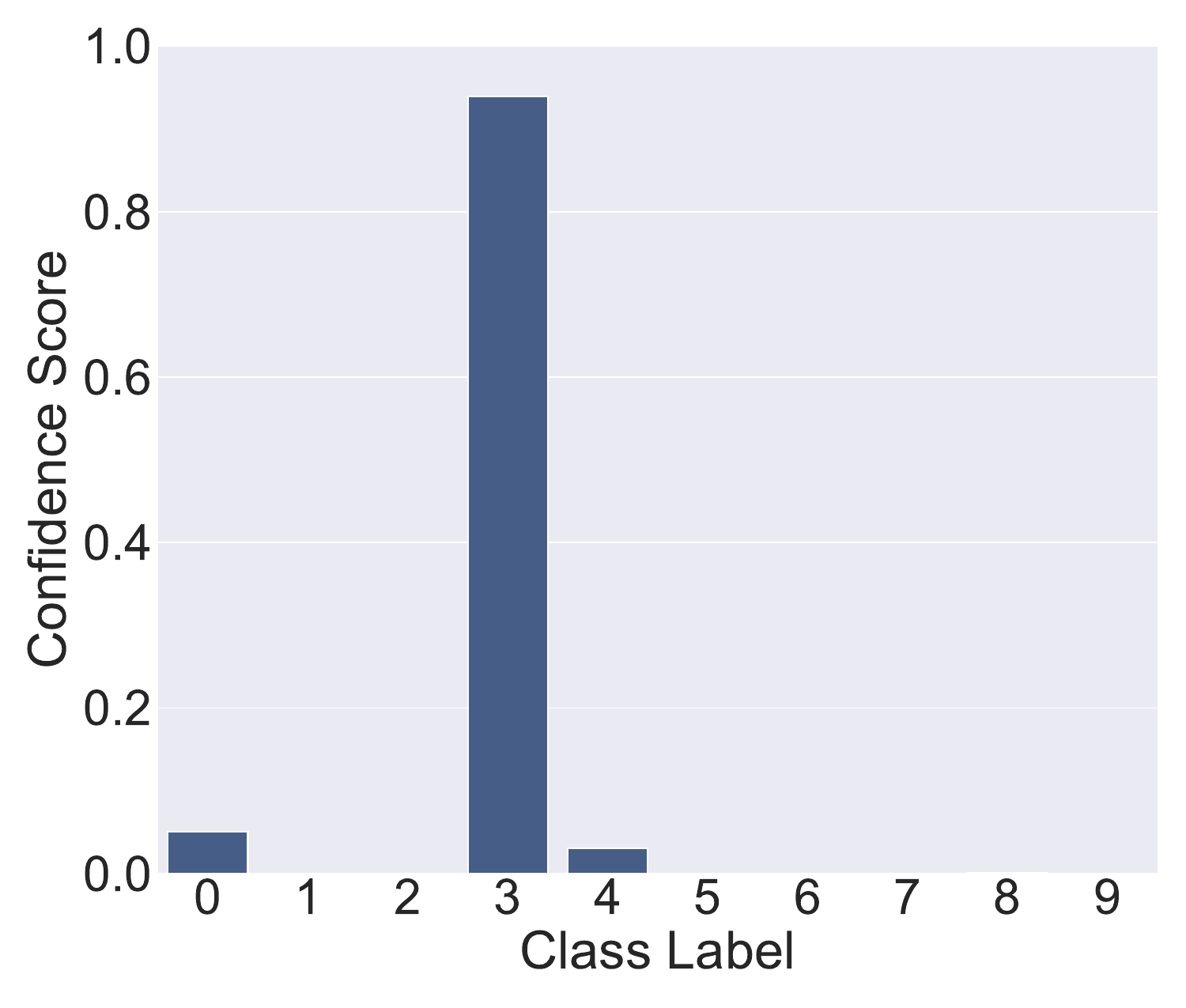}
\caption{Member data sample}
\label{fig:post-c} 
\end{subfigure}
\begin{subfigure}{0.49\columnwidth}
\includegraphics[width=\linewidth]{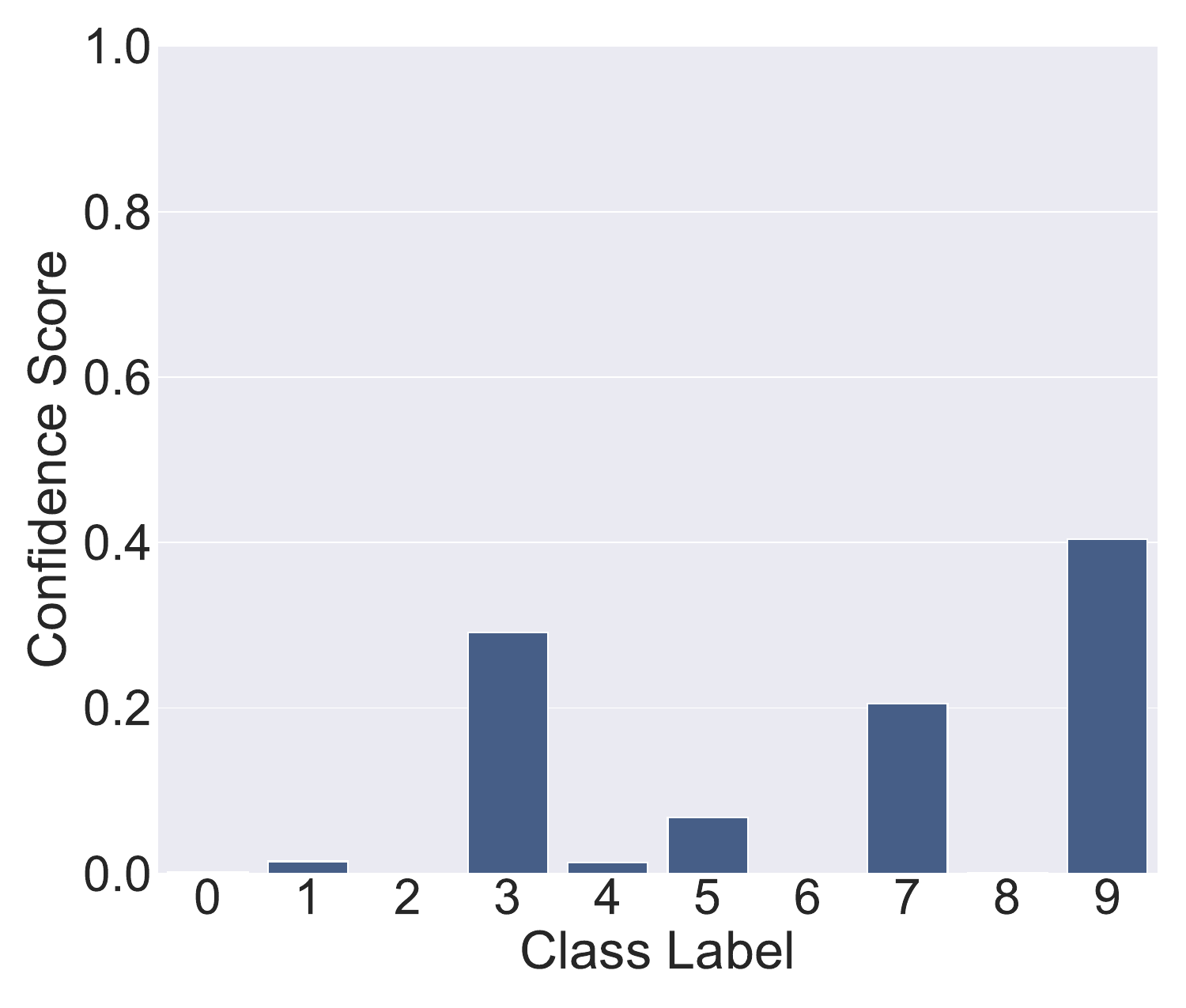}
\caption{Non-member data sample}
\label{fig:post-d}
\end{subfigure}
\caption{The probability distribution of the target model ($\model$-0, CIFAR-10) on member samples and  non-member samples.}
\label{fig:adv2intuition}
\end{figure}

Intuitively, \autoref{fig:adv2intuition} depicts the confidence scores for two randomly selected member data samples (\autoref{fig:post-a}, \autoref{fig:post-c}) and non-member data samples (\autoref{fig:post-b}, \autoref{fig:post-d}) with respect to $\model$-0 trained on CIFAR-10. 
We can observe that the maximal score for member samples is indeed much higher than the one of non-member samples. 
We further use cross entropy (\autoref{equ:CELoss}) to quantify the difficulty for an ML model to change its predicted label for a data sample to other labels.

\autoref{table:adv2intuition} shows the cross entropy between the confidence scores and other labels for these samples. 
We can see that member samples' cross entropy is significantly larger than non-member samples. 
This leads to the following observation on membership information. 

\mypara{Observation}
Given an ML model and a set of data samples, the cost of changing the target model's predicted labels for member samples is larger than the cost for non-member samples.
Furthermore, consider the label-only exposures in a black-box ML model, which means the adversary can only perturb the data samples to change the target model's predicted labels, thus the perturbation needed to change a member sample's predicted label is larger than non-members. 
Then, the adversary can exploit the magnitude of the perturbation to determine whether the sample is a member or not.

% ======================================================
\subsection{Methodology}
\label{subsec:methodology}
% ======================================================

Our attack methodology consists of the following three stages, i.e., decision change, perturbation measurement, and membership inference. 
The algorithm can be found in Appendix \autoref{alg:boundary}.

\mypara{Decision Change}
The goal of changing the final model decision, i.e., predicted label, is similar to that of adversarial attack~\cite{BCMNSLGR13,PMJFCS16,PMGJCS17,CW17,TKPGBM17,SHNSSDG18}, 
For simplicity, we utilize adversarial example techniques to perturb the input to mislead the target model. 
Specifically, we utilize two state-of-the-art black-box adversarial attacks, namely HopSkipJump~\cite{CJW20} and QEBA~\cite{LXZYL20}, which only require access to the model's predicted labels. 

\mypara{Perturbation Measurement}
Once the final model decision has changed, we measure the magnitude of the perturbations added to the candidate input samples. 
In general, adversarial attack techniques typically use $L_p$ distance (or Minkowski Distance), e.g., $L_{0}$, $L_{1}$, $L_{2}$, and $L_{\infty}$, to measure the perceptual similarity between a perturbed sample and its original one. 
Thus, we use $L_p$ distance to measure the perturbation.

\mypara{Membership Inference}
After obtaining the magnitude of the perturbations, the adversary simply considers a candidate sample with perturbations larger than a threshold as a member sample, and vice versa.
Similar to the transfer attack, we mainly use AUC as our evaluation metric. We also provide a general and simple method for choosing a threshold in \autoref{adv2_results}.

\begin{figure*}[!t]
\centering
\begin{subfigure}{0.5\columnwidth}
\includegraphics[width=\columnwidth]{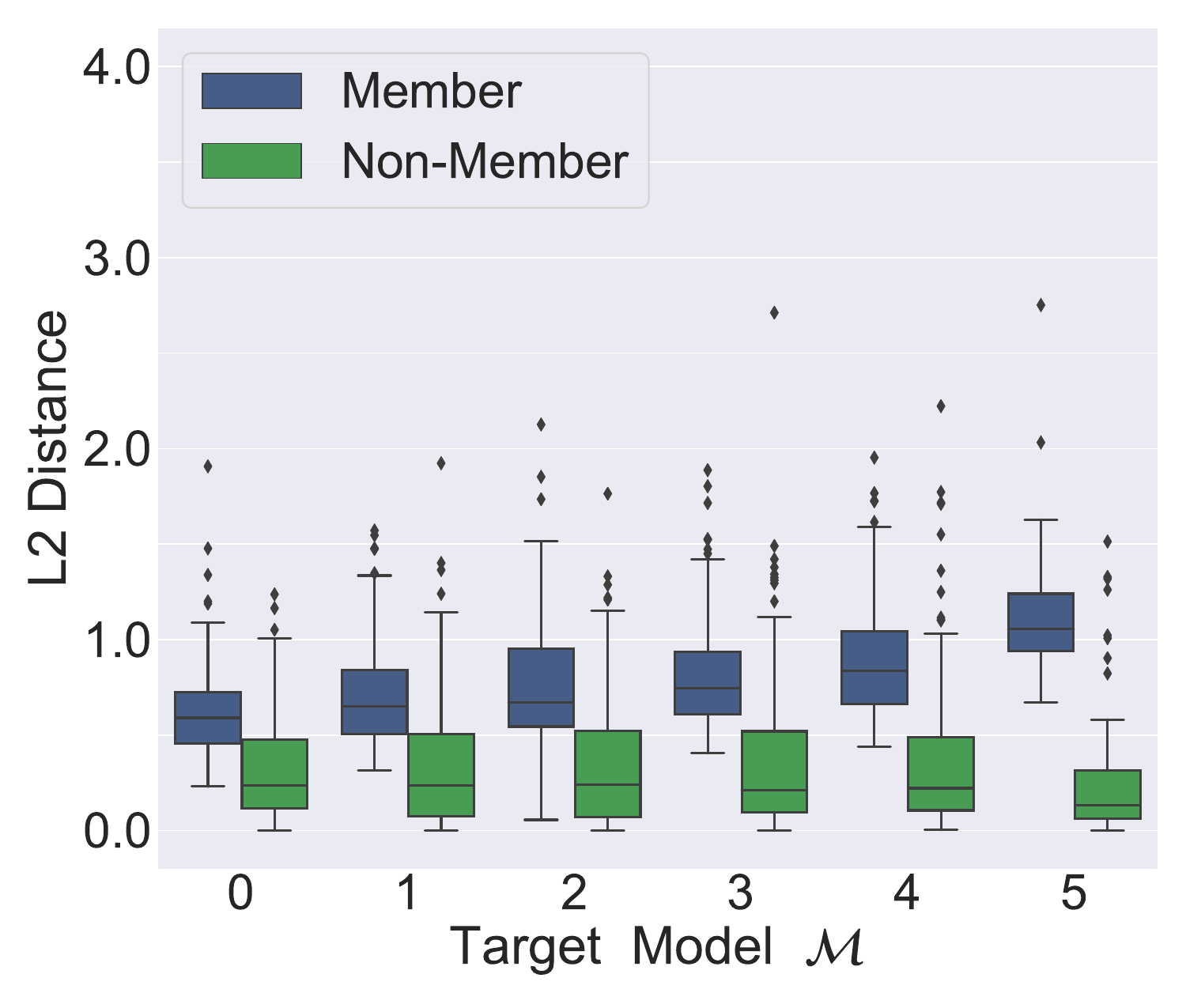}
\caption{CIFAR-10}
\end{subfigure}
\begin{subfigure}{0.5\columnwidth}
\includegraphics[width=\columnwidth]{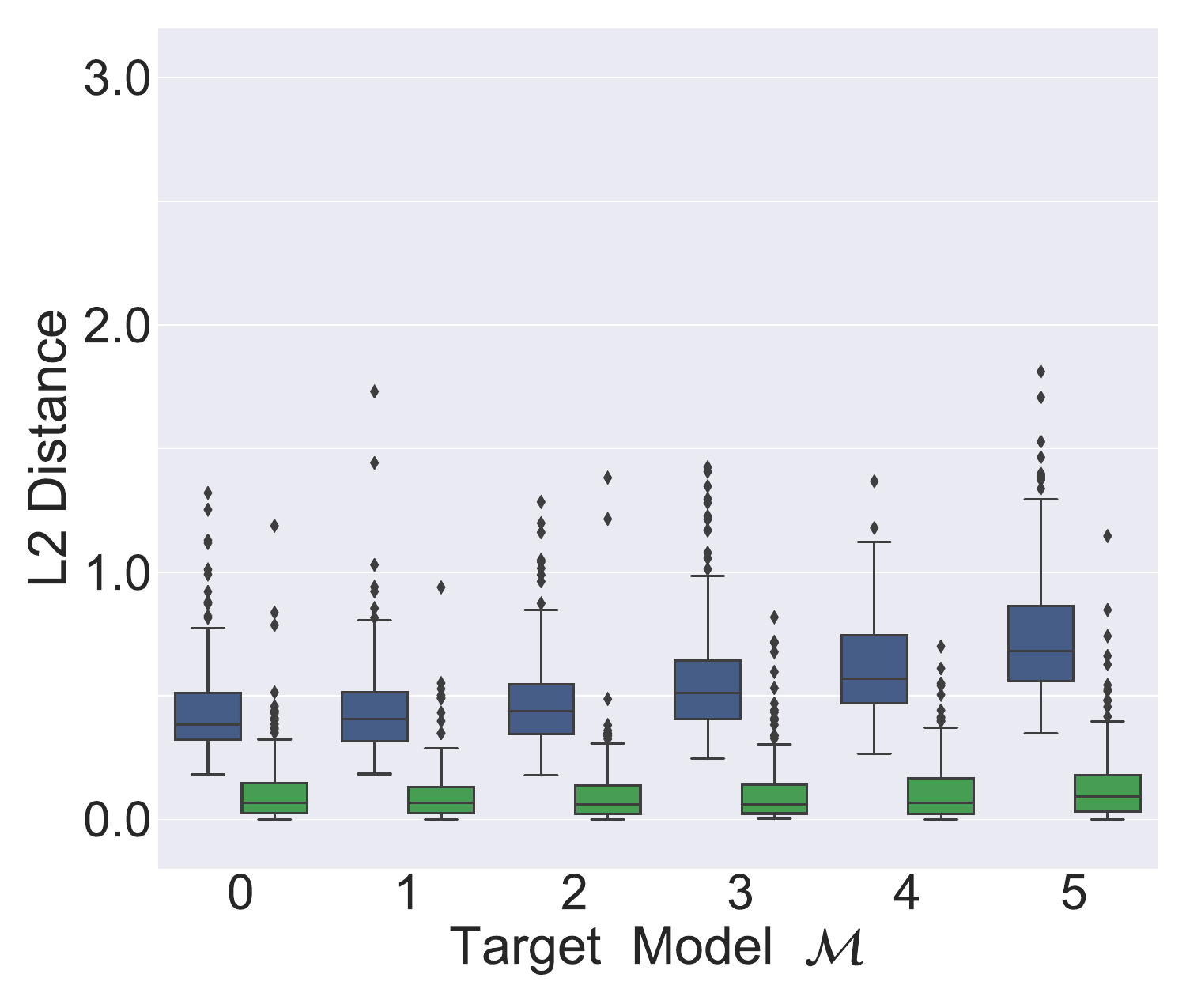}
\caption{CIFAR-100}
\end{subfigure}
\begin{subfigure}{0.5\columnwidth}
\includegraphics[width=\columnwidth]{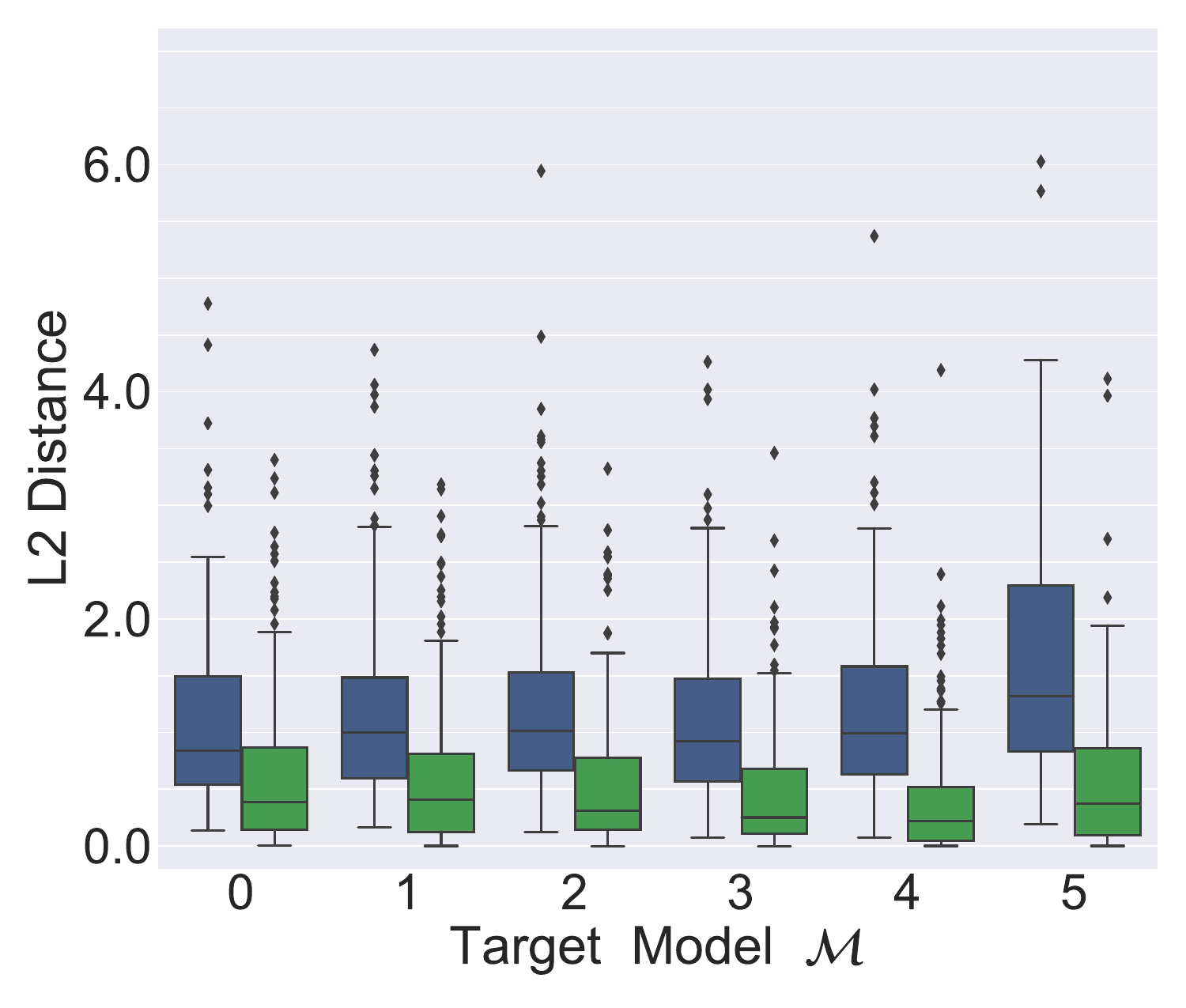}
\caption{GTSRB}
\end{subfigure}
\begin{subfigure}{0.5\columnwidth}
\includegraphics[width=\columnwidth]{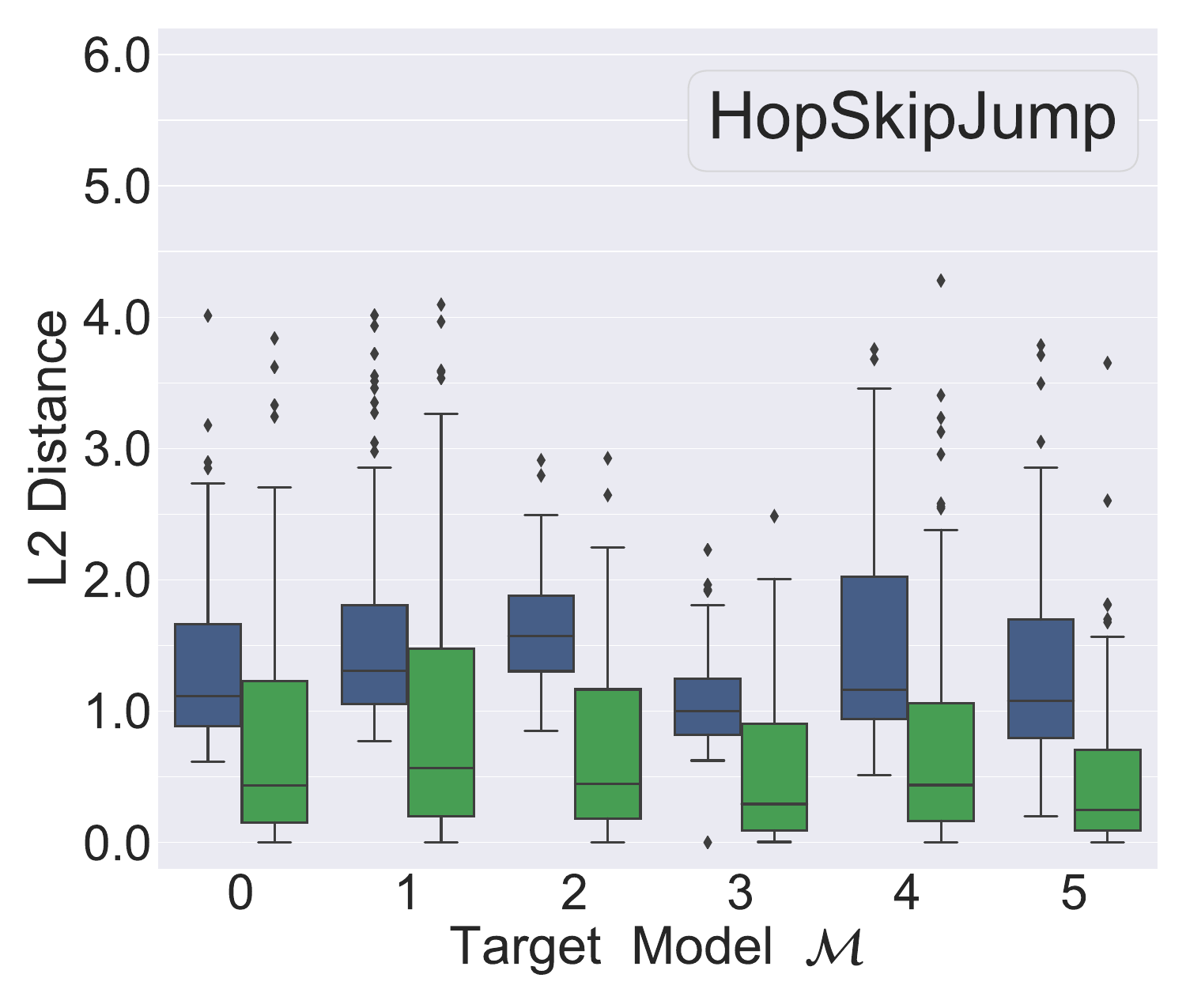}
\caption{Face}
\end{subfigure}
\caption{$L_2$ distance between the original sample and its perturbed samples generated by the HopSkipJump attack. 
The x-axis represents the target model being attacked and the y-axis represents the $L_2$ distance.}
\label{fig:dist_HSJA}
\end{figure*} 

\begin{figure*}[!ht]
\centering
\begin{subfigure}{0.5\columnwidth}
\includegraphics[width=\columnwidth,height=0.815\columnwidth]{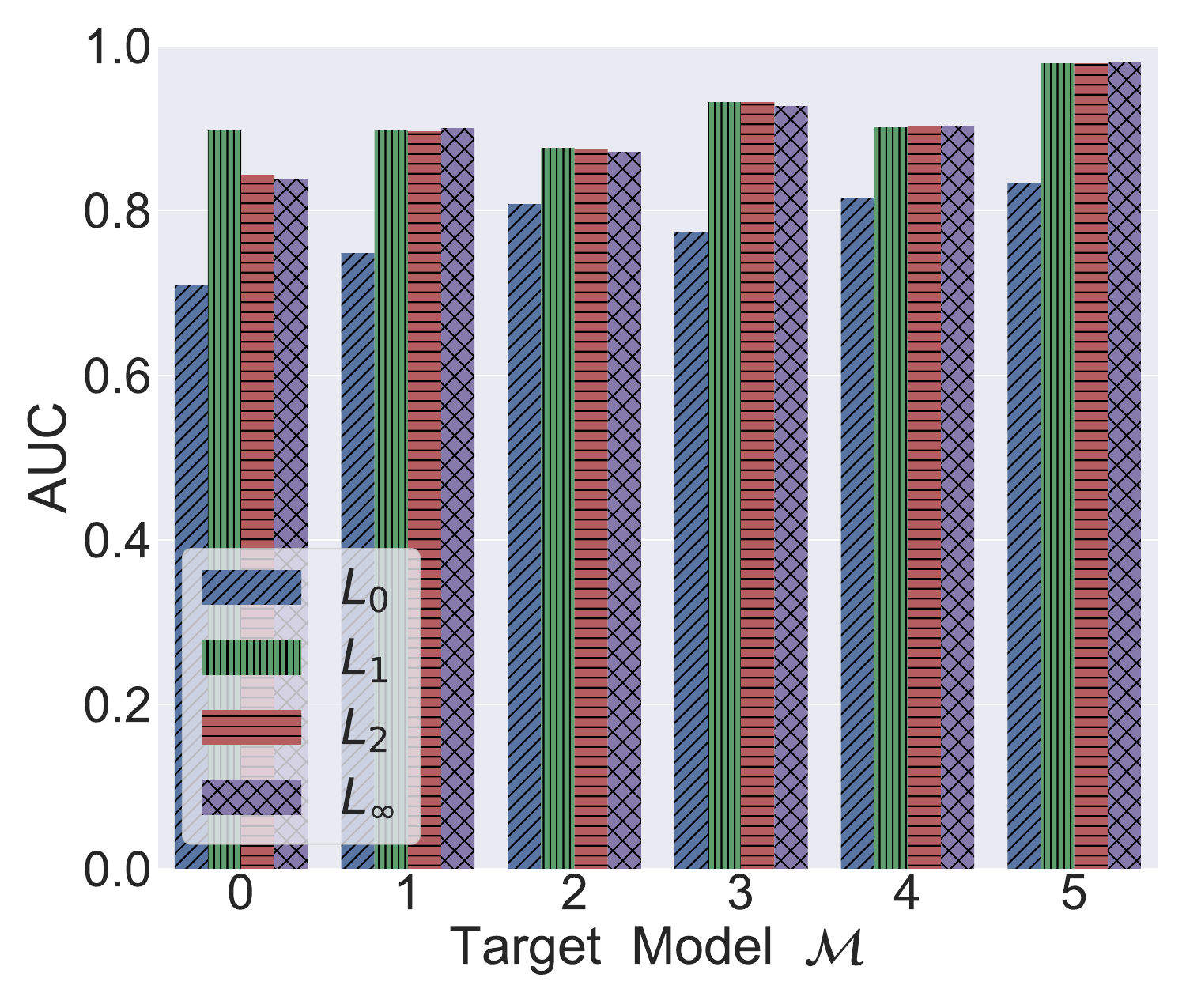}
\caption{CIFAR-10}
\end{subfigure}
\begin{subfigure}{0.5\columnwidth}
\includegraphics[width=\columnwidth,height=0.815\columnwidth]{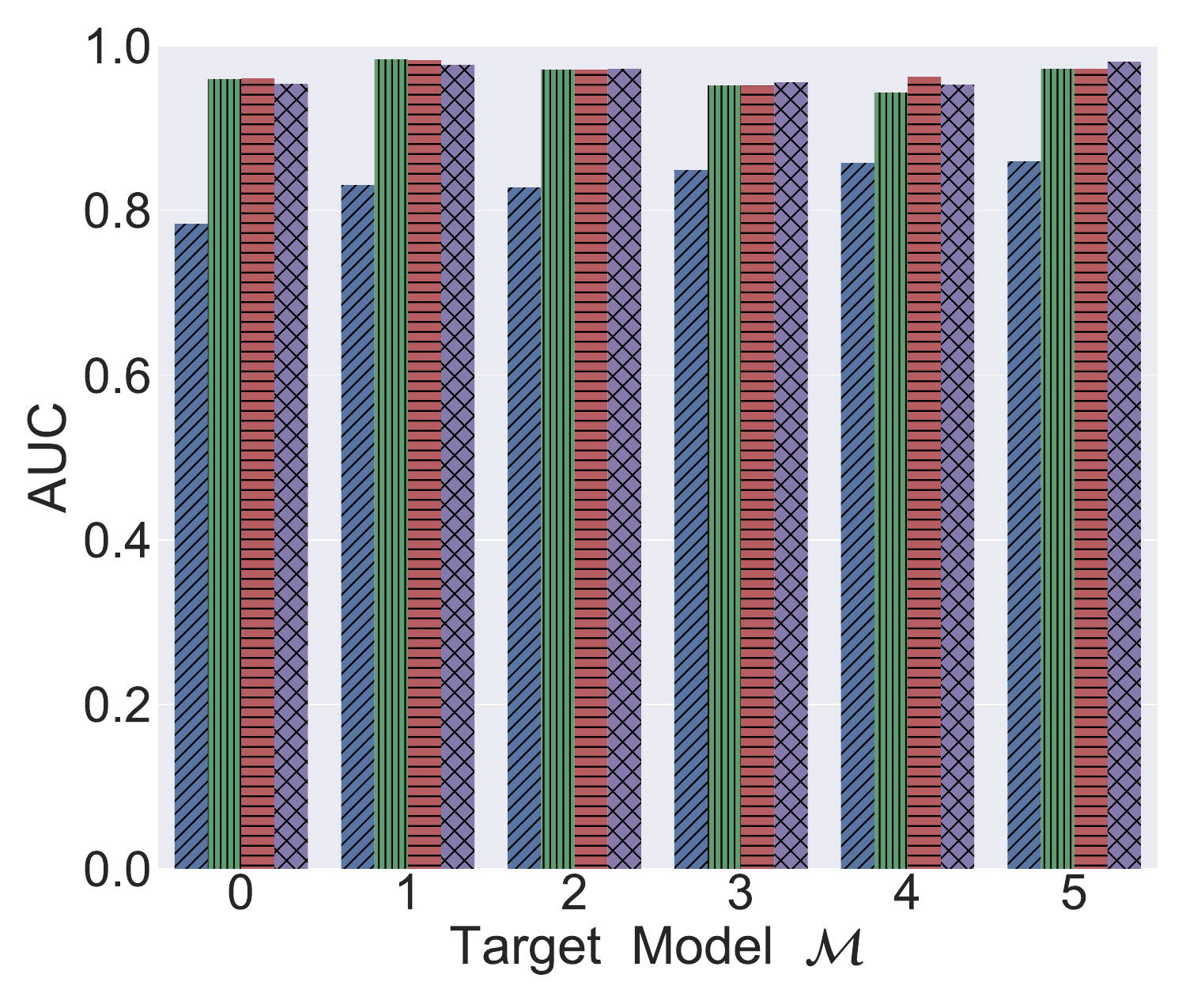}
\caption{CIFAR-100}
\end{subfigure}
\begin{subfigure}{0.5\columnwidth}
\includegraphics[width=\columnwidth,height=0.815\columnwidth]{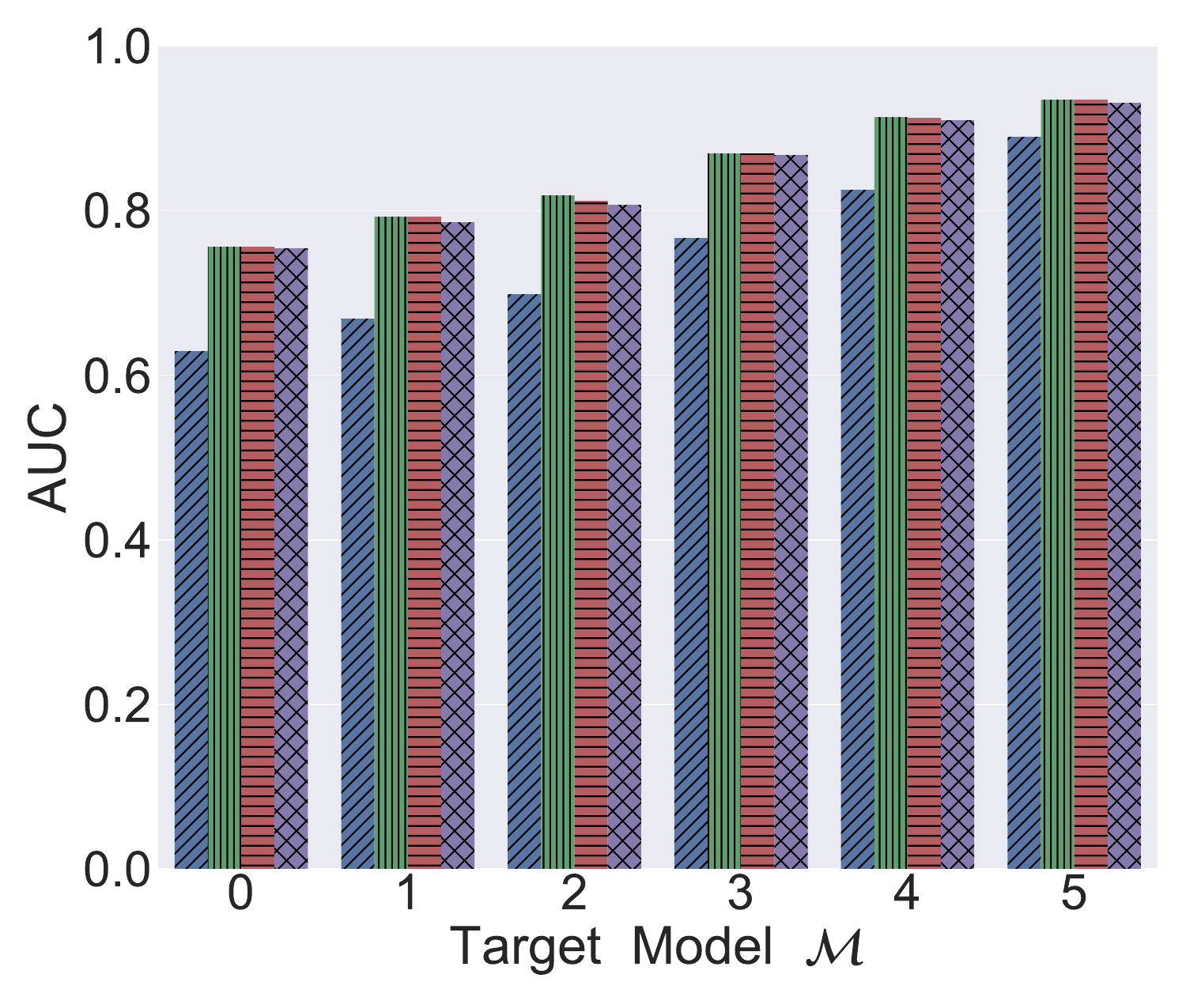}
\caption{GTSRB}
\end{subfigure}
\begin{subfigure}{0.5\columnwidth}
\includegraphics[width=\columnwidth,height=0.815\columnwidth]{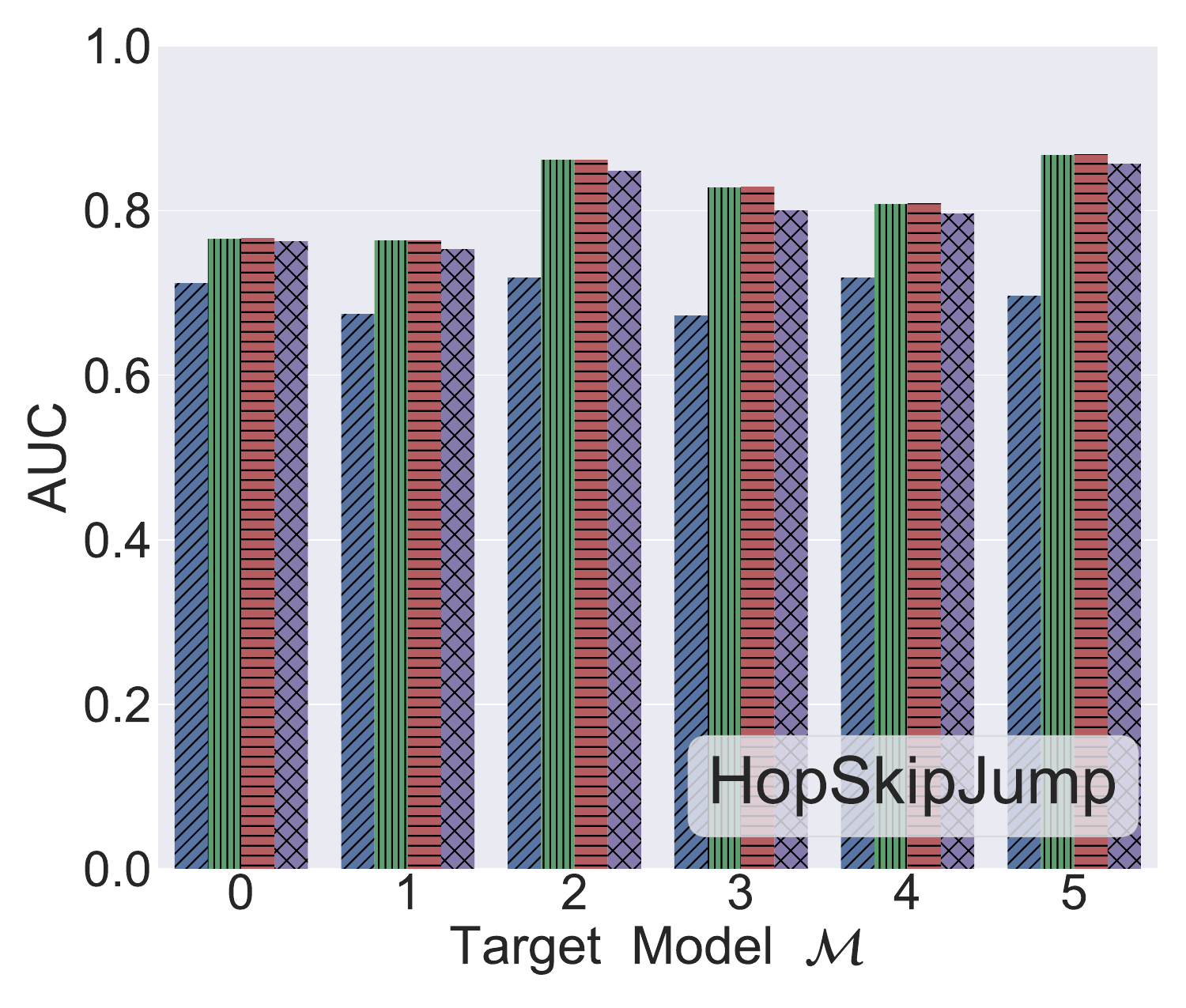}
\caption{Face}
\end{subfigure}
\caption{Attack AUC for four different $L_p$ distances between the original samples and its perturbed samples generated by the HopSkipJump attack. 
The x-axis represents the target model being attacked and the y-axis represents the AUC score.}
\label{fig:auc_HJSA}
\end{figure*} 

% ======================================================
\subsection{Experiment Setup}
% ======================================================

We use the same experimental setup as presented in~\autoref{adv1_setup}, such as the dataset splitting strategy and  6 target models trained on different size of training set $\train$.
In the decision change stage, we use the implementation of a popular python library (ART\footnote{\url{https://github.com/Trusted-AI/adversarial-robustness-toolbox}}) for HopSkipJump, and the authors' source code\footnote{\url{https://github.com/AI-secure/QEBA}} for QEBA. 
Note that we only apply untargeted decision change, i.e., changing the initial decision of the target model to any other random decision. 
Besides, both HopSkipJump and QEBA require multiple queries to perturb data samples to change their predicted labels. 
We set 15,000 for HopSkipJump and 7,000 for QEBA.
We further study the influence of the number of queries on the attack performance.
For space reasons, we report the results of HopSkipJump scheme in the main body of our paper.
Results of QEBA scheme can be found in Appendix \autoref{fig:dist_QEBA_SPnoise} and \autoref{fig:auc_QEBA_SPnoise}. 

% ======================================================
\subsection{Results}
\label{adv2_results}
% ======================================================

\mypara{Distribution of Perturbation}
First, we show the distribution of perturbation between a perturbed sample and its original one for member and non-member samples in \autoref{fig:dist_HSJA}. 
Both decision change schemes, i.e., HopSkipJump and QEBA, apply $L_{2}$ distance to limit the magnitude of perturbation, thus we report results of $L_{2}$ distance as well. 
As expected, the magnitude of the perturbation on member samples is indeed larger than that on non-member samples. 
For instance in \autoref{fig:dist_HSJA} ($\model$-5, CIFAR-10), the average $L_2$ distance of the perturbation for member samples is 1.0755, while that for non-member samples is 0.1102. 
In addition, models with a larger training set, i.e., lower overfitting level, require less perturbation to change the final prediction. 
As the overfitting level increases, the adversary needs to modify more on the member sample. 
The reason is that an ML model with a higher overfitting level has remembered its training samples to a larger extent, thus it is much harder to change their predicted labels, i.e., larger perturbation is required.

\mypara{Attack AUC Performance}
We report the AUC scores over all datasets in \autoref{fig:auc_HJSA}. 
In particular, we compare 4 different distance metrics, i.e., $L_{0}$, $L_{1}$, $L_{2}$, and $L_{\infty}$, for each decision change scheme. 
From \autoref{fig:auc_HJSA}, we can observe that $L_1$, $L_2$ and $L_{\infty}$ metrics achieve the best performance across all datasets. 
For instance in \autoref{fig:auc_HJSA} ($\model$-1, CIFAR-10), the AUC scores for $L_1$, $L_2$, and $L_{\infty}$ metrics are 0.8969, 0.8963, and 0.9033, respectively, while the AUC score for $L_0$ metric is 0.7405. 
From \autoref{fig:auc_QEBA_SPnoise} (in Appendix), we can also observe the same results of QEBA scheme: $L_1$, $L_2$  and $L_{\infty}$ metrics achieve the best performance across all datasets, while $L_0$ metric performs the worst. 
Therefore, an adversary can simply choose the same distance metric adopted by adversarial attacks to measure the magnitude of the perturbation.

\begin{figure}[!t]
\centering
\includegraphics[width=0.80\columnwidth]{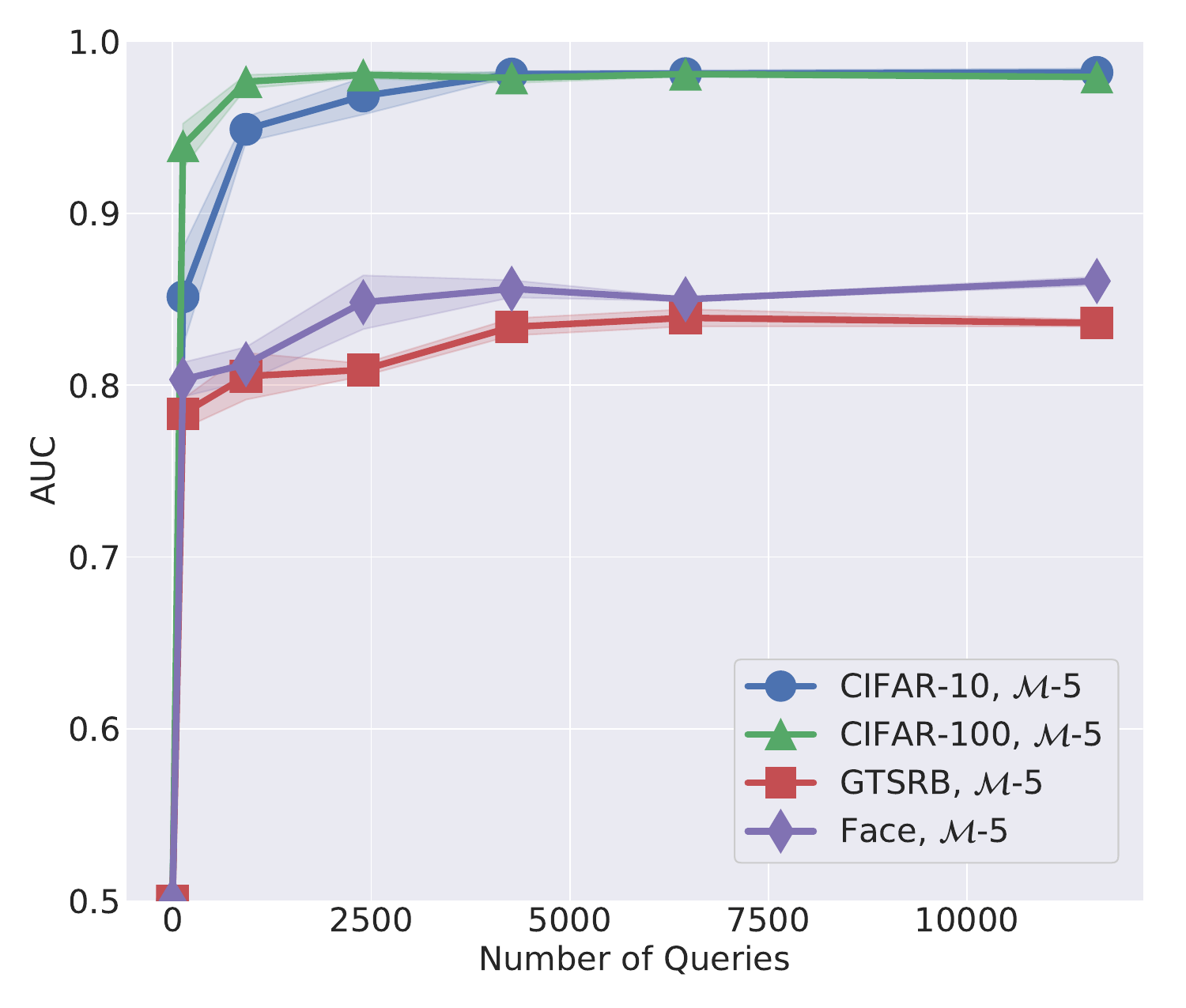}
\caption{Attack AUC under the effect of number of queries. 
The x-axis represents the number of queries and the y-axis represents the AUC score for perturbation-based attack.}
\label{fig:query}
\end{figure}

\begin{figure*}[!t]
\centering
\begin{subfigure}{0.5\columnwidth}
\includegraphics[width=\columnwidth,height=0.815\columnwidth]{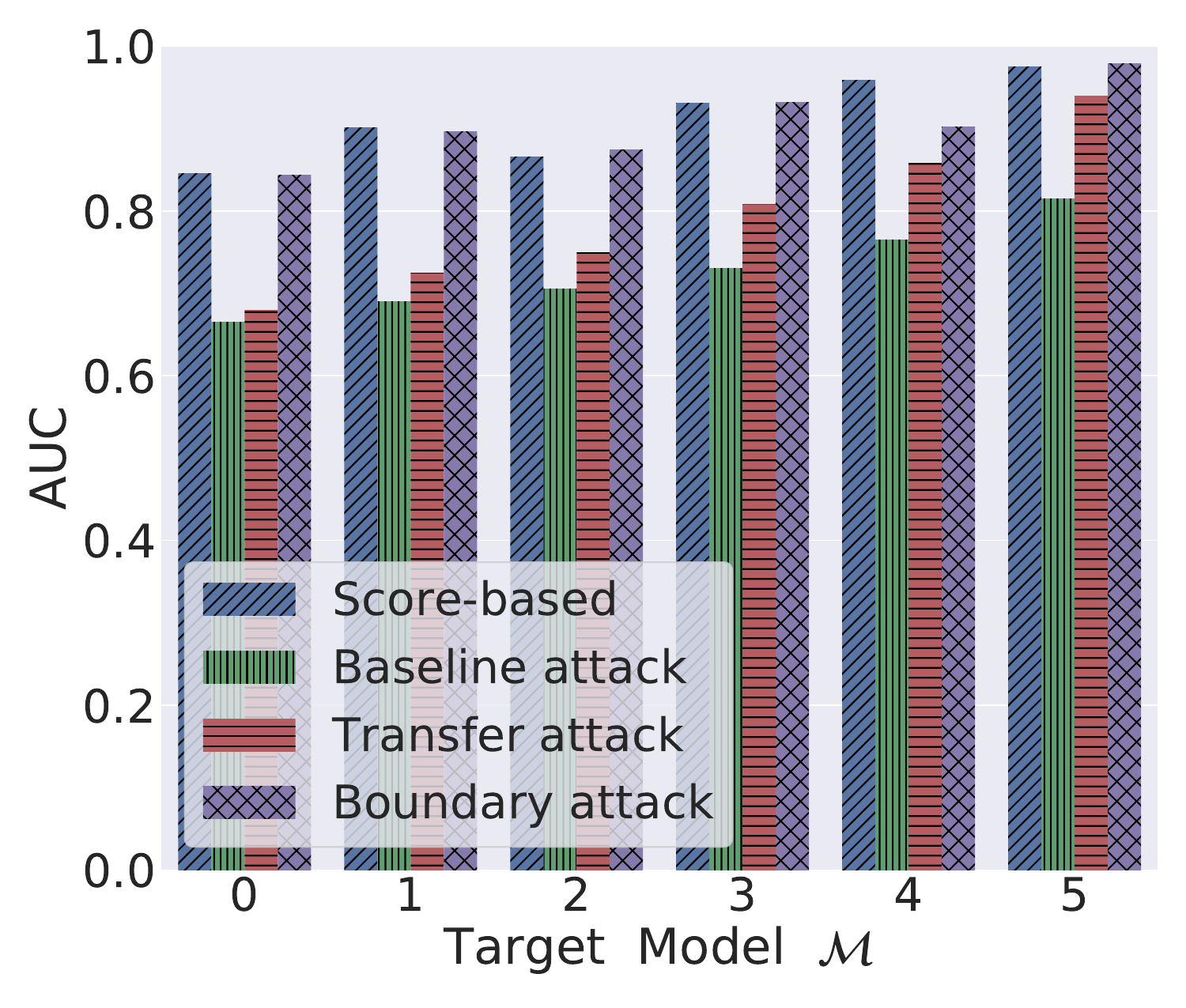}
\caption{CIFAR-10}
\end{subfigure}
\begin{subfigure}{0.5\columnwidth}
\includegraphics[width=\columnwidth,height=0.815\columnwidth]{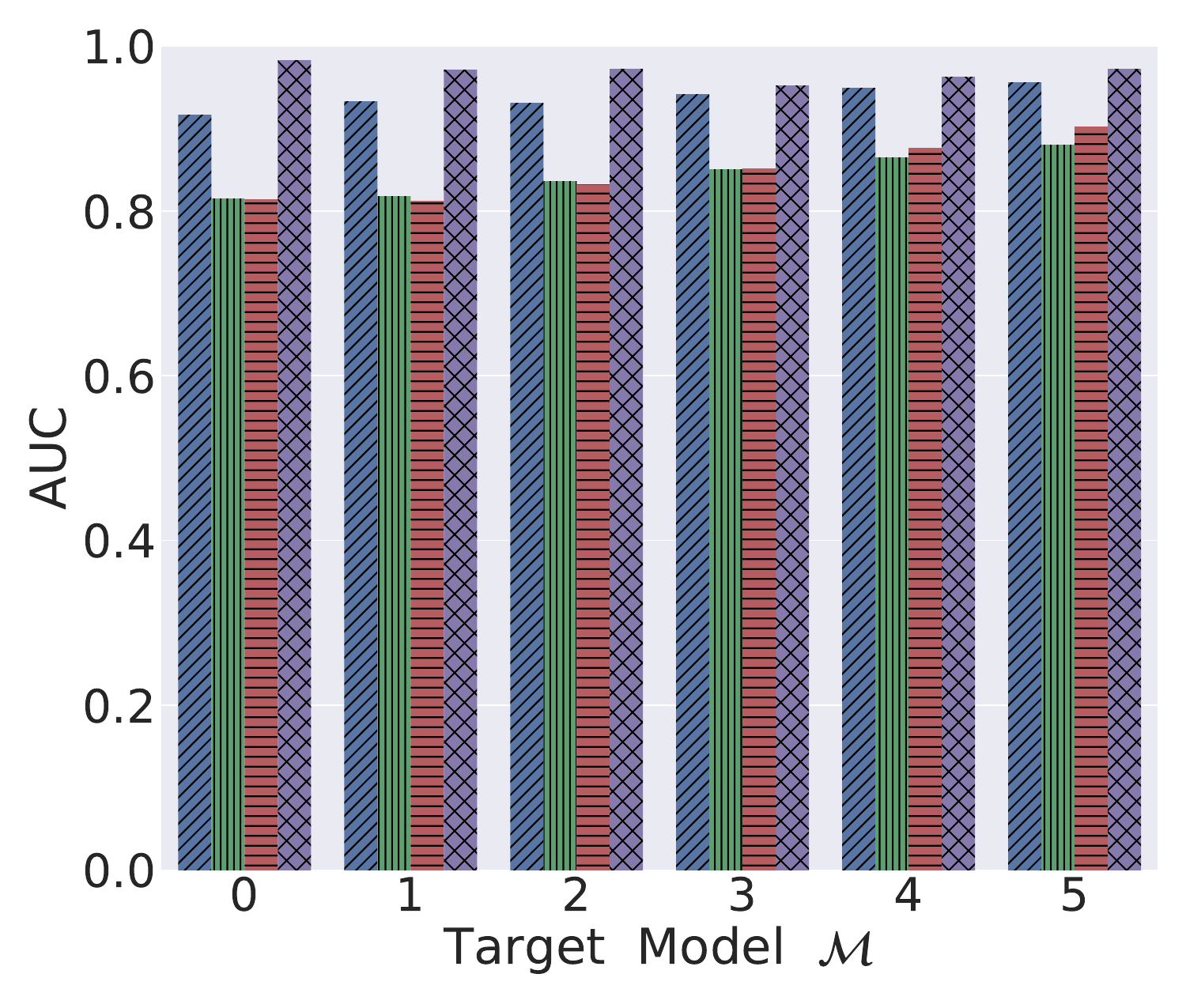}
\caption{CIFAR-100}
\end{subfigure}
\begin{subfigure}{0.5\columnwidth}
\includegraphics[width=\columnwidth,height=0.815\columnwidth]{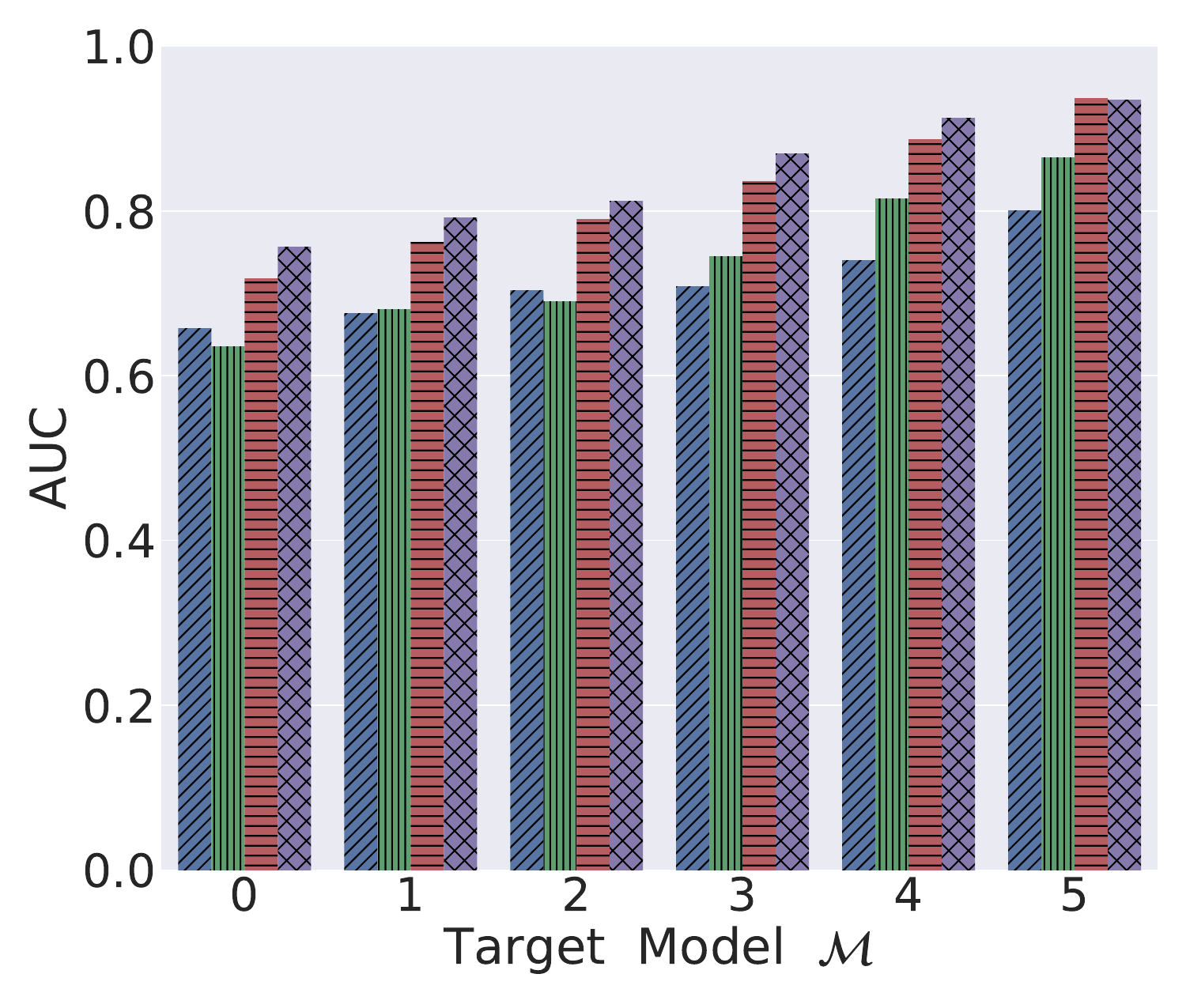}
\caption{GTSRB}
\end{subfigure}
\begin{subfigure}{0.5\columnwidth}
\includegraphics[width=\columnwidth,height=0.815\columnwidth]{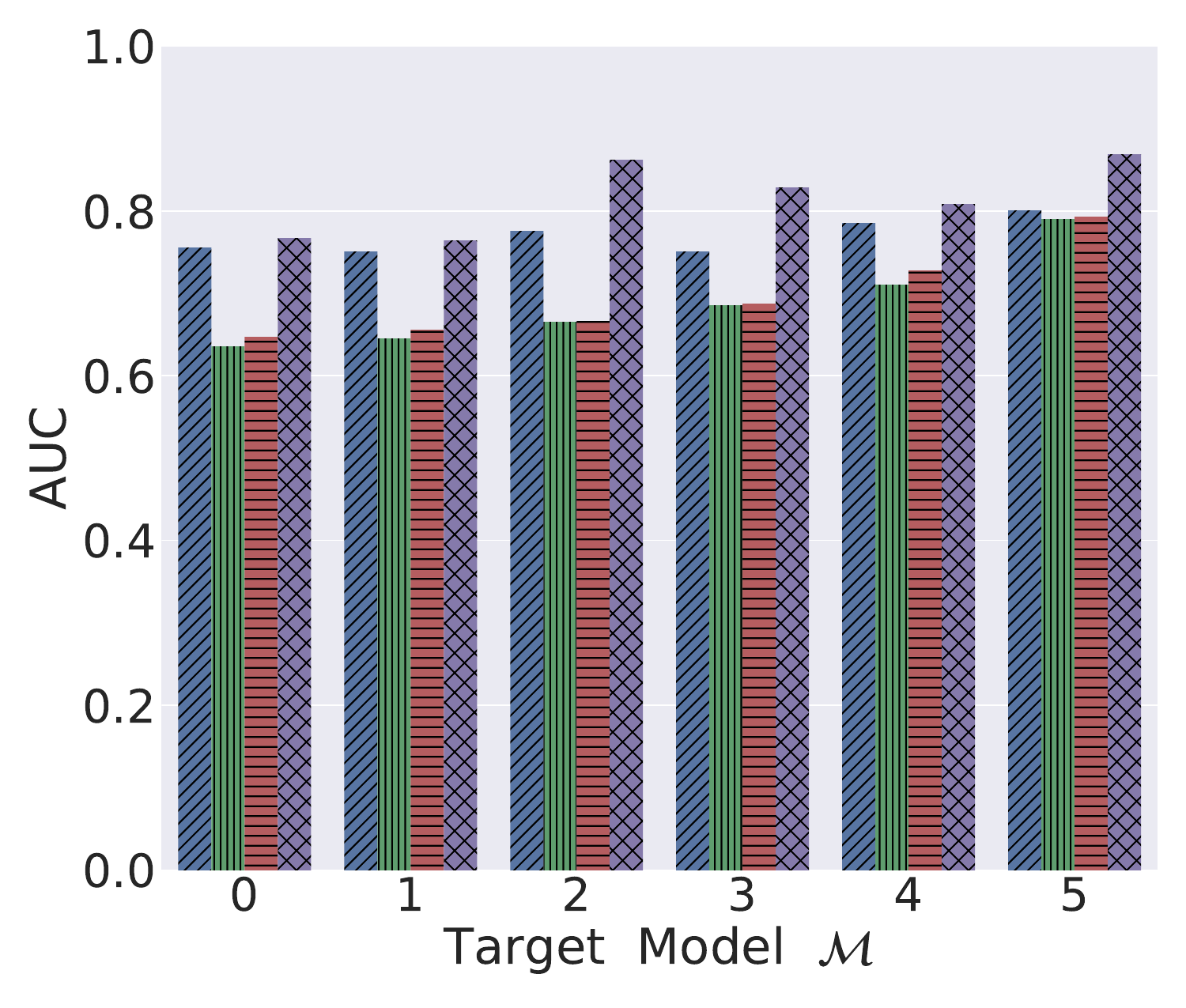}
\caption{Face}
\end{subfigure}
\caption{Comparison of our two types of attacks with the baseline attack and score-based attack. 
The x-axis represents the target model being attacked and the y-axis represents the AUC score.}
\label{fig:comparisonALL}
\end{figure*} 

\begin{figure}[!t]
\centering
\includegraphics[width=0.80\columnwidth]{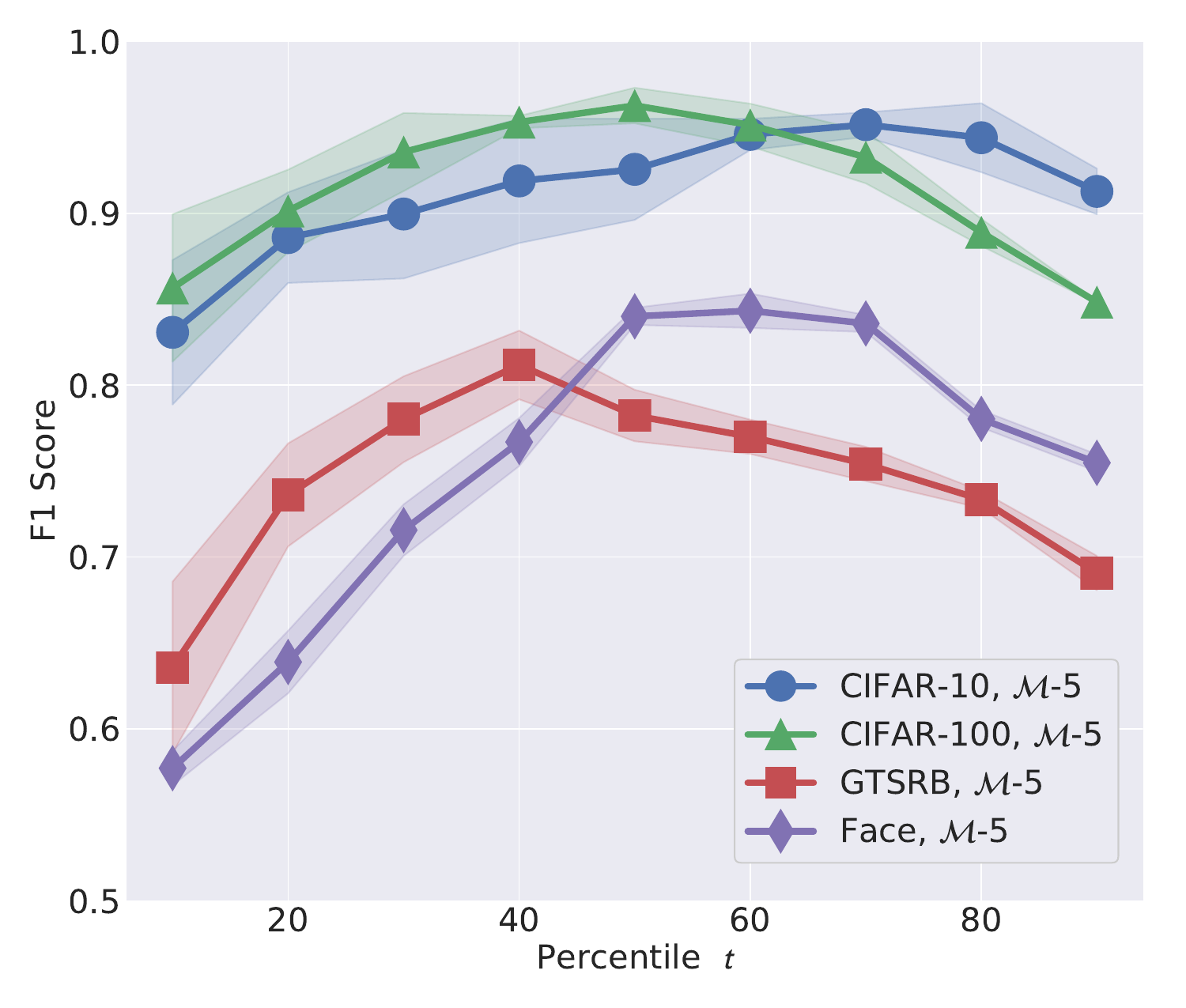}
\caption{The relation between the top $t$ percentile of the $L_2$ distance, i.e., threshold, and the attack performance.
The x-axis represents the top $t$ percentile and the y-axis represents the F1 score.}
\label{fig:threshold}
\end{figure}

\mypara{Effects of Number of Queries}
To mount boundary attack in real-world ML applications such as Machine Learning as a Service (MLaaS), the adversary cannot issue as many queries as they want to the target model, since a large number of queries increases the cost of the attack and may raise the suspicion of the model provider. 
Now, we evaluate the attack performance with different number of queries. 
Here, we show the results of the HopSkipJump scheme for $\model$-5 over all datasets. 
We vary the number of queries from 0 to 15,000 and evaluate the attack performance based on the $L_{2}$ metric. 
As we can see in \autoref{fig:query}, the AUC increases sharply as the number of queries increases in the beginning. 
After 2,500 queries, the attack performance becomes stable. 
From the results, we argue that query limiting would likely not be a suitable defense. For instance, when querying 131 times, the AUC for CIFAR-10 is 0.8228 and CIFAR-100 is 0.9266. 
At this time, though the perturbed sample is far away from its origin's decision boundary, the magnitude of perturbation for member samples is still relatively larger than that for non-member samples. 
Thus, the adversary can still differentiate member and non-member samples.

\mypara{Threshold Choosing}
Here, we focus on the threshold choosing for our boundary attack where the adversary is not equipped with a shadow dataset. 
We provide a simple and general method for choosing a threshold. 
Concretely, we generate a set of random samples in the feature space as the target model's training set. In the case of image classification, we sample each pixel for an image from a uniform distribution. 
Next, we treat these randomly generated samples as non-members and query them to the target model.
Then, we apply adversarial attack techniques on these random samples to change their initial predicted labels by the target model. 
Finally, we use these samples' perturbation to estimate a threshold, i.e., finding a suitable top $t$ percentile over these perturbations. 
The algorithm can be found in Appendix \autoref{alg:threshold}.

We experimentally generate 100 random samples for $\model$-5 trained across all datasets, and adopt HopSkipJump in decision change stage. 
We again use the $L_2$ distance to measure the magnitude of perturbation and F1 score as our evaluation metric. 
From \autoref{fig:threshold}, we make the following observations:

\begin{itemize}
\item The peak attack performance is bounded between $t=0\%$ and $t=100\%$, which means the best threshold can definitely be selected from these random samples' perturbation.
\item The powerful and similar attack performance ranges from $t=30\%$ to $t=80\%$, reaching half of the total percentile, which means that a suitable threshold can be easily selected.
\end{itemize}
Therefore, we conclude that our threshold choosing method is effective and can achieve excellent performance.

\begin{table*}[!ht]
\definecolor{mygray}{gray}{0.9}
\centering
\caption{Average Certified Radius (ACR)  of members and non-members for target models.}
\label{table:target_ACR}
\scalebox{0.83}
{
\begin{tabular}{c|>{\columncolor{mygray}}c|c|>{\columncolor{mygray}}c|c|>{\columncolor{mygray}}c|c|>{\columncolor{mygray}}c|c }
\toprule
Target&  \multicolumn{2}{c}{CIFAR-10}& \multicolumn{2}{c}{CIFAR-100} & \multicolumn{2}{c}{GTSRB} & \multicolumn{2}{c}{Face}\\
 Model&Member&Non-mem&Member&Non-mem&Member&Non-mem&Member&Non-mem\\
\midrule
 $\model$-0 &  {0.1392}& 0.1201& {0.0068}& 0.0033& {0.0300}& 0.0210& {0.0571}& 0.0607\\
 $\model$-1 &  {0.1866}& 0.1447& {0.0133}& 0.0079& {0.0358}& 0.0215& {0.0290}& 0.0190\\
 $\model$-2 &  {0.1398}& 0.1170& {0.0155}& 0.0079& {0.0692}& 0.0463& {0.0408}& 0.0313\\
 $\model$-3 &  {0.1808}& 0.1190& {0.0079}& 0.0074& {0.0430}& 0.0348& {0.1334}& 0.1143\\
 $\model$-4 &  {0.1036}& 0.1032& {0.0141}& 0.0116& {0.0212}& 0.0176& {0.0392}& 0.0292\\
 $\model$-5 &  {0.1814}& 0.0909& {0.0157}& 0.0080& {0.0464}& 0.0385& {0.1242}& 0.1110\\
\bottomrule
\end{tabular}
}
\end{table*}

\mypara{Comparison of Different Attacks}
Now we compare the performance of our two attacks and previous existing attacks. 
In particular, we also compare our attacks against prior score-based attacks. Following the score-based attack proposed by Salem et al.~\cite{SZHBFB19}, we train one shadow model using half of $\shadowData$ with its ground truth labels, and one attack model in a supervised manner based on the shadow model’s output scores. 
Here, we do not assume that the attacker knows the exact training set size of the target model, which is actually a strong assumption. 
Note that this is not a fair comparison, as our decision-based attacks only access to the final model's prediction, rather than the confidence scores.

We report attack performance for our boundary attack using $L_2$ metric in HopSkipJump scheme. 
From \autoref{fig:comparisonALL}, we can find that our boundary attack achieves similar or even better performance than the score-based attack in some cases. 
This demonstrates the efficacy of our proposed decision-based attack, thereby the corresponding membership leakage risks stemming from ML models are much more severe than previously shown.

As for cost analysis, the attack logic is different for each method, so it is difficult to evaluate the cost with standard metrics. 
Besides the adversarial knowledge acquired for each attack, we mainly report training costs and query costs in \autoref{table:cost}. 
We can find the baseline attack only queries once for a candidate sample. 
However, in our transfer attack, once a shadow model is built, the adversary will only query the shadow model for candidate samples without making any other queries to the target model. 
Therefore, we cannot prematurely claim that the baseline attack has the lowest cost, but should consider the actual situation.

\begin{table}[!ht]
\centering
\caption{The cost of each attack. Query cost is the number of queries to the target model.}
\scalebox{0.83}
{
\begin{tabular}{c|c|c|c}
\toprule
Attack& Shadow Model & Query for & Query for a\\
Type  & Training Epochs & $\shadowData$ &candidate sample\\
\midrule
score-based & 200 &- & 1\\
baseline attack & - & - & 1\\
transfer attack & 200 & |$\shadowData$| &-\\
boundary attack & - &- & Multiple\\
\bottomrule
\end{tabular}
}
\label{table:cost}
\end{table}

% ======================================================
\section{Membership Leakage Analysis}
\label{sec:analysis}
% ======================================================

The above results fully demonstrate the effectiveness of our decision-based attacks. 
Here, we delve more deeply into the reasons for the success of membership inference. 
Our boundary attack utilizes the magnitude of the perturbation to determine whether the sample is a member or not, and the key to stop searching perturbations is the final decision change of the model. 
Here, the status of decision change actually contains information about the decision boundary, i.e., the perturbed sample crosses the decision boundary. 
This suggests a new perspective on the relationship between member samples and non-member samples, and we intend to analyze membership leakage from this perspective. 
Since previous experiments have verified our key intuition that the perturbation required to change the predicted label of a member sample is larger than that of a non-member, we argue that the distance between the member sample and its decision boundary is typically larger than that of the non-member sample. 
Next, we will verify it both quantitatively and qualitatively.

% ======================================================
\subsection{Quantitative Analysis}
\label{subsec:QuantitativeAnalysis} 
% ======================================================

\begin{figure*}
\centering
\begin{subfigure}{0.5\columnwidth}
\includegraphics[width=\columnwidth]{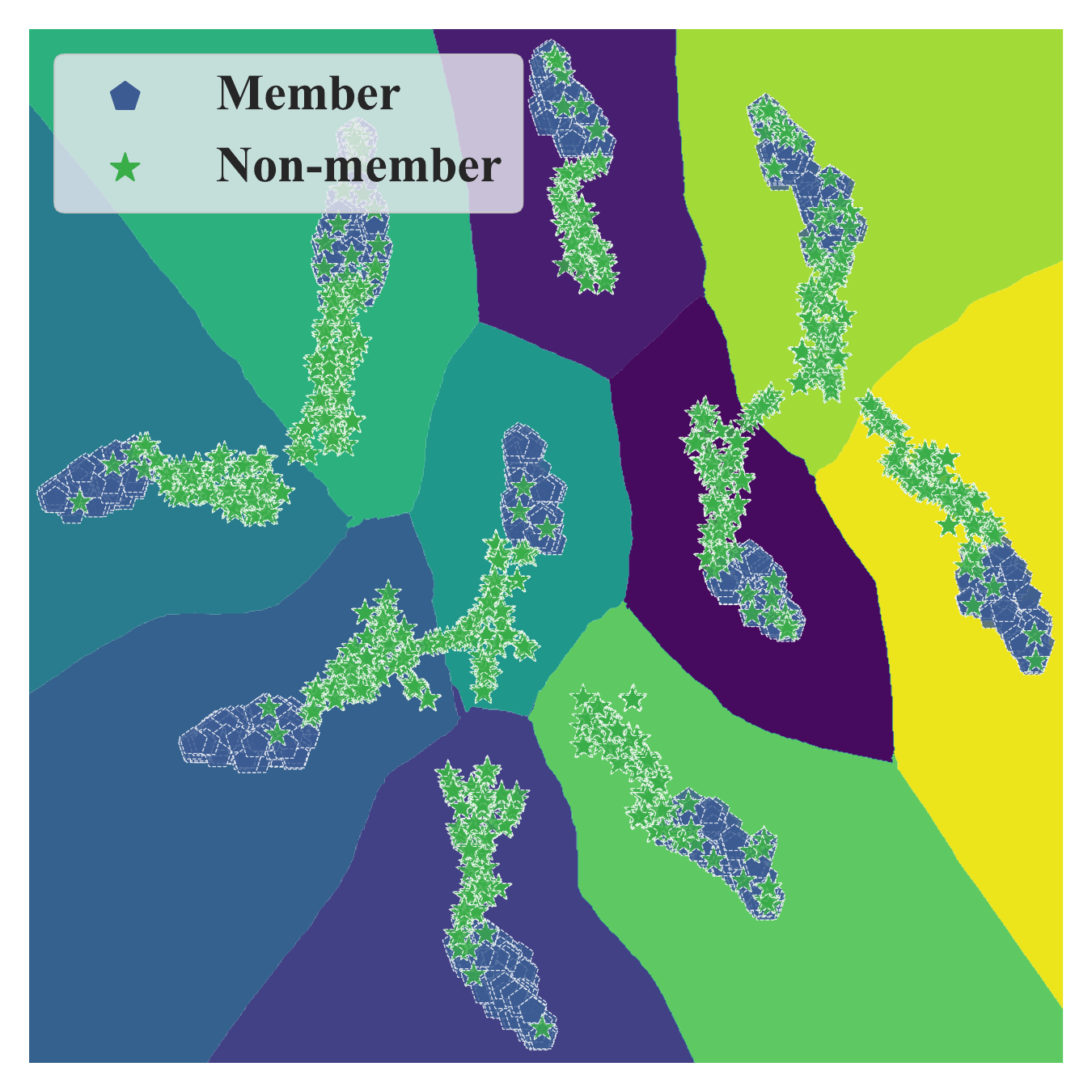}
\caption{Target Model, Zoom-out}
\label{fig:target_db}
\end{subfigure}
\begin{subfigure}{0.5\columnwidth}
\includegraphics[width=\columnwidth]{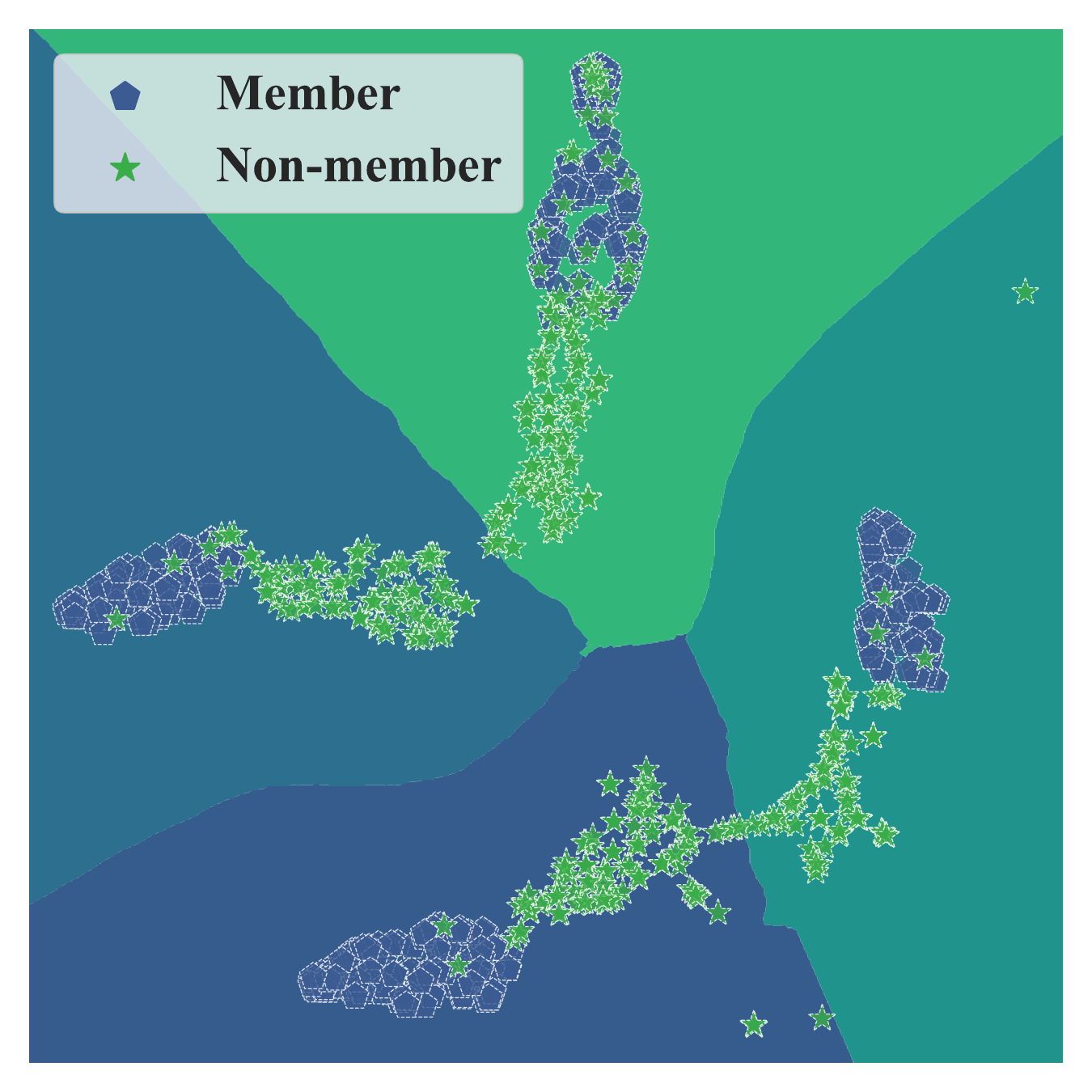}
\caption{Target Model, Zoom-in}
\label{fig:target_db_zoomin}
\end{subfigure}
\begin{subfigure}{0.5\columnwidth}
\includegraphics[width=\columnwidth]{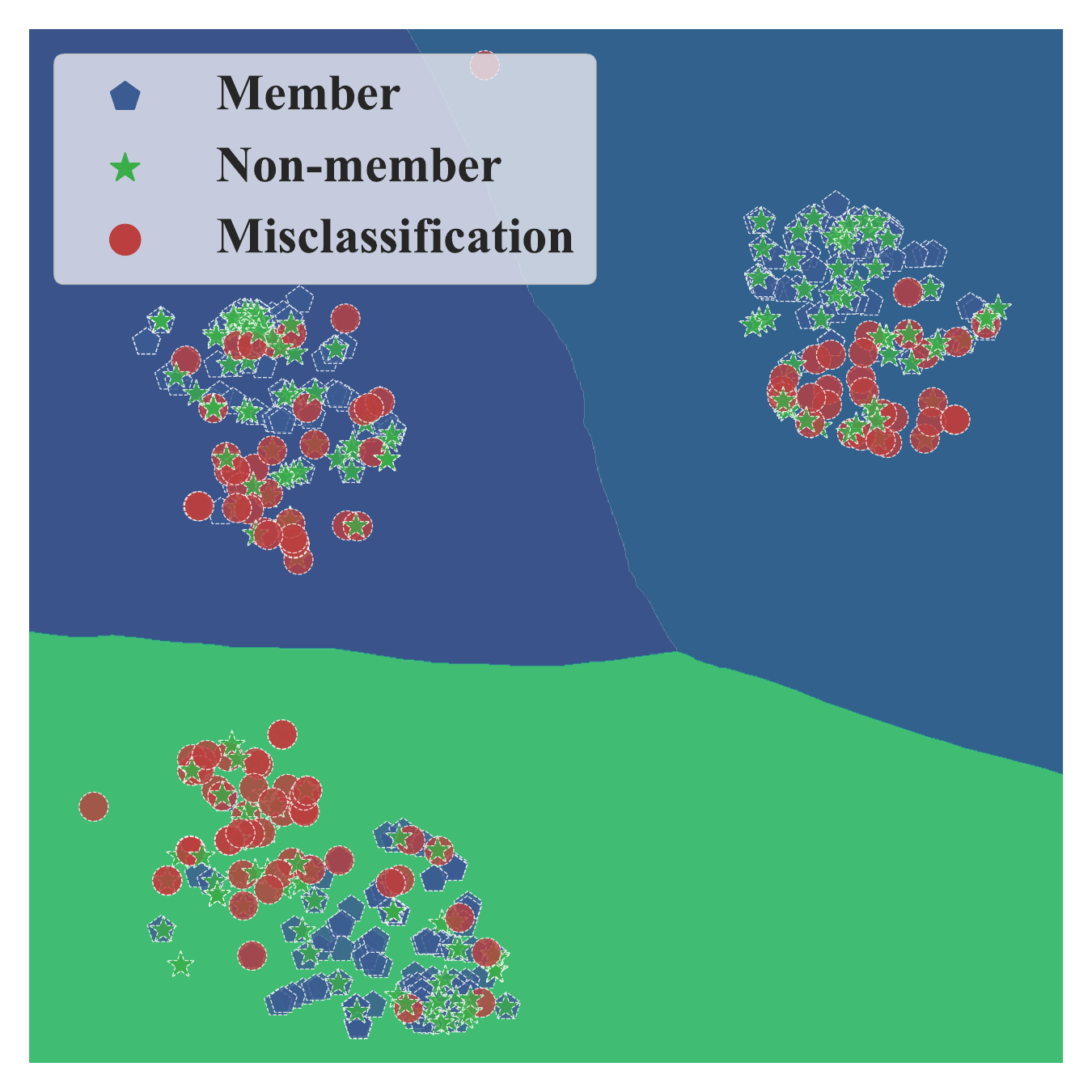}
\caption{Shadow Model, Zoom-in}
\label{fig:shadow_db_zoomin}
\end{subfigure}
\begin{subfigure}{0.5\columnwidth}
\includegraphics[width=\columnwidth, height=\columnwidth]{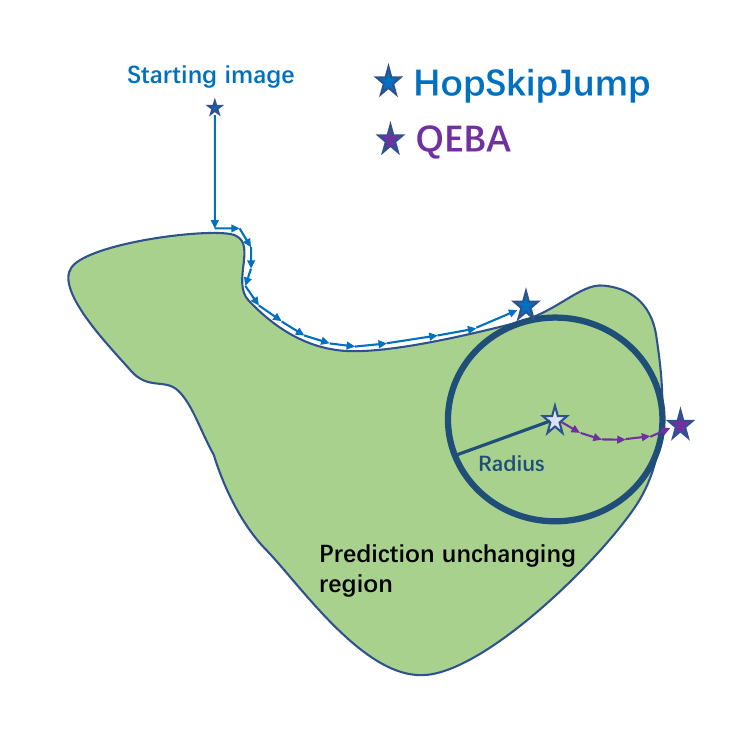}
\caption{Adversarial attack process}
\label{fig:attack_schema}
\end{subfigure}
\caption{The visualization of decision boundary for target model (a, b) and shadow model (c), and the search process of perturbed sample by HopSkipJump and QEBA (d).}
\label{fig:decision_boundary}
\end{figure*}

We introduce the neighboring $L_p$-radius ball to investigate the membership leakage of ML models. 
This neighboring $L_p$-radius ball, also known as \emph{Robustness Radius}, is defined as the $L_p$ robustness of the target model at a data sample, which represents the radius of the largest $L_p$ ball centered at the data sample in which the target model does not change its prediction, as shown in \autoref{fig:attack_schema}. 
Concretely, we investigate the $L_2$ robustness radius of the target model $\model$ at a data sample $x$. Unfortunately, computing the robustness radius of a ML model is a hard problem. 
Researchers have proposed many certification methods to derive a tight lower bound of robustness radius $ R(\model;x,y)$ for ML models. 
Here, we also derive a tight lower bound of robustness radius, namely \emph{Certified  Radius}~\cite{ZDHZGRHW20}, which satisfies $0 \leq CR(\model;x,y) \leq R(\model;x,y)$ for any $\model, x$ and its ground truth label $y \in \mathcal{Y} =\{1,2,\cdots,K\}$. 
More details about certified  radius can be found in Appendix \autoref{app:CR}.

\mypara{ACR of Members and Non-members}
As we can see from Theorem 1 (see Appendix \autoref{app:CR}), the value of the certified radius can be estimated by repeatedly sampling Gaussian noises. 
For the target model $\model$ and a data sample $(x,y)$, we can estimate the certified radius $CR(\model;x,y)$. 
Here, we use the \emph{average certified radius} (ACR) as a metric to estimate the average certified radius for members and non-members separately, i.e.,
\begin{align}
ACR_{member} = \frac{1}{|\train|}\sum_{(x,y)\in\train} CR(\model; x, y),\\
ACR_{non-member} = \frac{1}{|\test|}\sum_{(x,y)\in\test} CR(\model; x, y).
\end{align}

\begin{table}[!ht]
\definecolor{mygray}{gray}{0.9}
\centering
\caption{Average Certified Radius (ACR) of members and non-members for shadow models.}
\label{table:shadow_acr}
\scalebox{0.83}
{
\begin{tabular}{c|>{\columncolor{mygray}}c|c|>{\columncolor{mygray}}c|c }
\toprule
Shadow&  \multicolumn{2}{c}{CIFAR-10}&  \multicolumn{2}{c}{CIFAR-100}\\
 Model&Member&Non-mem&Member&Non-mem\\
\midrule
 $\model$-0 &  {0.1392}& 0.1301&  {0.0091}& 0.0039\\
 $\model$-1 &  {0.1873}& 0.1516&  {0.0150}& 0.0071\\
 $\model$-2 &  {0.1416}& 0.1463&  {0.0177}& 0.0068\\
 $\model$-3 &  {0.1962}& 0.1452&  {0.0121}& 0.0047\\
 $\model$-4 &  {0.1152}& 0.1046&  {0.0099}& 0.0092\\
 $\model$-5 &  {0.1819}& 0.0846&  {0.0176}& 0.0087\\
\bottomrule
\end{tabular}
}
\end{table}

We randomly select an equal number of members and non-members for target models and report the results in \autoref{table:target_ACR}. 
Note that the certified radius is actually an estimated value representing the lower bound of the robustness radius, not the exact radius. 
Therefore, we analyze the results from a macroscopic perspective and can draw the following observations.

\begin{itemize}
\item The ACR of member samples is generally larger than the ACR of non-member samples, which means that in the output space, the ML model maps member samples further away from its decision boundary than non-member samples.
\item As the level of overfitting increases, the macroscopic trend of the gap between the ACR of members and non-members is also larger, which exactly reflects the increasing attack performance in the aforementioned AUC results.
\end{itemize}

\begin{figure*}[!t]
\centering
\begin{subfigure}{0.5\columnwidth}
\includegraphics[width=\columnwidth]{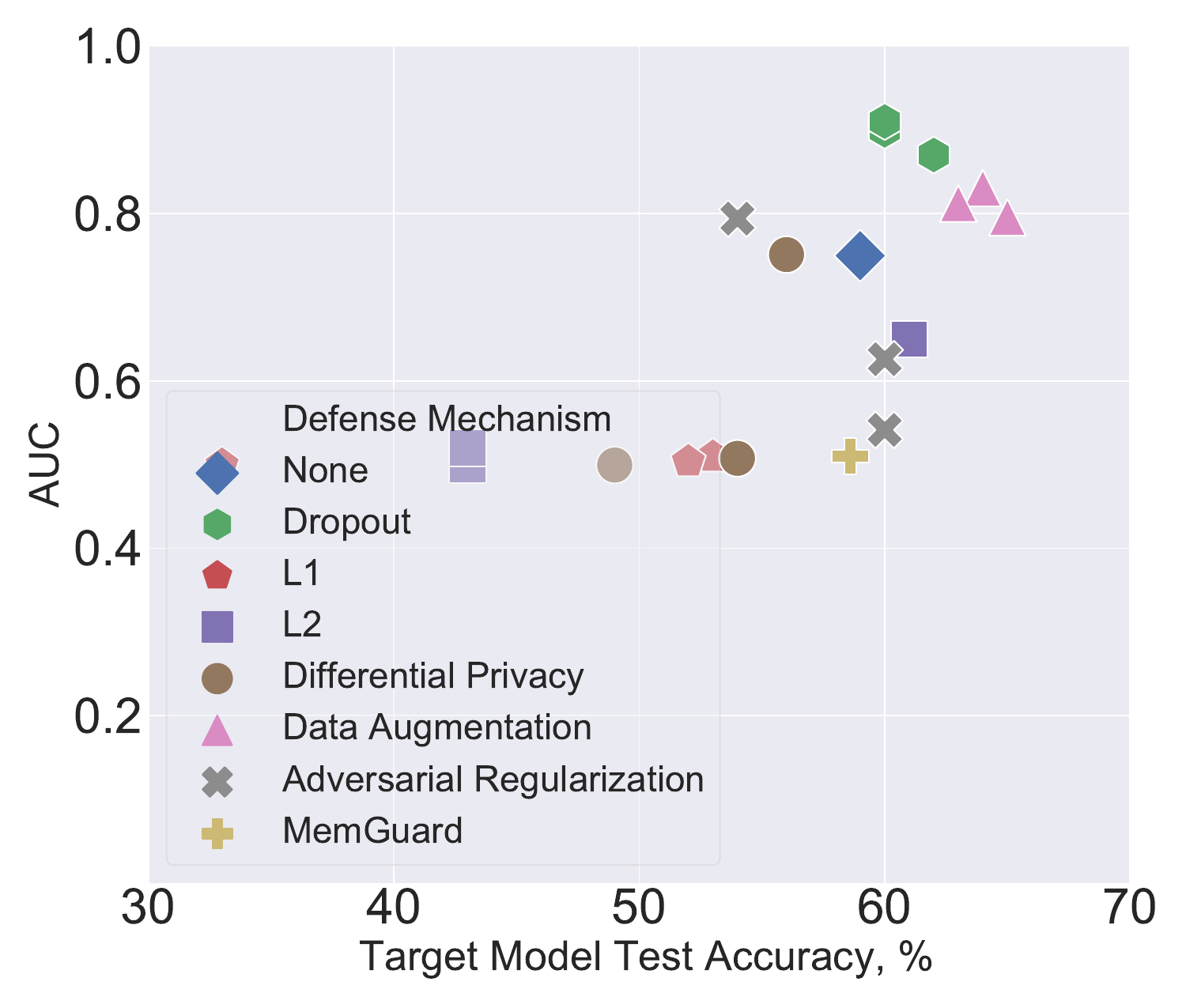}
\caption{Score-based attack}
\label{fig:score_defense} 
\end{subfigure}
\begin{subfigure}{0.5\columnwidth}
\includegraphics[width=\columnwidth]{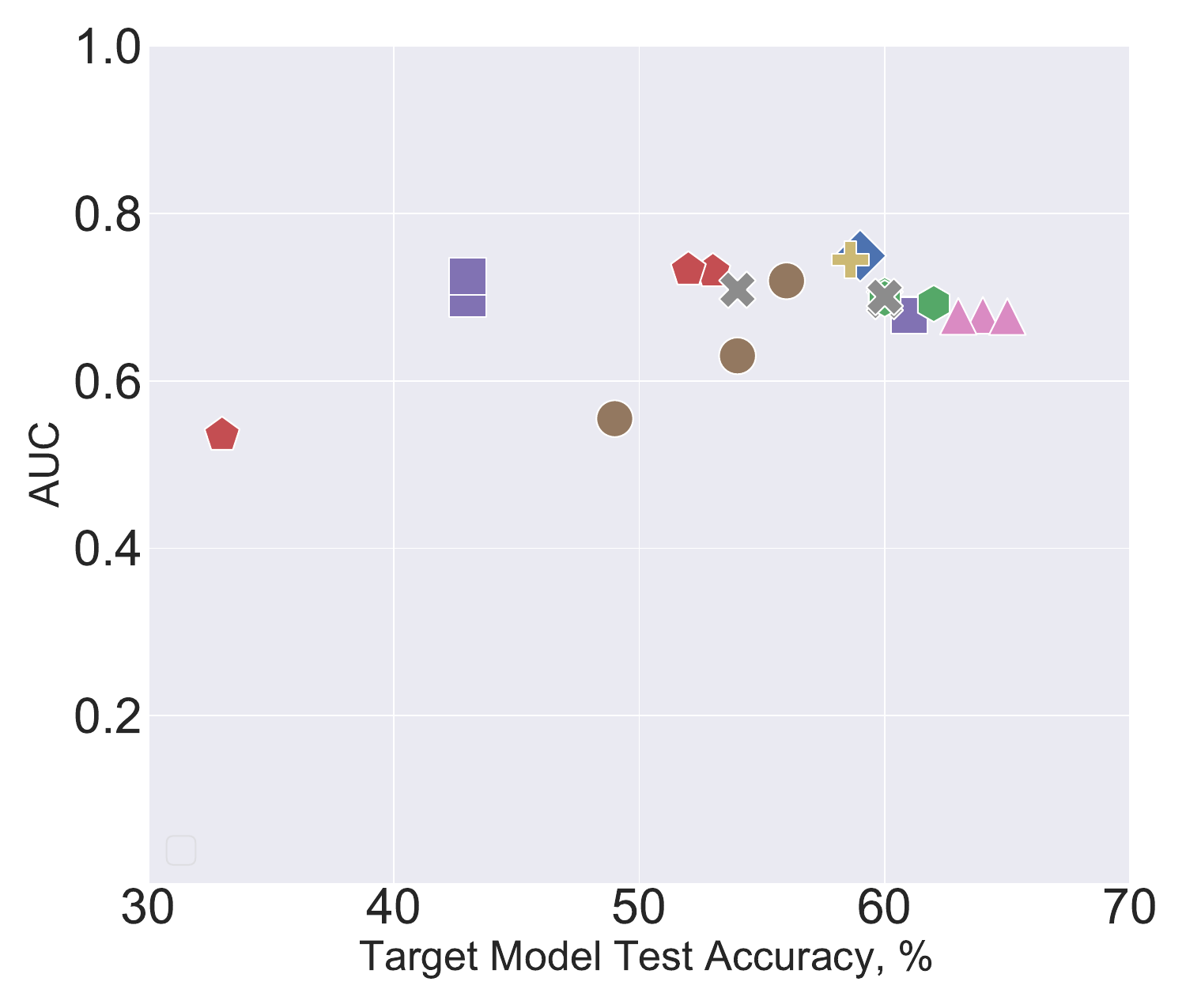}
\caption{Baseline attack}
\label{fig:baseline_defense} 
\end{subfigure}
\begin{subfigure}{0.5\columnwidth}
\includegraphics[width=\columnwidth]{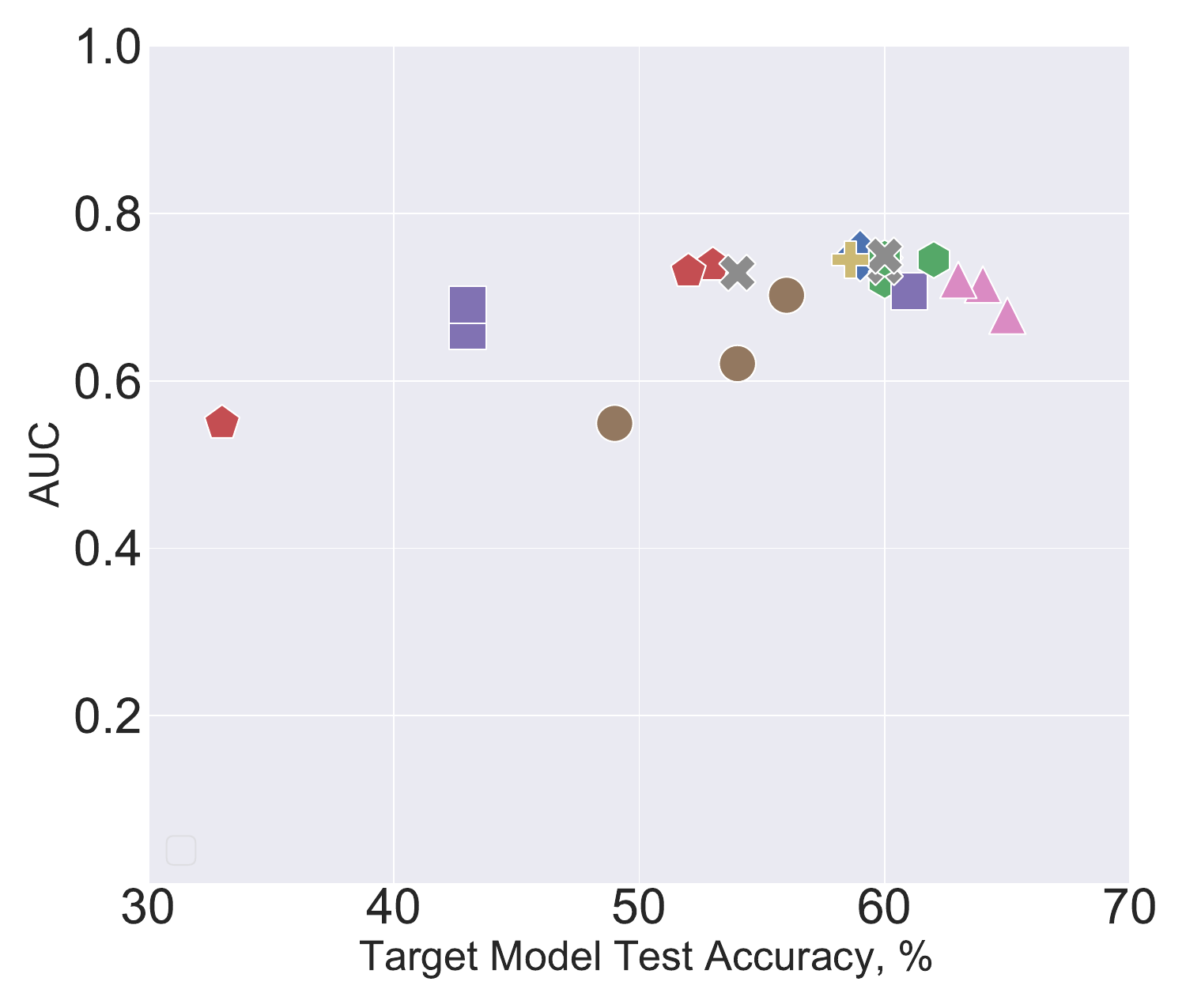}
\caption{Transfer attack}
\label{fig:transfer_defense} 
\end{subfigure}
\begin{subfigure}{0.5\columnwidth}
\includegraphics[width=\columnwidth]{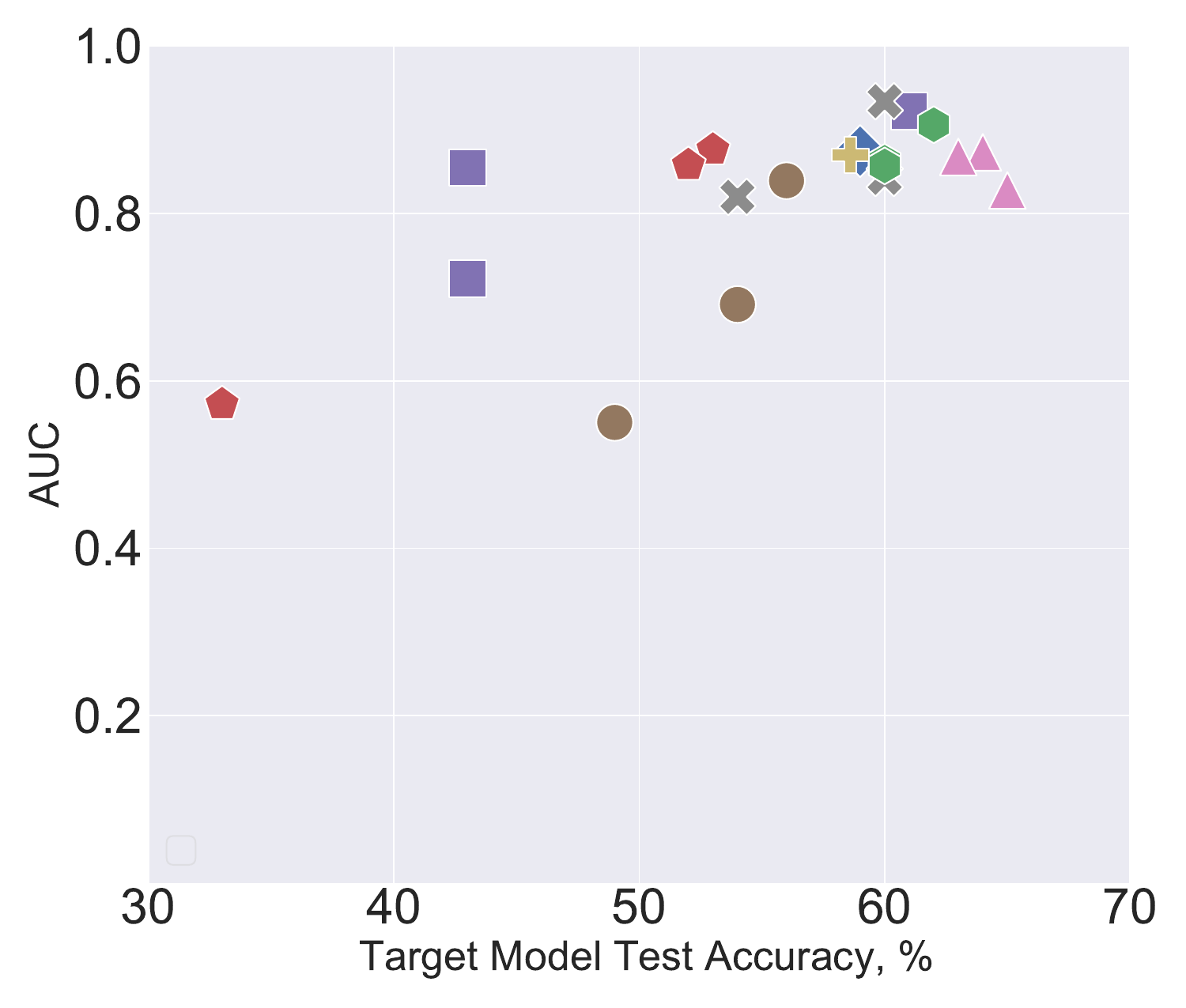}
\caption{Boundary attack}
\label{fig:boundary_defense}
\end{subfigure}
\caption{Attack AUC of transfer attack and boundary attack against multiple defense mechanisms.}
\label{fig:defense}
\end{figure*}

Furthermore, we also feed the equal member and non-member samples into each corresponding shadow model and obtain the ACR. 
Note that both member and non-member samples are never used to train the shadow model. 
We report the results in \autoref{table:shadow_acr}, and we can draw the same observations as for the target model. 
In other words, this again verifies our key intuition for transfer attack: The transferability of membership information holds between shadow model $\shadow$ and target model $\model$, i.e., the member and non-member samples behaving differently in $\model$ will also behave differently with high probability in $\shadow$.

% ======================================================
\subsection{Qualitative Analysis}
\label{subsec:QualitativeAnalysis}
% ======================================================

Next, we investigate the membership leakage of ML models from a visualization approach. 
We study the decision boundary of the target model (CIFAR-10, $\model$-3) with a given set of data samples, including 1,000 member samples and 1,000 non-member samples.
To better visualize the decision boundary, there are two points to note:
\begin{itemize}
\item Both member and non-member samples are mapped from the input space to the output space, which then presents the membership signal. 
Thus, we visualize the decision boundary in the output space, i.e., the transformed space of the last hidden layer which is fully connected with the final model decision.
\item Due to the limitation of the target dataset size, we further sample a large number of random data points in the output space and label them with different colors according to their corresponding classes. 
This can clearly visualize the decision boundary that distinguishes between different class regions. 
\end{itemize}
To this end, we map the given data samples into the transformed space and embed the output logits or scores into a 2D space using t-Distributed Stochastic Neighbor Embedding (t-SNE)~\cite{MH082}.
\autoref{fig:target_db} shows the results for 10 classes of CIFAR-10. 
We can see that the given data samples have been clearly classified into 10 classes and mapped to 10 different regions. 
For the sake of analysis, we purposely zoom in four different regions in the left of the whole space. 
From~\autoref{fig:target_db_zoomin}, we can make the following observations:
\begin{itemize}
\item The member samples and non-member samples belonging to the same class are tightly divided into 2 clusters, which explains why the previous score-based attacks can achieve effective performance.
\item More interestingly, we can see that the member samples are further away from the decision boundary than the non-member samples, that is, the distance between the members and the decision boundary is larger than that of the non-members. 
Again, this validates our key intuition.
\end{itemize}

Recall that in the decision change stage of boundary attack, we apply black-box adversarial attack techniques to change the final model decision. 
Here, we give an intuitive overview of how HopSkipJump and QEBA schemes work in \autoref{fig:attack_schema}. 
As we can see, though these two schemes adopt different strategies to find the perturbed sample, there is one thing in common: The search ends at the tangent samples between the neighboring $L_p$-radius ball of the original sample and its decision boundary. 
Only in this way they can mislead the target model and also generate a small perturbation. 
Combined with \autoref{fig:target_db_zoomin}, we can find that the magnitude of perturbation is essentially a reflection of the distance from the original sample to its decision boundary. 

We again feed the 1,000 member samples and 1,000 non-member samples to the shadow model (CIFAR-10, $\model$-3), and visualize its decision boundary in \autoref{fig:shadow_db_zoomin}. 
In particular, we mark in red the misclassified samples from non-members. 
First, looking at the correctly classified samples, we can also find that the member samples are relatively far from the decision boundary, i.e., the loss is relatively lower than that of non-member samples. 
As for the misclassified samples, it is easy to see that their loss is much larger than any other samples. 
Therefore, we can leverage the loss as metric to differentiate members and non-members. 
However, we should also note that compared to \autoref{fig:target_db_zoomin}, the difference between members and non-members towards the decision boundary is much smaller. 
Thus, if we do not adopt loss metric which considers the ground truth label, then the maximum confidence scores $Max(p_{i})$ and normalized entropy $\frac{-1}{\log (K)} \sum_{i} p_{i} \log \left(p_{i}\right)$ which are just based on self-information will lead to a much lower difference between members and non-members. 
This is the reason why the loss metric achieves the highest performance.

Summarizing the above quantitative and qualitative analysis, we verify our argument that the distance between the member sample and its decision boundary is larger than that of the non-member sample, thus revealing the reasons for the success of the membership inference, including score-based and decision-based attacks. 
In addition, we verify that membership information remains transferable between the target and shadow models.
Last but not least, we also show the reason why the loss metric of the transfer attack achieves the best performance.

% ======================================================
\section{Defenses Evaluation}
\label{sec:defenses}
% ======================================================

To mitigate the threat of membership leakage, a large body of defense mechanisms have been proposed in the literature. 
In this section, we evaluate the performance of current membership inference attacks against the state-of-the-art defenses. 
We summarize existing defenses in the following three broad categories.

\mypara{Generalization Enhancement}
As overfitting is the major reason for membership inference to be successful, multiple approaches have been proposed with the aim of reducing overfitting, which are first introduced by the machine learning community to encourage generalization. 
The standard generalization enhancement techniques, such as weight decay (L1/L2 regularization)~\cite{TLGYW18,SZHBFB19}, dropout~\cite{SHKSS14}, and data augmentation, have been shown to limit overfitting effectively, but may lead to a significant decrease in model accuracy.

\mypara{Privacy Enhancement}
Differential privacy~\cite{CMS11,DMNS06,INSTTW19} is a widely adopted for mitigating membership privacy. 
Many differential privacy based defense techniques add noise to the gradient to ensure the data privacy in the training process of the ML model. 
A representative approach in this category is DP-Adam~\cite{ACGMMTZ16}, and we adopt an open-source version of its implementation in our experiments.\footnote{\url{https://github.com/ebagdasa/pytorch-privacy}} 

\mypara{Confidence Score Perturbation}
Previous score-based attacks have demonstrated that the confidence score predicted by the target model clearly presents membership signal. 
Therefore, researchers have proposed several approaches to alter the confidence score. 
We focus on two representative approaches in this category: MemGuard~\cite{JSBZG19} and adversarial regularization~\cite{NSH18}, which changes the output probability distribution so that both members and non-members look like similar examples to the inference model built by the adversary. 
We adopt the original implementation of MemGuard,\footnote{\url{https://github.com/jjy1994/MemGuard}} and an open-source version of the adversarial regularization.\footnote{\url{https://github.com/SPIN-UMass/ML-Privacy-Regulization}}

For each mechanism, we train 3 target models (CIFAR-10, $\model-2$) using different hyper-parameters. 
For example, in L2 regularization, the $\lambda$ used to constrain the regularization loss is set to 0.01, 0.05, and 0.1, and the $\lambda$ in L1 regularization is set to 0.0001, 0.001 and 0.005, respectively. 
In differential privacy, the noise  is randomly sampled from a Gaussian distribution $\mathcal{N}(\epsilon ,\beta)$, wherein $\epsilon$ is fixed to 0 and $\beta$ is set to 0.1, 0.5 and 1.1, respectively.
\begin{table}[!ht]
\centering
\caption{Attack AUC performance under the defense of MemGuard.}
\scalebox{0.83}
{
\label{tab:defense_memguard}
\begin{tabular}{l|cc|cc}
\toprule
&\multicolumn{2}{c}{CIFAR-10, $\model$-2} &  \multicolumn{2}{c}{ Face, $\model$-2}         \\
 Attack &None & MemGuard &None & MemGuard \\
\midrule
score-based     &  0.8655  &     0.5151                     & 0.755  & 0.513 \\
baseline attack &  0.705   &     0.705                      & 0.665  & 0.665 \\
transfer attack &  0.7497  &     0.7497                     & 0.6664 & 0.6664 \\
boundary attack &  0.8747  &     0.8747                     & 0.8617 & 0.8617 \\
\bottomrule
\end{tabular}
}
\end{table}

We report the attack performance against models trained with a wide variety of different defensive mechanisms in \autoref{fig:defense}, and we make the following observations.
\begin{itemize}
\item Our decision-based attacks. i.e., both transfer attack and boundary attack, can bypass most types of defense mechanisms. 
\item  Strong differential privacy ($\beta$=1.1), L1 regularization ($\lambda=0.005$) and L2 regularization ($\lambda=0.1$) can reduce membership leakage but, as expected, lead to a significant degradation in the model’s accuracy. 
The reason is that the decision boundary between members and non-members is heavily blurred.
\item Data augmentation can definitely reduce overfitting, but it still does not reduce membership leakage. 
This is because data augmentation drives the model to strongly remember both the original samples and their augmentations.
\end{itemize}
In \autoref{tab:defense_memguard}, we further compare the performance of all attacks against MemGuard~\cite{JSBZG19}, which is the latest powerful defense technique and can be easily deployed. 
We can find that MemGuard cannot defend against decision-based attacks at all, but is very effective against previous score-based attacks. 

% ======================================================
\section{Related Works}
\label{sec:related}
% ======================================================

Various research has shown that machine learning models are vulnerable to security and privacy attacks. 
In this section, we mainly survey the domains that are most relevant to us.

\mypara{Membership Inference} 
Membership inference attack has been successfully performed in various data domains, ranging form biomedical data~\cite{BBHM16,HZHBTWB19,HSRDTMPSNC08} to mobility traces~\cite{PTC18}. 
Shokri et al.~\cite{SSSS17} present the first membership inference attack against machine learning models. 
The general idea behind this attack is to use multiple shadow models to generate data to train multiple attack models (one for each class). 
These attack models take the target sample’s confidence scores as input and output its membership status, i.e., member or non-member. 
Salem et al.~\cite{SZHBFB19} later present another attack by gradually relaxing the assumptions made by Shokri et al.~\cite{SSSS17} achieving a model and data independent membership inference. 
In addition, there are several other subsequent score-based membership inference attacks~\cite{LBG17,YGFJ18,SSM19,HYYBGC21,LLR21}.
In the area of decision-based attacks, Yeom et al.~\cite{YGFJ18} quantitatively analyzed the relationship between attack performance and loss for training and testing sets, and proposed the first decision-based attack, i.e., baseline attack aforementioned. 
We also acknowledge that a concurrent work~\cite{CTCP20} proposes an approach similar to our boundary attack. 
Specifically, the concurrent work assumes that an adversary has more knowledge of the target model, including training knowledge (model architecture, training algorithm, and training dataset size), and a shadow dataset from the same distribution as the target dataset to estimate the threshold. 
In our work, we relax all assumptions and propose a general threshold-choosing method. 
We further present a new perspective on the reasons for the success of membership inference. In addition, we introduce a novel transfer-attack.

\mypara{Defenses Against Membership Inference}
Researchers have proposed to improve privacy against membership inference via different types of generalization enhancement. 
For example, Shokri et al.~\cite{SSSS17} adopted L2 regularization with a polynomial in the model’s loss function to penalize large parameters. 
Salem et al.~\cite{SZHBFB19} demonstrated two effective method of defending MI attacks, namely dropout and model stacking. 
Nasr et al.~\cite{NSH18} introduced a defensive confidence score membership classifier in a min-max game mechanism to train models with membership privacy, namely adversarial regularization. 
There are other existing generalization enhancement method can be used to mitigate membership leakage, such as L1 regularization and data augmentation. 
Another direction is privacy enhancement. 
Many differential privacy-based defenses ~\cite{CMS11,DMNS06,INSTTW19} involve clipping and adding noise to instance-level gradients and is designed to train a model to prevent it from memorizing training data or being susceptible to membership leakage. 
Shokri et al.~\cite{SSSS17} designed a differential privacy method for collaborative learning of DNNs. 
As for confidence score alteration, Jia et al.~\cite{JSBZG19} introduce MemGuard, the first defense with formal utility-loss guarantees against membership inference. 
The basic idea behind this work is to add carefully crafted noise to confidence scores of an ML model to mislead the membership classifier. 
Yang et al.~\cite{YSXCZ20} also propose a similar defense in this direction.

\mypara{Attacks against Machine Learning} 
Besides membership inference attacks, there exist numerous other types of attacks against ML models.
A major attack type in this space is adversarial examples~\cite{PMSW18,PMGJCS17,PMJFCS16,TKPGBM17,CJW20,LXZYL20}. 
In this setting, an adversary adds carefully crafted noise to samples aiming at mislead the target classifier. 
Ganju et al.~\cite{GWYGB18} proposed a property inference attack aiming at inferring general properties of the training data (such as the proportion of each class in the training data). 
Model inversion attack~\cite{FJR15,FLJLPR14} focuses on inferring the missing attributes of the target ML model. 
A similar type of attacks is backdoor attack, where the adversary as a model trainer embeds a trigger into the model for her to exploit when the model is deployed~\cite{GDG17,LMALZWZ19,WYSLVZZ19}. 
Another line of work is model stealing, Tramèr et al.~\cite{TZJRR16} proposed the first attack on inferring a model’s parameters. Other works focus on protecting a model’s ownership~\cite{LHZG19,ZGJWSHM18,ABCPK18,RCK18}.

% ======================================================
\section{Conclusion}
\label{sec:con}
% ======================================================

In this paper, we perform a systematic investigation on membership leakage in label-only exposures of ML models, and propose two novel decision-based membership inference attacks, including transfer attack and boundary attack. 
Extensive experiments demonstrate that our two attacks achieve better performances than baseline attack, and even outperform prior score-based attacks in some cases. 
Furthermore, we propose a new perspective on the reasons for the success of membership inference and show that members samples are further away from the decision boundary than non-members. 
Finally, we evaluate multiple defense mechanisms against our decision-based attacks and  show that our novel attacks can still achieve reasonable performance unless heavy regularization has been applied. 
In particular, our evaluation demonstrates that confidence score perturbation is an infeasible defense mechanism in label-only exposures.

% ======================================================
\section*{Acknowledgements}
This work is partially funded by the Helmholtz Association within the project ``Trustworthy Federated Data Analytics'' (TFDA) (funding number ZT-I-OO1 4).
% ======================================================

% ======================================================
\newpage
\balance
\bibliographystyle{plain}
\bibliography{normal_generated_py3}

\begin{thebibliography}{10}

\bibitem{CIFAR}
\url{https://www.cs.toronto.edu/~kriz/cifar.html}.

\bibitem{GTSRB}
\url{http://benchmark.ini.rub.de/?section=gtsrb}.

\bibitem{FACE}
\url{http://vis-www.cs.umass.edu/lfw/}.

\bibitem{ACGMMTZ16}
Martin Abadi, Andy Chu, Ian Goodfellow, Brendan McMahan, Ilya Mironov, Kunal
  Talwar, and Li~Zhang.
\newblock {Deep Learning with Differential Privacy}.
\newblock In {\em {ACM SIGSAC Conference on Computer and Communications
  Security (CCS)}}, pages 308--318. ACM, 2016.

\bibitem{ABCPK18}
Yossi Adi, Carsten Baum, Moustapha Cisse, Benny Pinkas, and Joseph Keshet.
\newblock {Turning Your Weakness Into a Strength: Watermarking Deep Neural
  Networks by Backdooring}.
\newblock In {\em {USENIX Security Symposium (USENIX Security)}}, pages
  1615--1631. USENIX, 2018.

\bibitem{BBHM16}
Michael Backes, Pascal Berrang, Mathias Humbert, and Praveen Manoharan.
\newblock {Membership Privacy in MicroRNA-based Studies}.
\newblock In {\em {ACM SIGSAC Conference on Computer and Communications
  Security (CCS)}}, pages 319--330. ACM, 2016.

\bibitem{BHPZ17}
Michael Backes, Mathias Humbert, Jun Pang, and Yang Zhang.
\newblock {walk2friends: Inferring Social Links from Mobility Profiles}.
\newblock In {\em {ACM SIGSAC Conference on Computer and Communications
  Security (CCS)}}, pages 1943--1957. ACM, 2017.

\bibitem{BCMNSLGR13}
Battista Biggio, Igino Corona, Davide Maiorca, Blaine Nelson, Nedim Srndic,
  Pavel Laskov, Giorgio Giacinto, and Fabio Roli.
\newblock {Evasion Attacks against Machine Learning at Test Time}.
\newblock In {\em {European Conference on Machine Learning and Principles and
  Practice of Knowledge Discovery in Databases (ECML/PKDD)}}, pages 387--402.
  Springer, 2013.

\bibitem{BFDB11}
Philippe Burlina, David~E. Freund, B.~Dupas, and Neil~M. Bressler.
\newblock {Automatic Screening of Age-related Macular Degeneration and Retinal
  Abnormalities}.
\newblock In {\em {Annual International Conference of the {IEEE} Engineering in
  Medicine and Biology Society (EMBC)}}, pages 3962--3966. IEEE, 2011.

\bibitem{CW17}
Nicholas Carlini and David Wagner.
\newblock {Towards Evaluating the Robustness of Neural Networks}.
\newblock In {\em {IEEE Symposium on Security and Privacy (S\&P)}}, pages
  39--57. IEEE, 2017.

\bibitem{CMS11}
Kamalika Chaudhuri, Claire Monteleoni, and Anand~D Sarwate.
\newblock {Differentially Private Empirical Risk Minimization}.
\newblock {\em {Journal of Machine Learning Research}}, 2011.

\bibitem{CJW20}
Jianbo Chen, Michael~I. Jordan, and Martin~J. Wainwright.
\newblock {HopSkipJumpAttack: {A} Query-Efficient Decision-Based Attack}.
\newblock In {\em {IEEE Symposium on Security and Privacy (S\&P)}}, pages
  1277--1294. IEEE, 2020.

\bibitem{CTCP20}
Christopher A.~Choquette Choo, Florian Tram{\`e}r, Nicholas Carlini, and
  Nicolas Papernot.
\newblock {Label-Only Membership Inference Attacks}.
\newblock {\em {CoRR abs/2007.14321}}, 2020.

\bibitem{CRK19}
Jeremy~M. Cohen, Elan Rosenfeld, and J.~Zico Kolter.
\newblock {Certified Adversarial Robustness via Randomized Smoothing}.
\newblock In {\em {International Conference on Machine Learning (ICML)}}, pages
  1310--1320. PMLR, 2019.

\bibitem{DMPJBONR19}
Ambra Demontis, Marco Melis, Maura Pintor, Matthew Jagielski, Battista Biggio,
  Alina Oprea, Cristina Nita{-}Rotaru, and Fabio Roli.
\newblock {Why Do Adversarial Attacks Transfer? Explaining Transferability of
  Evasion and Poisoning Attacks}.
\newblock In {\em {USENIX Security Symposium (USENIX Security)}}, pages
  321--338. USENIX, 2019.

\bibitem{MH082}
Laurens~Van der Maaten and Geoffrey Hinton.
\newblock {Visualizing Data Using t-SNE}.
\newblock {\em {Journal of Machine Learning Research}}, 2008.

\bibitem{DMNS06}
Cynthia Dwork, Frank McSherry, Kobbi Nissim, and Adam Smith.
\newblock {Calibrating Noise to Sensitivity in Private Data Analysis}.
\newblock In {\em {Theory of Cryptography Conference (TCC)}}, pages 265--284.
  Springer, 2006.

\bibitem{FJR15}
Matt Fredrikson, Somesh Jha, and Thomas Ristenpart.
\newblock {Model Inversion Attacks that Exploit Confidence Information and
  Basic Countermeasures}.
\newblock In {\em {ACM SIGSAC Conference on Computer and Communications
  Security (CCS)}}, pages 1322--1333. ACM, 2015.

\bibitem{FLJLPR14}
Matt Fredrikson, Eric Lantz, Somesh Jha, Simon Lin, David Page, and Thomas
  Ristenpart.
\newblock {Privacy in Pharmacogenetics: An End-to-End Case Study of
  Personalized Warfarin Dosing}.
\newblock In {\em {USENIX Security Symposium (USENIX Security)}}, pages 17--32.
  USENIX, 2014.

\bibitem{GWYGB18}
Karan Ganju, Qi~Wang, Wei Yang, Carl~A. Gunter, and Nikita Borisov.
\newblock {Property Inference Attacks on Fully Connected Neural Networks using
  Permutation Invariant Representations}.
\newblock In {\em {ACM SIGSAC Conference on Computer and Communications
  Security (CCS)}}, pages 619--633. ACM, 2018.

\bibitem{GDG17}
Tianyu Gu, Brendan Dolan-Gavitt, and Siddharth Grag.
\newblock {Badnets: Identifying Vulnerabilities in the Machine Learning Model
  Supply Chain}.
\newblock {\em {CoRR abs/1708.06733}}, 2017.

\bibitem{HZHBTWB19}
Inken Hagestedt, Yang Zhang, Mathias Humbert, Pascal Berrang, Haixu Tang,
  XiaoFeng Wang, and Michael Backes.
\newblock {MBeacon: Privacy-Preserving Beacons for DNA Methylation Data}.
\newblock In {\em {Network and Distributed System Security Symposium (NDSS)}}.
  Internet Society, 2019.

\bibitem{HHB19}
Benjamin Hilprecht, Martin H{\"{a}}rterich, and Daniel Bernau.
\newblock {Monte Carlo and Reconstruction Membership Inference Attacks against
  Generative Models}.
\newblock {\em {Symposium on Privacy Enhancing Technologies Symposium}}, 2019.

\bibitem{HSRDTMPSNC08}
Nils Homer, Szabolcs Szelinger, Margot Redman, David Duggan, Waibhav Tembe,
  Jill Muehling, John~V. Pearson, Dietrich~A. Stephan, Stanley~F. Nelson, and
  David~W. Craig.
\newblock {Resolving Individuals Contributing Trace Amounts of DNA to Highly
  Complex Mixtures Using High-Density SNP Genotyping Microarrays}.
\newblock {\em {PLOS Genetics}}, 2008.

\bibitem{HYYBGC21}
Bo~Hui, Yuchen Yang, Haolin Yuan, Philippe Burlina, Neil~Zhenqiang Gong, and
  Yinzhi Cao.
\newblock {Practical Blind Membership Inference Attack via Differential
  Comparisons}.
\newblock In {\em {Network and Distributed System Security Symposium (NDSS)}}.
  Internet Society, 2021.

\bibitem{INSTTW19}
Roger Iyengar, Joseph~P. Near, Dawn~Xiaodong Song, Om~Dipakbhai Thakkar,
  Abhradeep Thakurta, and Lun Wang.
\newblock {Towards Practical Differentially Private Convex Optimization}.
\newblock In {\em {IEEE Symposium on Security and Privacy (S\&P)}}, pages
  299--316. IEEE, 2019.

\bibitem{JSBZG19}
Jinyuan Jia, Ahmed Salem, Michael Backes, Yang Zhang, and Neil~Zhenqiang Gong.
\newblock {MemGuard: Defending against Black-Box Membership Inference Attacks
  via Adversarial Examples}.
\newblock In {\em {ACM SIGSAC Conference on Computer and Communications
  Security (CCS)}}, pages 259--274. ACM, 2019.

\bibitem{KSMB16}
Ira Kemelmacher{-}Shlizerman, Steven~M. Seitz, Daniel Miller, and Evan
  Brossard.
\newblock {The MegaFace Benchmark: 1 Million Faces for Recognition at Scale}.
\newblock In {\em {IEEE Conference on Computer Vision and Pattern Recognition
  (CVPR)}}, pages 4873--4882. IEEE, 2016.

\bibitem{KEEKF15}
Konstantina Kourou, Themis~P. Exarchos, Konstantinos~P. Exarchos, Michalis~V.
  Karamouzis, and Dimitrios~I. Fotiadis.
\newblock {Machine Learning Applications in Cancer Prognosis and Prediction}.
\newblock {\em {Computational and Structural Biotechnology Journal}}, 2015.

\bibitem{LXZYL20}
Huichen Li, Xiaojun Xu, Xiaolu Zhang, Shuang Yang, and Bo~Li.
\newblock {{QEBA:} Query-Efficient Boundary-Based Blackbox Attack}.
\newblock In {\em {IEEE Conference on Computer Vision and Pattern Recognition
  (CVPR)}}, pages 1218--1227. IEEE, 2020.

\bibitem{LLR21}
Jiacheng Li, Ninghui Li, and Bruno Ribeiro.
\newblock {Membership Inference Attacks and Defenses in Supervised Learning via
  Generalization Gap}.
\newblock In {\em {ACM Conference on Data and Application Security and Privacy
  (CODASPY)}}, pages 5--16. ACM, 2021.

\bibitem{LHZG19}
Zheng Li, Chengyu Hu, Yang Zhang, and Shanqing Guo.
\newblock {How to Prove Your Model Belongs to You: A Blind-Watermark based
  Framework to Protect Intellectual Property of DNN}.
\newblock In {\em {Annual Computer Security Applications Conference (ACSAC)}},
  pages 126--137. ACM, 2019.

\bibitem{LCLS16}
Yanpei Liu, Xinyun Chen, Chang Liu, and Dawn Song.
\newblock {Delving into Transferable Adversarial Examples and Black-box
  Attacks}.
\newblock {\em {CoRR abs/1611.02770}}, 2016.

\bibitem{LMALZWZ19}
Yingqi Liu, Shiqing Ma, Yousra Aafer, Wen-Chuan Lee, Juan Zhai, Weihang Wang,
  and Xiangyu Zhang.
\newblock {Trojaning Attack on Neural Networks}.
\newblock In {\em {Network and Distributed System Security Symposium (NDSS)}}.
  Internet Society, 2019.

\bibitem{LBG17}
Yunhui Long, Vincent Bindschaedler, and Carl~A. Gunter.
\newblock {Towards Measuring Membership Privacy}.
\newblock {\em {CoRR abs/1712.09136}}, 2017.

\bibitem{LBWBWTGC18}
Yunhui Long, Vincent Bindschaedler, Lei Wang, Diyue Bu, Xiaofeng Wang, Haixu
  Tang, Carl~A. Gunter, and Kai Chen.
\newblock {Understanding Membership Inferences on Well-Generalized Learning
  Models}.
\newblock {\em {CoRR abs/1802.04889}}, 2018.

\bibitem{NKKKP19}
Muzammal Naseer, Salman~H. Khan, Muhammad~Haris Khan, Fahad~Shahbaz Khan, and
  Fatih Porikli.
\newblock {Cross-Domain Transferability of Adversarial Perturbations}.
\newblock In {\em {Annual Conference on Neural Information Processing Systems
  (NeurIPS)}}, pages 12885--12895. NeurIPS, 2019.

\bibitem{NSH18}
Milad Nasr, Reza Shokri, and Amir Houmansadr.
\newblock {Machine Learning with Membership Privacy using Adversarial
  Regularization}.
\newblock In {\em {ACM SIGSAC Conference on Computer and Communications
  Security (CCS)}}, pages 634--646. ACM, 2018.

\bibitem{PMG16}
Nicolas Papernot, Patrick McDaniel, and Ian Goodfellow.
\newblock {Transferability in Machine Learning: from Phenomena to Black-Box
  Attacks using Adversarial Samples}.
\newblock {\em {CoRR abs/1605.07277}}, 2016.

\bibitem{PMSW18}
Nicolas Papernot, Patrick McDaniel, Arunesh Sinha, and Michael Wellman.
\newblock {SoK: Towards the Science of Security and Privacy in Machine
  Learning}.
\newblock In {\em {IEEE European Symposium on Security and Privacy (Euro
  S\&P)}}, pages 399--414. IEEE, 2018.

\bibitem{PMGJCS17}
Nicolas Papernot, Patrick~D. McDaniel, Ian Goodfellow, Somesh Jha, Z.~Berkay
  Celik, and Ananthram Swami.
\newblock {Practical Black-Box Attacks Against Machine Learning}.
\newblock In {\em {ACM Asia Conference on Computer and Communications Security
  (ASIACCS)}}, pages 506--519. ACM, 2017.

\bibitem{PMJFCS16}
Nicolas Papernot, Patrick~D. McDaniel, Somesh Jha, Matt Fredrikson, Z.~Berkay
  Celik, and Ananthram Swami.
\newblock {The Limitations of Deep Learning in Adversarial Settings}.
\newblock In {\em {IEEE European Symposium on Security and Privacy (Euro
  S\&P)}}, pages 372--387. IEEE, 2016.

\bibitem{PTC18}
Apostolos Pyrgelis, Carmela Troncoso, and Emiliano~De Cristofaro.
\newblock {Knock Knock, Who's There? Membership Inference on Aggregate Location
  Data}.
\newblock In {\em {Network and Distributed System Security Symposium (NDSS)}}.
  Internet Society, 2018.

\bibitem{RCK18}
Bita~Darvish Rouhani, Huili Chen, and Farinaz Koushanfar.
\newblock {DeepSigns: A Generic Watermarking Framework for IP Protection of
  Deep Learning Models}.
\newblock {\em {CoRR abs/1804.00750}}, 2018.

\bibitem{SDSOJ19}
Alexandre Sablayrolles, Matthijs Douze, Cordelia Schmid, Yann Ollivier, and
  Herv{\'e} J{\'e}gou.
\newblock {White-box vs Black-box: Bayes Optimal Strategies for Membership
  Inference}.
\newblock In {\em {International Conference on Machine Learning (ICML)}}, pages
  5558--5567. PMLR, 2019.

\bibitem{SZHBFB19}
Ahmed Salem, Yang Zhang, Mathias Humbert, Pascal Berrang, Mario Fritz, and
  Michael Backes.
\newblock {ML-Leaks: Model and Data Independent Membership Inference Attacks
  and Defenses on Machine Learning Models}.
\newblock In {\em {Network and Distributed System Security Symposium (NDSS)}}.
  Internet Society, 2019.

\bibitem{SHNSSDG18}
Ali Shafahi, W~Ronny Huang, Mahyar Najibi, Octavian Suciu, Christoph Studer,
  Tudor Dumitras, and Tom Goldstein.
\newblock {Poison Frogs! Targeted Clean-Label Poisoning Attacks on Neural
  Networks}.
\newblock In {\em {Annual Conference on Neural Information Processing Systems
  (NeurIPS)}}, pages 6103--6113. NeurIPS, 2018.

\bibitem{SSSS17}
Reza Shokri, Marco Stronati, Congzheng Song, and Vitaly Shmatikov.
\newblock {Membership Inference Attacks Against Machine Learning Models}.
\newblock In {\em {IEEE Symposium on Security and Privacy (S\&P)}}, pages
  3--18. IEEE, 2017.

\bibitem{SSM19}
Liwei Song, Reza Shokri, and Prateek Mittal.
\newblock {Privacy Risks of Securing Machine Learning Models against
  Adversarial Examples}.
\newblock In {\em {ACM SIGSAC Conference on Computer and Communications
  Security (CCS)}}, pages 241--257. ACM, 2019.

\bibitem{SHKSS14}
Nitish Srivastava, Geoffrey Hinton, Alex Krizhevsky, Ilya Sutskever, and Ruslan
  Salakhutdinov.
\newblock {Dropout: A Simple Way to Prevent Neural Networks from Overfitting}.
\newblock {\em {Journal of Machine Learning Research}}, 2014.

\bibitem{SWFJH10}
Mary~H. Stanfill, Margaret Williams, Susan~H. Fenton, Robert~A. Jenders, and
  William~R. Hersh.
\newblock {A Systematic Literature Review of Automated Clinical Coding and
  Classification Systems}.
\newblock {\em {J. Am. Medical Informatics Assoc.}}, 2010.

\bibitem{TKPGBM17}
Florian Tram{\`e}r, Alexey Kurakin, Nicolas Papernot, Ian Goodfellow, Dan
  Boneh, and Patrick McDaniel.
\newblock {Ensemble Adversarial Training: Attacks and Defenses}.
\newblock In {\em {International Conference on Learning Representations
  (ICLR)}}, 2017.

\bibitem{TZJRR16}
Florian Tram{\`e}r, Fan Zhang, Ari Juels, Michael~K. Reiter, and Thomas
  Ristenpart.
\newblock {Stealing Machine Learning Models via Prediction APIs}.
\newblock In {\em {USENIX Security Symposium (USENIX Security)}}, pages
  601--618. USENIX, 2016.

\bibitem{TLGYW18}
Stacey Truex, Ling Liu, Mehmet~Emre Gursoy, Lei Yu, and Wenqi Wei.
\newblock {Towards Demystifying Membership Inference Attacks}.
\newblock {\em {CoRR abs/1807.09173}}, 2018.

\bibitem{WYSLVZZ19}
Bolun Wang, Yuanshun Yao, Shawn Shan, Huiying Li, Bimal Viswanath, Haitao
  Zheng, and Ben~Y. Zhao.
\newblock {Neural Cleanse: Identifying and Mitigating Backdoor Attacks in
  Neural Networks}.
\newblock In {\em {IEEE Symposium on Security and Privacy (S\&P)}}, pages
  707--723. IEEE, 2019.

\bibitem{YSXCZ20}
Ziqi Yang, Bin Shao, Bohan Xuan, Ee-Chien Chang, and Fan Zhang.
\newblock {Defending Model Inversion and Membership Inference Attacks via
  Prediction Purification}.
\newblock {\em {CoRR abs/2005.03915}}, 2020.

\bibitem{YGFJ18}
Samuel Yeom, Irene Giacomelli, Matt Fredrikson, and Somesh Jha.
\newblock {Privacy Risk in Machine Learning: Analyzing the Connection to
  Overfitting}.
\newblock In {\em {IEEE Computer Security Foundations Symposium (CSF)}}, pages
  268--282. IEEE, 2018.

\bibitem{ZDHZGRHW20}
Runtian Zhai, Chen Dan, Di~He, Huan Zhang, Boqing Gong, Pradeep Ravikumar,
  Cho-Jui Hsieh, and Liwei Wang.
\newblock {MACER: Attack-free and Scalable Robust Training via Maximizing
  Certified Radius}.
\newblock In {\em {International Conference on Learning Representations
  (ICLR)}}, 2020.

\bibitem{ZGJWSHM18}
Jialong Zhang, Zhongshu Gu, Jiyong Jang, Hui Wu, Marc~Ph. Stoecklin, Heqing
  Huang, and Ian Molloy.
\newblock {Protecting Intellectual Property of Deep Neural Networks with
  Watermarking}.
\newblock In {\em {ACM Asia Conference on Computer and Communications Security
  (ASIACCS)}}, pages 159--172. ACM, 2018.

\bibitem{ZJPWLS20}
Yuheng Zhang, Ruoxi Jia, Hengzhi Pei, Wenxiao Wang, Bo~Li, and Dawn Song.
\newblock {The Secret Revealer: Generative Model-Inversion Attacks Against Deep
  Neural Networks}.
\newblock In {\em {IEEE Conference on Computer Vision and Pattern Recognition
  (CVPR)}}, pages 250--258. IEEE, 2020.

\bibitem{ZDH17}
Tianyue Zheng, Weihong Deng, and Jiani Hu.
\newblock {Cross-Age {LFW:} {A} Database for Studying Cross-Age Face
  Recognition in Unconstrained Environments}.
\newblock {\em {CoRR abs/1708.08197}}, 2017.

\end{thebibliography}
% ======================================================

% ======================================================
\appendix
% ======================================================

% ======================================================
\newpage
\section{Appendix}
% ======================================================

% ======================================================
\subsection{Datasets Description}
\label{appendix:datasets}
% ======================================================

\mypara{CIFAR-10/CIFAR-100}
CIFAR-10~\cite{CIFAR} and CIFAR-100~\cite{CIFAR} are benchmark datasets used to evaluate image recognition algorithms. 
CIFAR-10 is composed of 32$\times$32 color images in 10 classes, with 6000 images per class. 
In total, there are 50000 training images and 10000 test images. 
CIFAR-100 has the same format as CIFAR-10, but it has 100 classes containing 600 images each. 
There are 500 training images and 100 testing images per class. 

\mypara{GTSRB}
The GTSRB~\cite{GTSRB} dataset is an image collection consisting of 43 traffic signs. Images vary in size and are RGB-encoded.  
It consists of over 51,839 color images, whose dimensions range from 15$\times$15 to 250$\times$250 pixels (not necessarily square). 
Of these 51,839 images, 39,209 are used for training, and 12,630 are used for testing. 
Due to the varying sizes of the images, the images are resized to 64$\times$64 before being passed to the model for classification.

\mypara{Face}
The Face~\cite{FACE} dataset consists of about 13,000 images of human faces crawled from the web. 
It is collected from 1,680 participants with each participant having at least two distinct images in the dataset. 
In our evaluation, we only consider people with more than 40 images, which leaves us with 19 people’s data, i.e., 19 classes. 
The Face dataset is challenging for facial recognition, as the images are taken from the web and not under a controlled environment, such as a lab. 
It is also worth noting that this dataset is unbalanced.

% ======================================================
\subsection{Certified Radius}\label{app:CR}
% ======================================================

\mypara{Randomized Smoothing}
In this work, we apply a recent technique, called randomized smoothing ~\cite{CRK19}, which can be extended to any architecture to obtain the certified radius of smoothed deep neural networks. 
The core of randomized smoothing is to use the smoothed version of $\model$, which is denoted by $\smodel$, to make predictions. 
The formulation of $\smodel$ is defined as follows.

\mypara{Definition 1}
For an arbitrary classifier $\model$ and $\sigma > 0$, the smoothed classifier $\smodel$ of $\model$ is defined as
\begin{align}
\label{equ:smoothed}
\smodel(x)=\argmax_{c\in \mathcal{Y}}P_{\varepsilon \sim \mathcal{N}(0, \sigma^2 \boldsymbol{I})}(\model(x+\varepsilon)=c).
\end{align}
In short, the smoothed classifier $\smodel$ returns the label most likely to be returned by $\model$ when its input is sampled from a Gaussian distribution $\mathcal{N}(x,\sigma^2 \boldsymbol{I})$ centered at $x$. 
Cohen et al.~\cite{CRK19} prove the following theorem, which provides an analytic form of certified radius:

\mypara{Theorem 1}~\cite{CRK19}
\label{thm:CR}
Let $\model: x \rightarrow y$, and $\varepsilon \sim \mathcal{N}(0,\sigma^2 \boldsymbol{I})$. Let the smoothed classifier $\smodel$ be defined as in (\ref{equ:smoothed}). Let the ground truth of an input $x$ be $y$. If $\smodel$ classifies $x$ correctly, i.e.,
\begin{align}
P_\varepsilon(\model(x+\varepsilon)=y) \geq \max_{y'\neq y}P_\varepsilon(\model(x+\varepsilon)=y').
\end{align}
Then, $\smodel$ is provably robust at $x$, with the certified radius given by 

\begin{align}
CR(\smodel;x,y) = &\frac{\sigma}{2}[\Phi^{-1}(P_\varepsilon(\model(x+\varepsilon)=y))\nonumber\\
&-\Phi^{-1}(\max_{y' \neq y}P_\varepsilon(\model(x+\varepsilon)=y'))] \nonumber\\ 
= &\frac{\sigma}{2}[\Phi^{-1}(\mathbb{E}_\varepsilon \mathbf{1}_{\{\model(x+\varepsilon)=y \}}) \nonumber\\
&-  \Phi^{-1}(\max_{y' \neq y} \mathbb{E}_\varepsilon \mathbf{1}_{\{\model(x+\varepsilon)=y' \}})],
\label{equ:rand_smooth_db}
\end{align}
where $\Phi$ is the $c.d.f.$ of the standard Gaussian distribution.

\begin{table}[!htbp]
\centering
\caption{Dataset splitting strategy. $\train$ is used to train the target model and serves as the members, while the other $\test$ serves as the non-members. 
$\shadowData$ is used to train the shadow model after relabelled by the target model.}
\setlength{\tabcolsep}{5pt}
\scalebox{0.83}
{
\begin{tabular}{c|cc|cc|cc|cc }
\toprule
Target&  \multicolumn{2}{c}{CIFAR10} & \multicolumn{2}{c}{CIFAR100} & \multicolumn{2}{c}{GTSRB} & \multicolumn{2}{c}{Face}\\
Model    & $\train$ &$\test$ & $\train$ &$\test$& $\train$ &$\test$& $\train$ &$\test$\\
\midrule
$\model$-0 &  3000 & 1000  & 7000& 1000& 600& 500& 350& 100\\
$\model$-1 &  2000 & 1000& 6000& 1000& 500& 500& 300& 100\\
$\model$-2 &  1500 & 1000& 5000& 1000& 400& 500& 250& 100\\
$\model$-3 &  1000 & 1000& 4000& 1000& 300& 500& 200& 100\\
$\model$-4 &  500 & 1000& 3000& 1000& 200& 500& 150& 100\\
$\model$-5 &  100 & 1000& 2000& 1000& 100& 500& 100& 100\\
\midrule
Shadow&\multicolumn{8}{c}{$\shadowData$}\\
Model&\multicolumn{2}{c|}{46000}&\multicolumn{2}{c|}{42000}&\multicolumn{2}{c|}{38109}&\multicolumn{2}{c}{1417}\\
\bottomrule
\end{tabular}
}
\label{table:datasetsplity}
\end{table}

\begin{figure*}[!htbp]
\centering
\begin{subfigure}{0.5\columnwidth}
\includegraphics[width=\columnwidth]{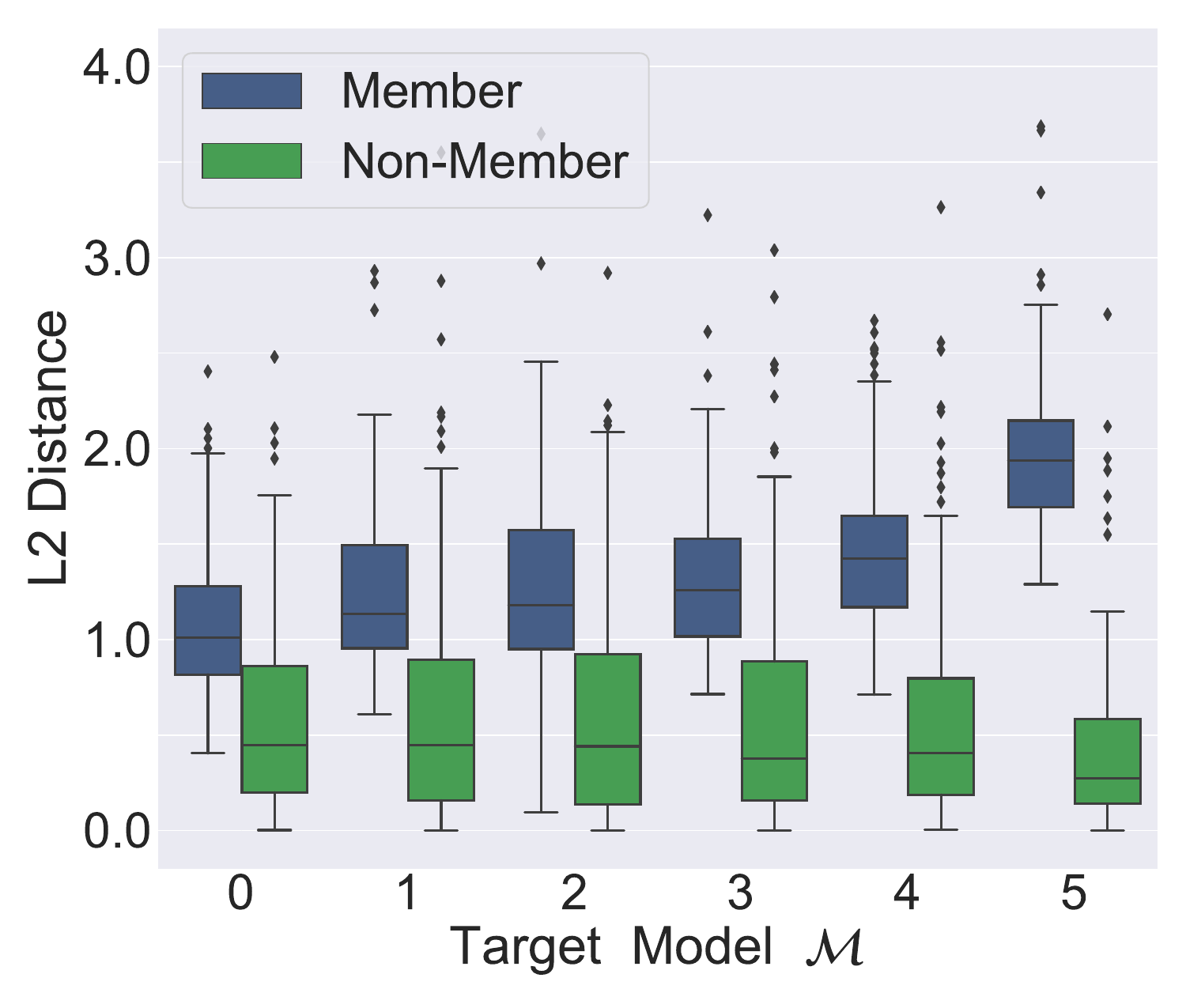}
\caption{CIFAR-10}
\end{subfigure}
\begin{subfigure}{0.5\columnwidth}
\includegraphics[width=\columnwidth]{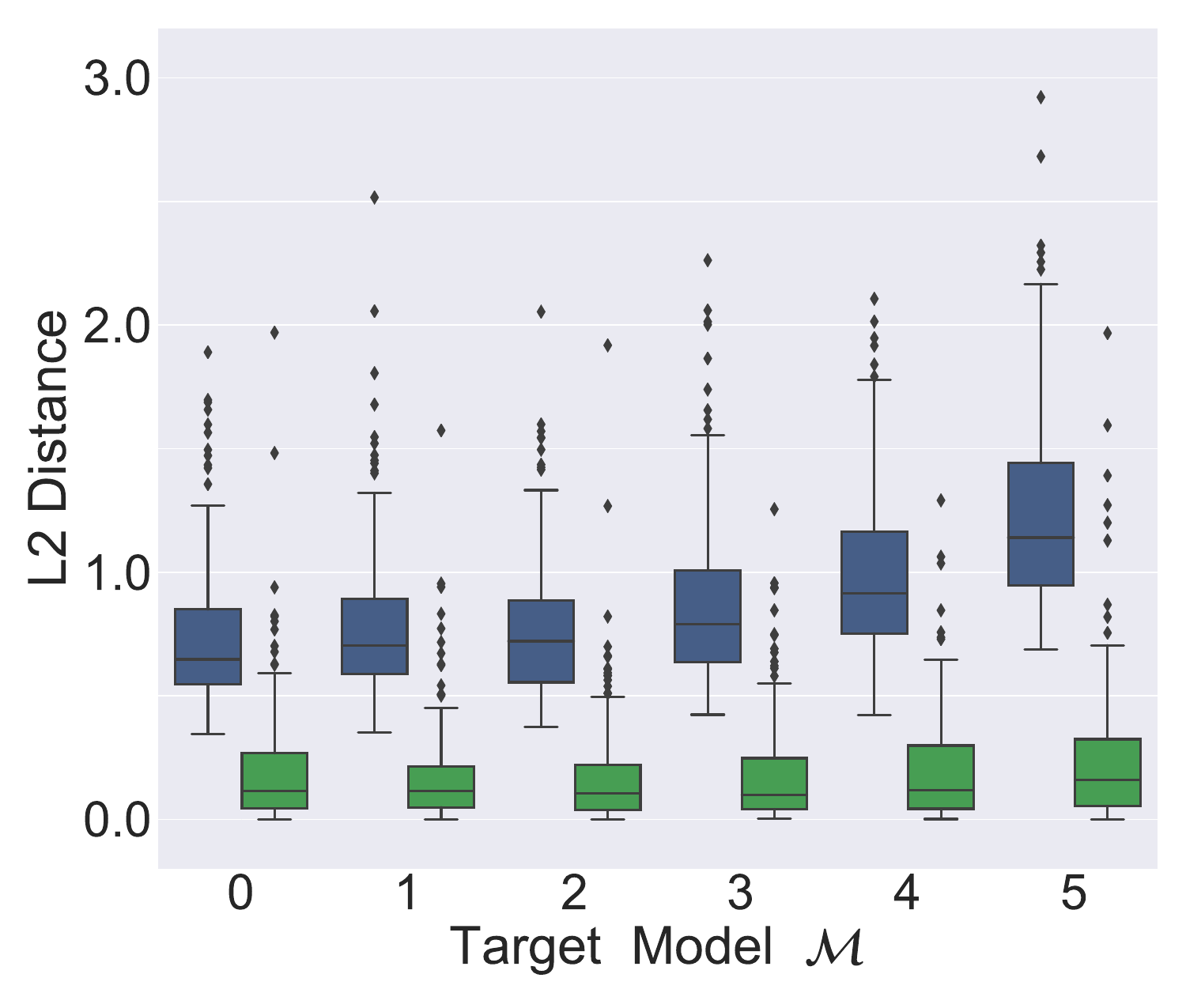}
\caption{CIFAR-100}
\end{subfigure}
\begin{subfigure}{0.5\columnwidth}
\includegraphics[width=\columnwidth]{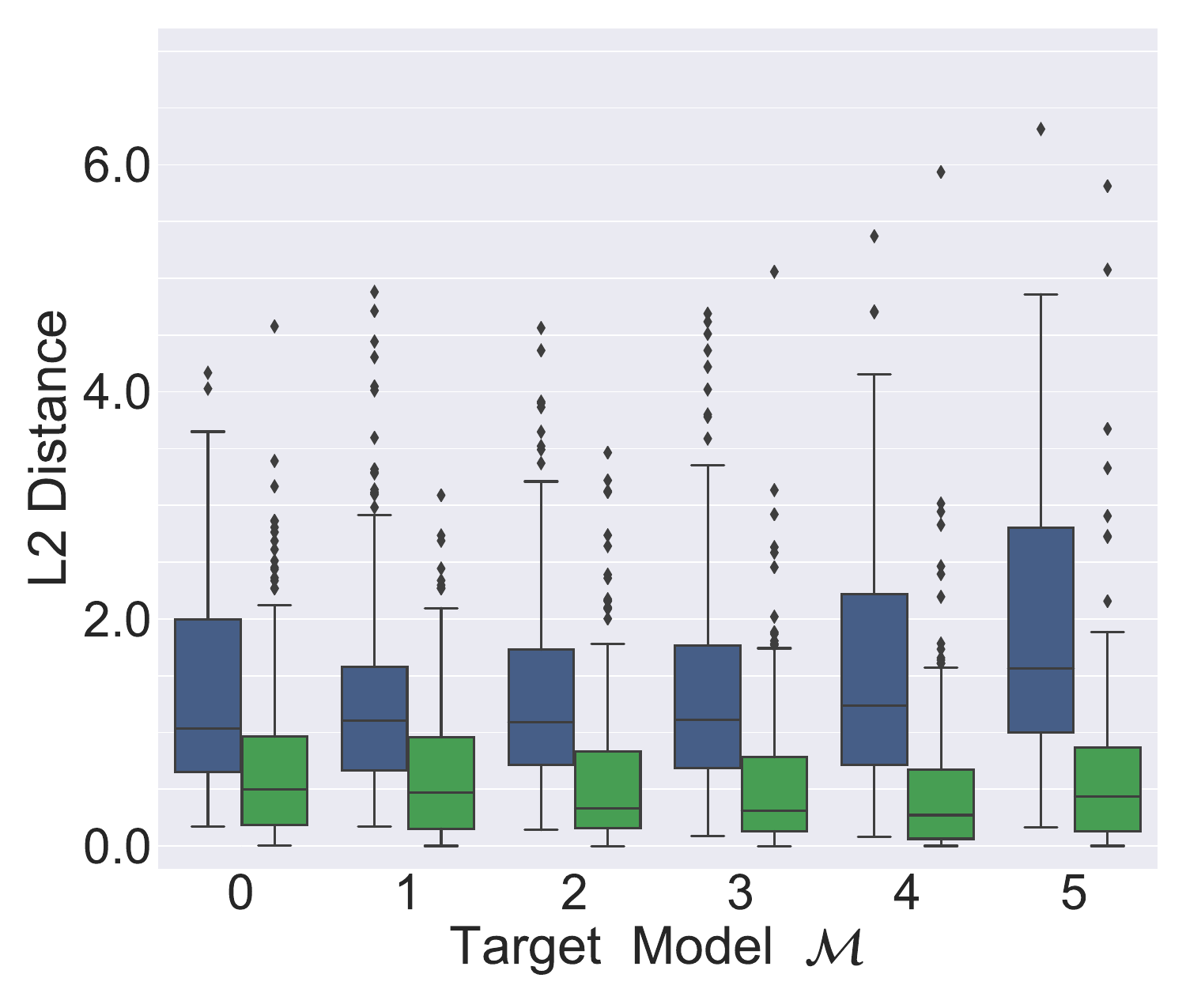}
\caption{GTSRB}
\end{subfigure}
\begin{subfigure}{0.5\columnwidth}
\includegraphics[width=\columnwidth]{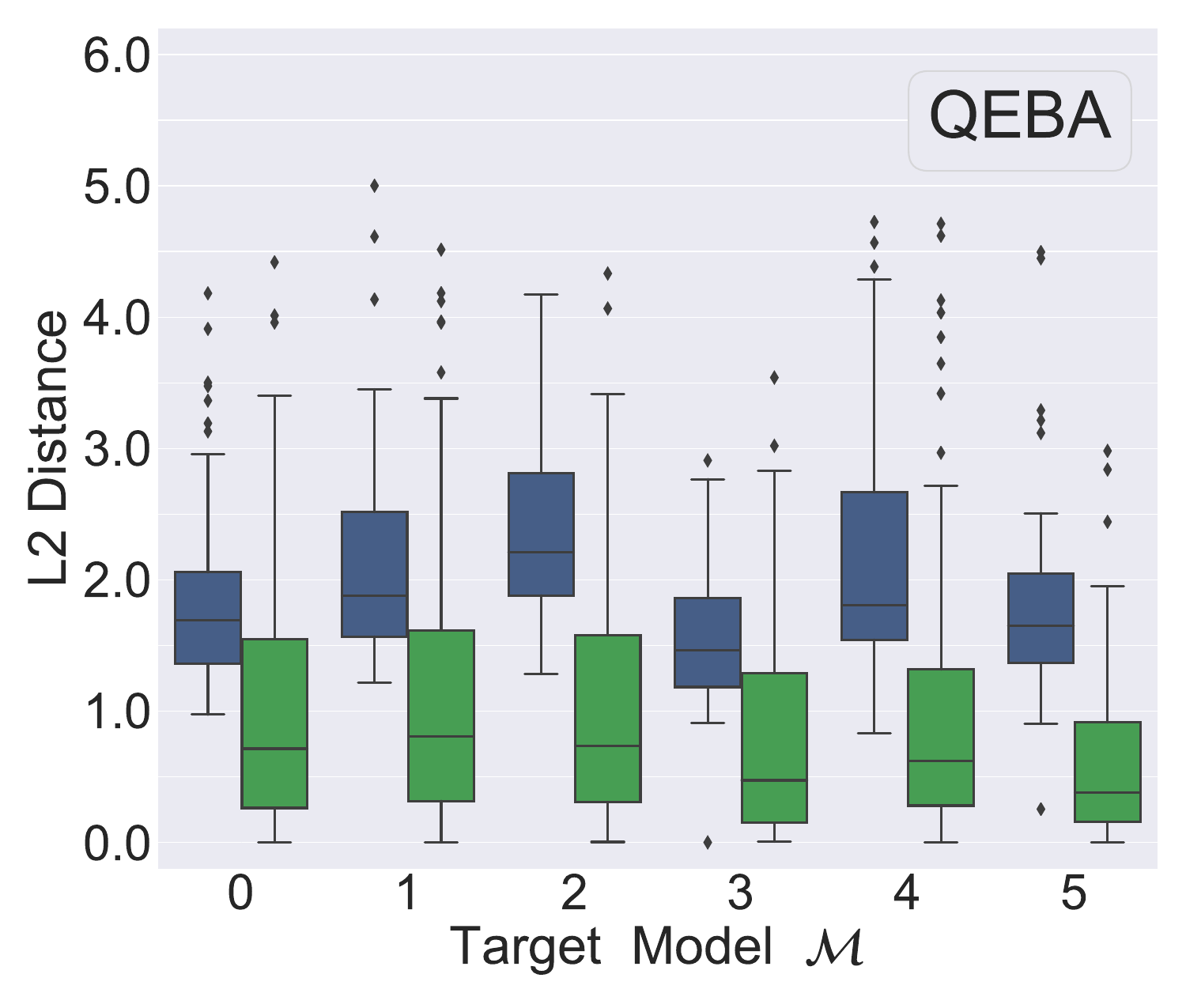}
\caption{Face}
\end{subfigure}
\caption{$L_2$ distance between the original sample and its perturbed samples generated by the QEBA attack. 
The x-axis represents the target model being attacked and the y-axis represents the $L_2$ distance.}
\label{fig:dist_QEBA_SPnoise}
\end{figure*} 

\begin{figure*}[!htbp]
\centering
\begin{subfigure}{0.5\columnwidth}
\includegraphics[width=\columnwidth,height=0.815\columnwidth]{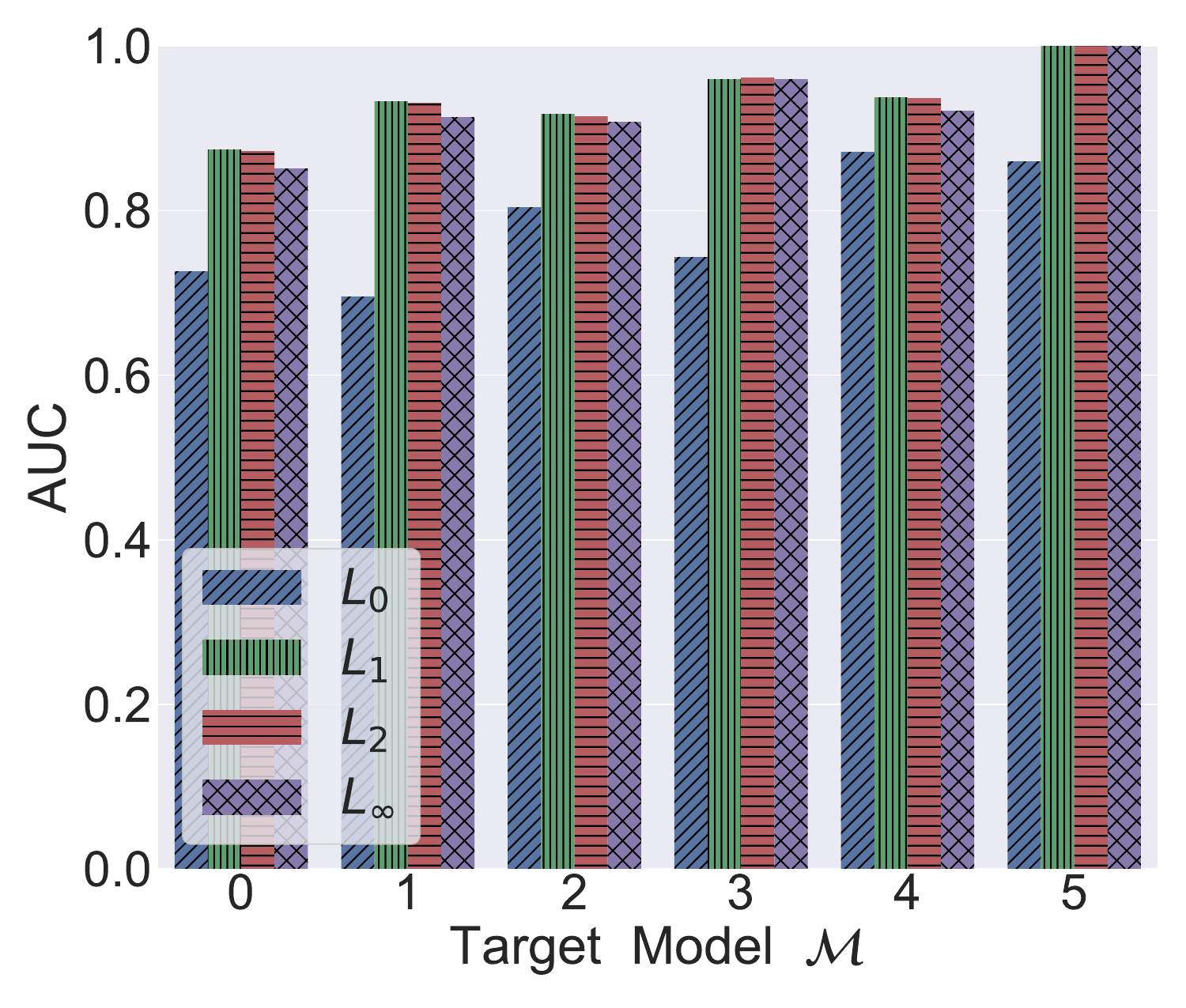}
\caption{CIFAR-10}
\end{subfigure}
\begin{subfigure}{0.5\columnwidth}
\includegraphics[width=\columnwidth,height=0.815\columnwidth]{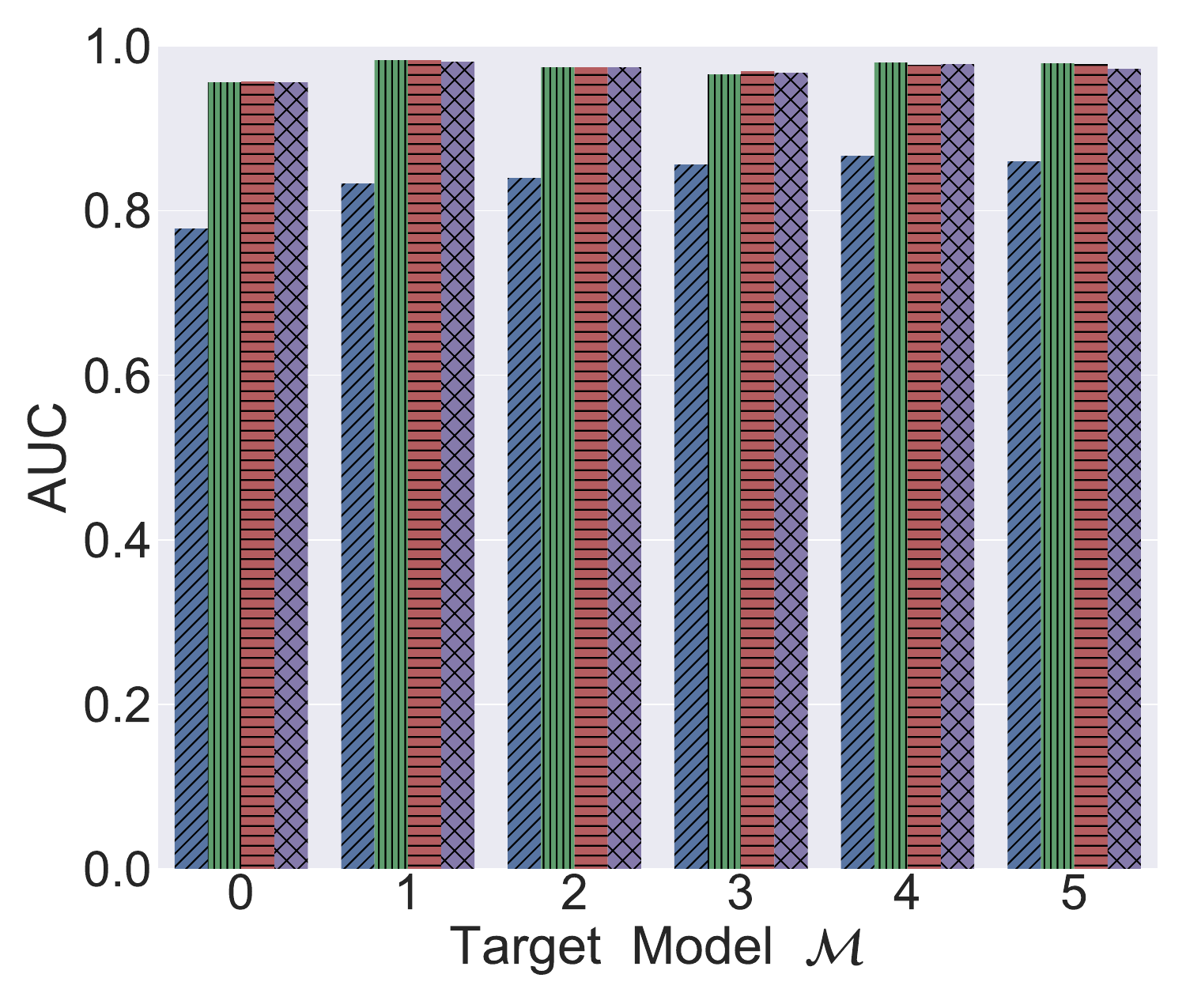}
\caption{CIFAR-100}
\end{subfigure}
\begin{subfigure}{0.5\columnwidth}
\includegraphics[width=\columnwidth,height=0.815\columnwidth]{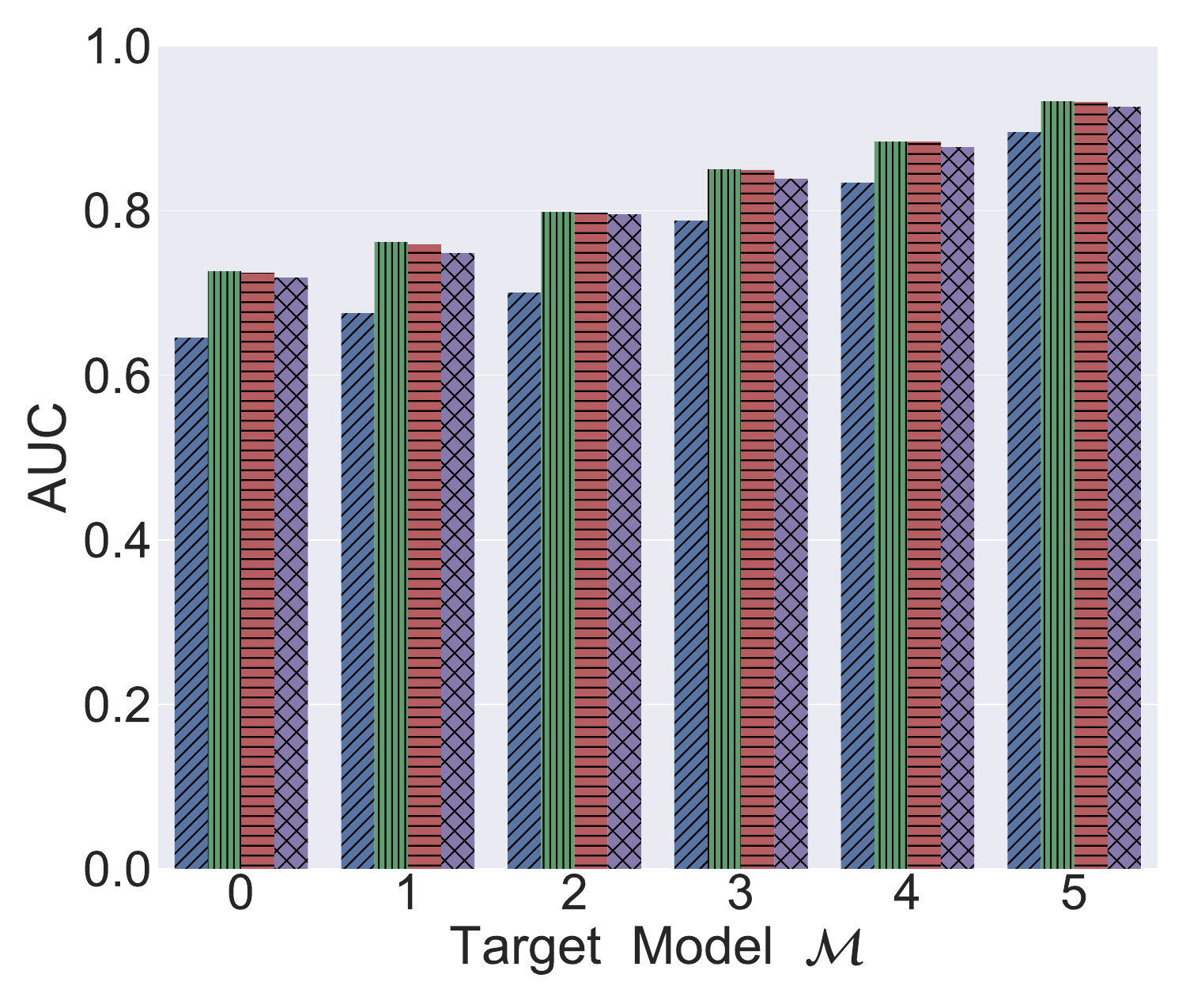}
\caption{GTSRB}
\end{subfigure}
\begin{subfigure}{0.5\columnwidth}
\includegraphics[width=\columnwidth,height=0.815\columnwidth]{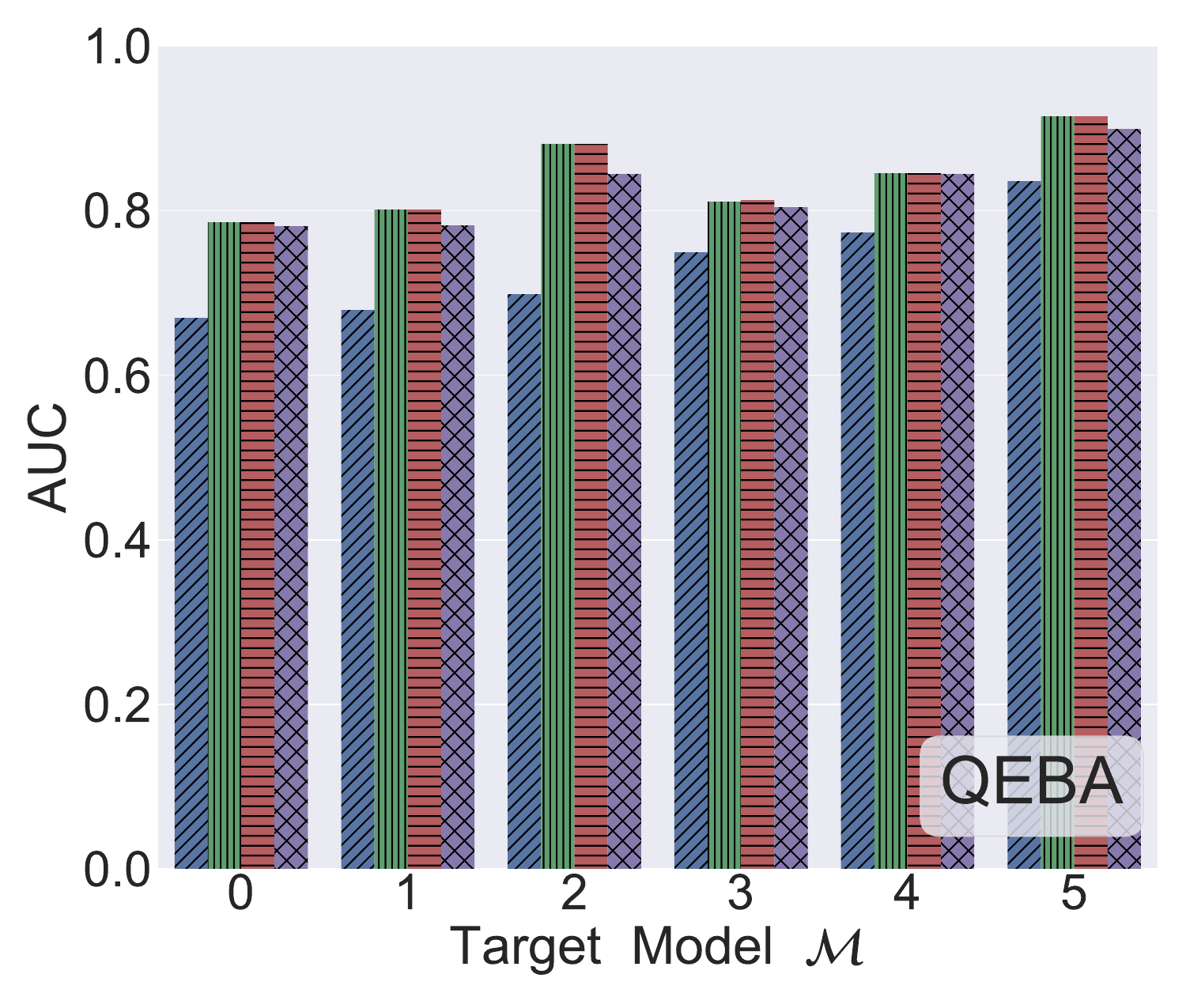}
\caption{Face}
\end{subfigure}
\caption{Attack AUC for four different $L_p$ distances between the original sample and its perturbed samples generated by the QEBA attack.
The x-axis represents the target model being attacked and the y-axis represents the AUC score.}
\label{fig:auc_QEBA_SPnoise}
\end{figure*} 

\begin{algorithm}[!htbp]
\caption{Transfer attack algorithm.}
\label{alg:transfer}  
\KwIn{shadow dataset $\shadowData$, shadow model $\shadow$, target model $\model$, a candidate sample $(x,y)$, threshold $\tau $, minibatch $m$, membership indicator $T$;}
\KwOut{Trained shadow model $\shadow$, $x$ is member or not;}  
Initialize the parameters of $shadow$\;
Relabel $\shadowData$ by querying to $\model$\; 
\For{number of training epochs}
{
 \For{$i=1;i \leq \frac{|\shadowData|}{m};i++$ }{
 sample minibatch of $m$ samples from $\shadowData$\;
 update $\shadow$ by descending its adam gradient}
 }
 Feed $x$ into $\shadow$ to obtain $p_i$\;
 calculate loss: $l=-\sum_{i=0}^{K} \mathbf{1}_{y} \log (p_{i}$)\;
 \uIf {$l\leq \tau$} { 
       $T$ = 1;             \tcc*{$x$ is a member}
} \Else  {
     $T$ = 0;       \tcc*{$x$ is a non-member}
}
return $\shadow$, $T$;  
\end{algorithm} 

\begin{algorithm}[!htbp]
\caption{Boundary attack algorithm.}
\label{alg:boundary}  
\KwIn{adversarial attack technique \emph{HopSkipJump}, target model $\model$, a candidate sample $(x,y) $, threshold $\tau $, membership indicator $T$;}
\KwOut{ $x$ is member or not;}  
\For{number of query}
{
Feed $x$ into $\model$\ to obtain predicted label $y'$\;
 \uIf {$y' \neq y$} { 
       $x'$ = x; \tcc*{perturbed sample $x'$}
} \Else  {
     Apply \emph{HopSkipJump} to perturb $x$ ;
}
}
calculate perturbation $P$ = $|x-x'|_2$\;
\uIf {$P\leq \tau$} { 
       $T$ = 0;             \tcc*{$x$ is a non-member}
} \Else  {
     $T$ = 1;       \tcc*{$x$ is a member}
}
return $T$;  
\end{algorithm} 

\begin{algorithm}[!htbp]
\caption{Threshold choosing for boundary attack.}
\label{alg:threshold}  
\KwIn{adversarial attack technique \emph{HopSkipJump}, target model $\model$; Gaussian distribution $\mathcal{N}(\epsilon ,\beta)$, queue $q$, top $t$ percentile;}
\KwOut{threshold $\tau $;}  
Initialize $q$\;
Sample multiple random samples $\mathcal{X}$ from $\mathcal{N}(\epsilon ,\beta)$
\For{number of random samples}
{ Select one sample $x\in\mathcal{X}$\;
Feed $x$ into $\model$\ to obtain predicted label $y$\;
\For{number of query}{
Apply \emph{HopSkipJump} to perturb $x$ to obtain $x'$ \;
Feed $x'$ into $\model$\ to obtain predicted label $y'$\;
\uIf {$y' \neq y$} { 
   push $|x-x'|_2$ into $q$; 
   break\;
} \Else  {
    $x = x'$\;
 }
}
}
sort $q$ in descending order\;
$\tau$ = $q(t)$\;
return $\tau$;  
\end{algorithm} 

% ======================================================
\end{document}